\documentclass[10pt]{article} 
\usepackage[accepted]{tmlr}

\usepackage{amsmath,amsfonts,bm}

\def\eqref#1{equation~\ref{#1}}

\def\1{\bm{1}}

\DeclareMathAlphabet{\mathsfit}{\encodingdefault}{\sfdefault}{m}{sl}
\SetMathAlphabet{\mathsfit}{bold}{\encodingdefault}{\sfdefault}{bx}{n}

\PassOptionsToPackage{sort}{natbib} 
\usepackage{amssymb,amsmath}
\usepackage{graphicx}
\usepackage{microtype}
\usepackage{tabularx, colortbl, booktabs} %
\usepackage[table,xcdraw,dvipsnames]{xcolor}
\usepackage[normalem]{ulem}
\usepackage{subcaption}
\usepackage{times}
\usepackage{epsfig}
\usepackage{color}
\usepackage[T1]{fontenc}
\usepackage{array}
\usepackage{svg}
\usepackage{placeins}
\usepackage{import}
\usepackage{xifthen}
\usepackage{pdfpages}
\usepackage{transparent}
\usepackage{pifont}
\usepackage{xspace}
\usepackage{multirow}
\usepackage{etoolbox,xpatch}
\usepackage{rotating}
\usepackage{enumitem}
\usepackage{pdflscape}
\usepackage{titletoc}
\usepackage{wrapfig}
\usepackage{abbrv}
\usepackage{tikz}
\usetikzlibrary{calc,tikzmark,arrows.meta} 
\tikzset{>={Stealth[length=2mm]}} 
\usepackage[pagebackref,breaklinks=true,colorlinks]{hyperref}
\renewcommand{\b}{\boldsymbol}
\hypersetup{
    colorlinks=true,
    citecolor=gray,
    breaklinks=true,
}
\usepackage{url}
\usepackage[all=normal, mathspacing=normal,mathdisplays=normal, floats=tight, paragraphs=tight, lists=normal]{savetrees}
\usepackage[capitalize]{cleveref}
\crefname{section}{Sec.}{Secs.}
\Crefname{section}{Section}{Sections}
\crefname{table}{Tab.}{Tabs.}
\Crefname{table}{Table}{Tables}

\usepackage[most]{tcolorbox}

\definecolor{pastelorange}{RGB}{255,179,102} 
\definecolor{DarkTurquoise}{RGB}{0,128,128}   

\newtcolorbox{takeawaybox}{
  enhanced,
  colback=Dandelion!10,
  colframe=Dandelion!50!gray,
  fonttitle=\bfseries,
  coltitle=black,
  title={Takeaway:},             
  attach title to upper={\ },    
  boxrule=2pt,
  arc=4pt,
  left=2pt, right=2pt, top=2pt, bottom=2pt,
}

\title{A Practical Investigation of Spatially-Controlled \\Image Generation with Transformers}

\author{\name Guoxuan Xia, Harleen Hanspal, Petru-Daniel Tudosiu,\email g.xia21@imperial.ac.uk 
\name  Shifeng Zhang \& Sarah Parisot \email zhangshifeng4@huawei.com 
      \addr Work done at Huawei Noah's Ark Lab}

\def\month{11}  
\def\year{2025} 
\def\openreview{\url{https://openreview.net/forum?id=loT6xhgLYK}} 

\begin{document}

\maketitle
\vspace{-2mm}
\begin{abstract}
Enabling image generation models to be \textit{spatially controlled} is an important area of research, empowering users to better generate images according to their own fine-grained specifications via \eg edge maps, poses. Although this task has seen impressive improvements in recent times, a focus on rapidly producing stronger models has come at the cost of detailed and fair scientific comparison. Differing training data, model architectures and generation paradigms make it difficult to disentangle the factors contributing to performance. Meanwhile, the motivations and nuances of certain approaches become lost in the literature.  
In this work, we aim to provide clear takeaways \textit{across} generation paradigms for practitioners wishing to develop \textit{transformer}-based systems for spatially-controlled generation, clarifying the literature and addressing knowledge gaps. We perform \textit{controlled experiments} on ImageNet across diffusion-based/flow-based and autoregressive (AR) models. First, we establish control token prefilling as a simple, general and performant baseline approach for transformers. We then investigate previously underexplored \textit{sampling time} enhancements, showing that extending classifier-free guidance to control, as well as softmax truncation, have a strong impact on control-generation consistency. Finally, we re-clarify the motivation of adapter-based approaches, demonstrating that they mitigate ``forgetting'' and maintain generation quality when trained on limited downstream data, but underperform full training in terms of generation-control consistency.
Code: \url{https://github.com/guoxoug/transformer-imagenet-ctrl}.
\end{abstract}
\vspace{-3mm}
\section{Introduction}\vspace{-1mm}
In recent years, deep-learning-based image generation has advanced at an unprecedented pace, finding widespread adoption in many applications. This has been driven by diffusion-based \citep{scorebased,lipman2023flow}, and more recently, autoregressive  (AR) \citep{var,llamagen} and masked \citep{maskgit,magvit2} modelling approaches that provide a stable and scalable paradigm for learning complex and diverse image distributions. Additionally, \textit{latent space} modelling \citep{taming,LDM} has facilitated efficient, high-resolution generative modelling. Notably, although earlier approaches for large-scale image generation were based on diffusion with the UNet architecture \citep{unet,LDM,sdxl}, the state-of-the-art is now dominated by \textit{transformers} \citep{transformer}, thanks to their scalability \citep{SiT,var,sd3,flux2024}.
%
%
Users most commonly control the content of generated images via text conditioning; however, this lacks spatial precision due to the inherent ambiguity of text and models' limited text understanding \citep{huang2023t2icompbench}. This has led to increasing interest in \textit{spatially-conditioned} image generation, especially for creative applications, where the user can \textit{control} the fine-grained structure of generations via an input such as an edge or depth map \citep{controlnet,t2iadapter}. In this work, we aim to \textit{clarify} design choices for the above task for the latest (transformer-based) generation paradigms, with the objective of providing practitioners with useful and practical takeaways. To this end, we identify a number of \textbf{gaps in the research literature}, which may encumber a practitioner wishing to build their own system:
%

%
1) \textbf{Unclear comparisons}: Existing work on spatially-conditioned image generation tends to focus on rapidly producing high-performing models that can be open-sourced and contribute to the research and general practitioner communities \citep{omnigen,tan2024ominicontrol,controlar,zhang2025unified}. Although this advances the field, it comes at the cost of scientific clarity/understanding. Approaches with different training data (pre-training and control finetuning), different architectures and different generation paradigms are compared against each other, limiting useful insight into why each approach may be more or less effective for practitioners. 
2) \textbf{Sampling enhancements}: Research has focused on new architectures \citep{controlnet,tan2024ominicontrol,controlar}, and new training losses \citep{controlnet_plus_plus,ctrlu} but has neglected \textit{generation-time} algorithmic adjustments, even though such approaches may be simple to implement and offer low-cost post-training improvements.
3) \textbf{Adapters vs full training}: There is a wide range of approaches to the problem, with differing \textit{design motivations} and \textit{architectural choices}, without clear positioning and comparison. In particular, the choice between fully training all parameters versus freezing the weights of a pretrained model and only training an adapter \citep{controlnet} is often poorly presented or ignored.

Motivated by the above, this work aims to provide a clear comparative investigation for spatially-conditioned image generation, with practical takeaways \textit{across} transformer-based architectures and generation paradigms. To this end, we perform \textit{controlled experiments} on ImageNet \citep{imagenet} over two representative but contrasting generative modelling approaches:
%
Scalable Interpolant Transformer (SiT) \citep{SiT} and Visual Autoregressive Modelling (VAR) \citep{var}. The former represents widespread diffusion/flow-based approaches, the latter recently appearing but under-investigated LLM-inspired autoregressive approaches. Our \textbf{key takeaways} are summarised as follows: 

\begin{enumerate}[left=0pt, topsep=0pt,itemsep=-1ex,partopsep=1ex,parsep=1ex]
    \item 
    Out of the plethora of possible approaches, we argue that transformer prefilling with control tokens is an obvious starting point for architectural design. We demonstrate empirically that this simple and general approach is a strong baseline that performs well out of the box across different generation paradigms.
    \item We investigate simple sampling enhancements, finding some to have \textit{meaningful practical impact}. Extending classifier-free guidance (CFG) to spatial control greatly improves generation-control consistency but trades off quality and inference cost, whilst softmax truncation improves \textit{both} image quality \textit{and} control consistency for VAR.
    \item 
    We re-clarify the motivation of control-adapters, where a pretrained generative model is \textit{frozen} and an adapter module is trained -- to \textit{preserve} generation ability and mitigate ``forgetting'' when training on limited/undiverse downstream data. We empirically demonstrate this property, but also find that adapters consistently underperform prefill + fully training all parameters for generation-control consistency when downstream data is not limited.
\end{enumerate}

\vspace{-2mm}\section{Preliminaries}\vspace{-1mm}
\paragraph{Image generation with spatial control.} This is a task popularised by ControlNet \citep{controlnet}, where the user wants to generate an output that adheres spatially to a supplied conditioning image, \eg an edge, depth or segmentation map. This enables greater user control over generations, which is important for creative applications \eg a user may wish to generate a person in a specific pose. Formally, we generate samples $\b x$, given control condition $\b c$ and prompt or class conditioning $\b y$ using a model of the conditional distribution over $\b x$ parameterised by $\b \theta$,
\begin{equation}\label{eq:cond-img}
    \b x \sim p_{\b \theta}(\b x|\b y, \b c)~.
\end{equation}
\paragraph{Latent image modelling.} For high-resolution generation, models are typically broken into two stages: an autoencoder, which maps images to a spatially compressed lower-resolution latent space, and a generative model that operates in this latent space \citep{vqvae,taming,LDM}. 
Since image \emph{semantics} are still preserved at lower resolutions, greater computational efficiency can be achieved by performing generative modelling in the latent space \citep{deepcomp}. Formally, the autoencoder is made up of an encoder $\mathcal{E}$ that compresses image $\b x$ to latent embedding $\b z$ and a decoder $\mathcal{D}$ that aims to reconstruct the image from the latent. Generation is then performed in the \textit{latent} space, with sampled latents then decoded into images using $\mathcal{D}$, 
\begin{equation}
    \b z = \mathcal{E}(\b x), \quad \hat{\b x} = \mathcal{D}(\b z),\quad \b z \sim p_{\b \theta}(\b z|\b y, \b c)~.
\end{equation}
From this point onwards, we assume that all generative modelling is \wrt $\b z$ in the latent space.
\paragraph{Diffusion-based image generation.} The most popular approach for image generation currently is diffusion (or flow)-based models \citep{scorebased,lipman2023flow}. These models learn to denoise data, generating samples by starting from pure noise and iteratively denoising to a clean sample. The \textit{forward process}, over time $t\in [0,1]$, is defined by,
\begin{equation}
    \b z_t = \alpha_t\b z_0 + \sigma_t \b \epsilon, \quad \epsilon \sim \mathcal{N}(\b 0, \b I),
\end{equation}
where noise $\epsilon$ is mixed with the clean sample $\b z_0$ according to the \emph{noise schedule} $(\alpha_t,\sigma_t)$. Clean data $\b z_0$ can then be generated from noise $\b z_1$ by integrating the following \textit{probability flow ODE}\footnote{We omit stochastic samplers/SDE solvers \citep{scorebased} for brevity, as we do not use them in this work.} \citep{scorebased},
\begin{equation}\label{eq:ode}
    \underbrace{\b u_t=\frac{d \b z_t}{dt}}_\text{flow vector} = f_t \b z_t - \frac{1}{2}g^2_t\underbrace{\nabla_{\b z_t}\log\; p(\b z_t)}_{\text{score }\b s(\b z_t)},  \quad \b z_1 \sim \mathcal N(\b z_1;\b 0,\b I),\quad \quad\quad    \text{where } f_t = \frac{d a_t}{dt}\frac{1}{a_t}, \quad g_t^2 = \frac{d \sigma_t^2}{dt} - 2f_t\sigma_t^2.
\end{equation}
Diffusion models are trained to model various parameterisations of the score $\b s_{\b \theta}(\b z_t,t)\approx\nabla_{\b z_t}\log\; p(\b z_t)$, \eg data ($\b z_0$) or noise ($\b \epsilon$) prediction $\b s_{\b \theta}(\b z_t,t) = [\alpha_t\b z_{\b \theta}(\b z_t,t) -\b z_t]/\sigma_t^2=-\b \epsilon_{\b \theta}(\b z_t,t)/\sigma_t$, whilst flow matching directly models the probability flow $\b u_{\b \theta}(\b z_t,t)\approx \b u_t$. We note that $\b u, \b s, \b \epsilon$ can be directly calculated from each other given $\b z_t$ \citep{gao2025diffusionmeetsflow}.
\paragraph{Visual Autoregressive modelling (VAR).} Recently, inspired by the success of large language models, interest has grown in developing autoregressive-style approaches for image generation \citep{taming,llamagen}. In particular, Visual Autoregressive modelling (VAR) \citep{var}, stands out as a strong performer, rivalling the generation quality of diffusion-based approaches. The core idea is to model image generation as a course-to-fine progression of ``scales'', \ie image representations of increasing spatial resolution, allowing a transformer-based model to progressively generate an image conditioned on lower-resolution representations via ``next-scale prediction''. This image-specific approach outperforms existing raster-scan AR models. The joint distribution over scales is modelled as 
\begin{equation}
   p_{\b\theta}(\b r^1, \b r^2,\dots,\b r^K) = \prod\nolimits_{k=1}^Kp_{\b\theta}(\b r^k\mid\b r^1, \b r^2,\dots,\b r^{k-1})\quad  \text{for image scales } \{\b r^1, \b r^2,\dots,\b r^K\},
\end{equation}
allowing each scale of $h^k\times w^k$ tokens to be generated given the sequence of previously generated scales (within-scale tokens are generated in parallel). 
VAR discretises the latent space using a vector-quantised variational autoencoder (VQVAE) \citep{vqvae}, allowing for modelling using softmax and training via cross entropy.\footnote{We omit some of the details with respect to the multi-scale and residual nature of VAR's VQVAE \citep{var} for brevity. We also note that there have been a number of iterations on VAR \citep{ren2025flowar,flexvar}, however, we stick to evaluating the original in this work.} 

\begin{figure}[t]
 \vspace{-4mm}
    \centering
    \includegraphics[width=1.05\linewidth]{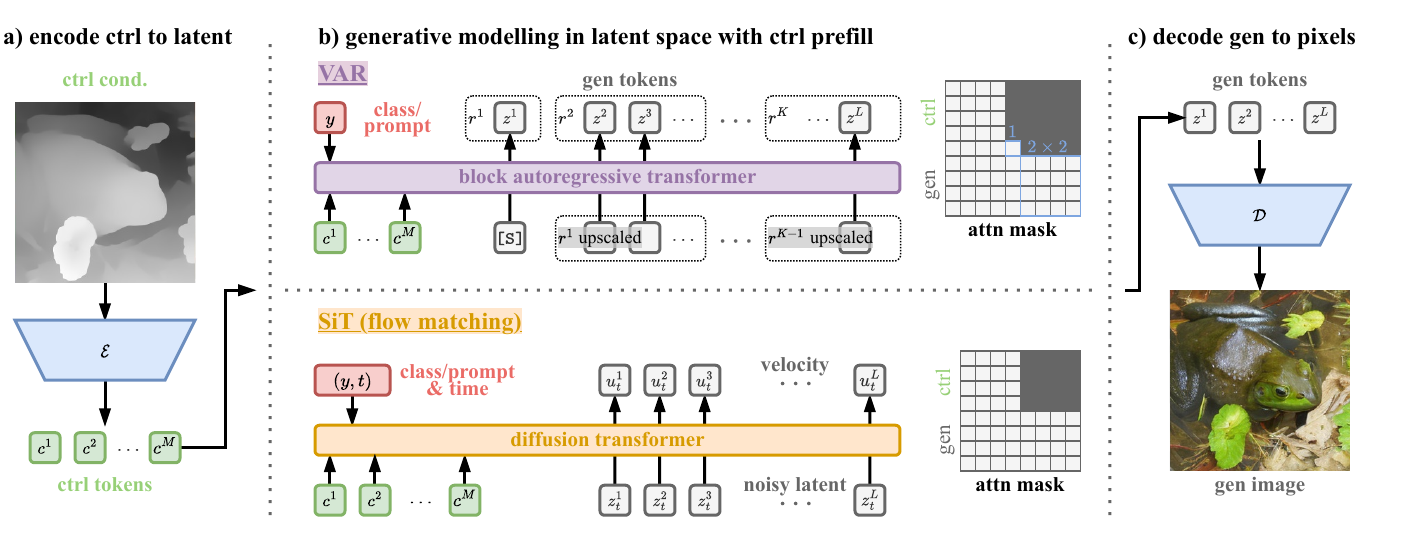}
    \vspace{-3mm}\caption{Illustration of our \textit{prefill} baseline for spatial control with transformers. \textbf{a)} The control conditioning is encoded to the generative model's latent space. \textbf{b)} Generative modelling is performed in the latent space using VAR (top) or SiT (bottom). Generative tokens \textit{attend back} to conditioning control tokens using a \textit{block casual} mask, allowing for KV-caching at inference. \textbf{c)} Generated tokens are decoded from the latent space to the pixel space.}
    \label{fig:prefill_illust}
\end{figure}

\paragraph{Classifier-free guidance (CFG).}  A standard method for improving generation quality is classifier-free guidance \citep{cfg}, which works by adjusting the generative distribution so that probability mass/density is concentrated in regions where the probability of the prompt/class $p(\b y|\b z)$ is implicitly higher,
\begin{equation}\label{eq:cfg}
    p^{\text{guid}}_{\b \theta}(\b z) = \frac{1}{Z}p_{\b \theta}(\b z| \b y)\left[\frac{p_{\b \theta}(\b z| \b y)}{p_{\b \theta}(\b z)}\right]^{\gamma_y-1}, \quad \text{since  } p(\b y|\b z) \propto \frac{p(\b z| \b y)}{p(\b z)}~,
\end{equation}
where $\gamma$ (typically $\geq 1$) is the guidance scale/strength and $Z$ is a normalising constant. To achieve this typically a model is trained to model both $p_{\b \theta}(\b z|\b y)$ and $p_{\b \theta}(\b z)$ via an empty condition $\b y = \varnothing$. Applying guidance to diffusion and VAR, 
\begin{equation}
    \b s^{\text{guid}}_{\b \theta}(\b z_t|\b y) = \b s_{\b \theta}(\b z_t|\b y) +( \gamma_y\!-\!1)\left[\b s_{\b \theta}(\b z_t|\b y)\!-\!\b s_{\b \theta}(\b z_t) \right], \quad
        \b v^{\text{guid}}_{\b \theta}(\b r^{}|\b y) = \b v_{\b \theta}(\b r^{}|\b y) + (\gamma_y\!-\!1)\left[\b v_{\b \theta}(\b r^{}|\b y)\!-\!\b v_{\b \theta}(\b r^{}) \right],
\end{equation}
results in simple linear combinations of the score estimates $\b s_{\b \theta}(\b z_t,t)\approx\nabla_{\b z_t}\log\; p(\b z_t)$  (see \cref{eq:ode}) and pre-softmax logits $\b v$. Note that the normalising constant $Z$ is naturally dealt with for diffusion due to the score being a gradient, whilst in VAR it is taken care of by the softmax denominator. Note that setting $\gamma_y=0$ recovers the unconditional model $ p_{\b \theta}(\b z)$, whilst $\gamma_y=1$ recovers the conditional one $ p_{\b \theta}(\b z|\b y)$. $\gamma_y$ is typically set to be $\geq 1$ in order to improve generation quality. 

\paragraph{Related work in spatially controlled image generation.}
Earlier approaches to this problem involve freezing the UNet \citep{unet} of Stable Diffusion \citep{LDM} and training an additional ``adapter'' module that injected control information via attention (Gligen \citep{gligen}), or addition, (ControlNet \citep{controlnet}, T2I-Adapter \citep{t2iadapter}). Later research focuses on improving ControlNet by unifying controls into a single model \citep{unicontrol,unicontrolnet} and refining the training loss to optimise control-generation consistency \citep{controlnet_plus_plus,ctrlu}. As image generation models diversified from UNets to transformers \citep{transformer}, and from diffusion to AR \citep{var,llamagen} and masked generation \citep{maskgit}, a wider range of architectural approaches have arisen. Adapters for diffusion/flow transformers include Pixart-$\delta$ \citep{chen2024pixartdelta} which adapts ControlNet for Pixart-$\alpha$ \citep{pixartalpha} and OminiControl \citep{tan2024ominicontrol,tan2025ominicontrol2}, which explores attention adapters for Flux \citep{flux2024}. ControlAR \citep{controlar} adds spatial control by fully fine-tuning a raster-scan AR model with an addition-based control encoder, whilst ControlVAR \citep{controlvar} proposes to jointly model image and control data using a VAR model. Recently, there has been a surge in research aiming to fully train unified transformer foundation models that can perform generation and understanding over various tasks and modalities \citep{zhang2025unified}. Notably, Omnigen \citep{omnigen} and UniReal \citep{chen2024UniReal} train spatially-conditioned diffusion transformers where the conditions are simply processed as additional tokens.
This wide array of diverse algorithms and architectures in the literature may seem daunting to newcomers. This work aims to provide some clear and practical takeaways to practitioners wanting to use transformers for spatially controlled image generation.

\section{Evaluation Setup}
\paragraph{Data.} In order to perform a clean and fair comparison and avoid potential confounding factors related to data quality (during both pre-training and control finetuning), we train and evaluate on class-conditioned ImageNet \citep{imagenet}, a well-established benchmark for image generation. ImageNet is at a scale ($\sim$1.2M samples, 256$\times$256 resolution, baseline models with $10^8\sim10^9$ parameters) that is close to deployed real-world image generation models, whilst still having manageable training and evaluation costs for an academic compute budget. 
We train and evaluate at a resolution of 256$\times$256 over two distinct spatial control conditions, which we extract from the ImageNet dataset on the fly during training and evaluation: 1) Canny edge maps \citep{canny}, using \texttt{kornia} with fixed thresholds $(0.1,0.2)$, and 2) dense depth maps using DPT-Hybrid \citep{depth}.

\paragraph{Evaluating generated images.} For most evaluations we generate 10K samples for evaluation, conditioned on controls extracted from the first 10 images of each of the 1000 classes in the ImageNet validation dataset. In a few cases, to compare with the literature, we generate using controls from all 50K validation images. We use fixed random seeds. We measure image generation performance using three metrics. Fréchet Inception Distance (FID) \citep{FID} measures the distributional distance between the data distribution and generated samples, capturing both realism and diversity, whilst Inception Score (IS) \citep{IS} measures the (class-)diversity and clarity of generations, reflecting how confidently a pretrained classifier can assign labels and how varied those labels are across samples. These are widely used image quality metrics.
Generation-control consistency \citep{controlnet_plus_plus} intuitively measures how closely generations adhere to spatial control, via similarity between the original control and the control extracted from the generated image. Following \citet{controlnet_plus_plus}, for Canny edge maps we compare binary pixels using F1 score $\uparrow$, whilst for depth maps we use root mean square error $\downarrow$ (RMSE) for pixel values in $[0,255]$. Consistency is a naturally desirable property, \ie to have the generation closely match the provided conditioning, and is especially motivated by applications with tight spatial tolerances such as industrial design. Additionally we emphasise the following:
\begin{takeawaybox}
    Metrics for generation quality (FID, IS) and control consistency (F1, RMSE) quantify certain aspects of performance; however, practitioners' requirements subjectively depend on their individual use cases (\eg prioritise consistency vs aesthetics, don't care about diversity). Readers should approach the quantitative results presented in this paper considering both the nature of the metrics reported and their own individual requirements.
\end{takeawaybox}
\paragraph{Evaluating inference cost.} We compare generation throughput in images per second (img/s) and latency in milliseconds (ms) by directly measuring time in python. Inference is performed on a single NVIDIA Tesla V100-32GB-SXM2. We do not include the decoder to the pixel space, or the extraction and encoding of control data to the latent space. This is to isolate the impact on the transformer component of the model, as that is the design space we are exploring.
\paragraph{Limitations.} Although the aim of the above experimental setup is to provide practically useful and generalisable takeaways, we also qualify that it is not comprehensive. For example, other than the additional experiment in \cref{app:ctrlnet}, we do not perform experiments on text-to-image models. Neither do we directly evaluate raster-scan AR image generation models. Readers should take this into consideration when interpreting our results/contributions for their own use.
\section{Prefilling -- a Baseline for Transformer-Based Controllable Generation}


In order to investigate design and deployment choices for image generation with spatial control for \textit{transformer}-based  \citep{transformer} model architectures, we need to first establish a baseline approach.
There are many potential options, an obvious choice being to adapt ControlNet from UNet to transformer as in \cite{chen2024pixartdelta}. However, we argue that control token \textit{prefilling} is a natural and simpler starting point. That is to say simply passing spatial control tokens $\b c^1,\dots,\b c^M$ through the transformer stack, and then allowing generation-related tokens $\b z^1,\dots,\b z^L$ to attend back to the control tokens (via block-causal attention mask during training and KV-cache at inference). Compared to other approaches \citep{chen2024pixartdelta,controlar} it is simpler, leaving the architecture untouched, and can be broadly applied to \textit{any} transfomer-based generation paradigm (diffusion, AR, \etc), rather than being \textit{paradigm specific}. In addition, the use of a KV-cache means that inference is only performed once on the conditioning tokens, whereas other approaches may incur considerably more overhead running once for each generation step \citep{tan2024ominicontrol,tan2025ominicontrol2}. Finally, this simpler, unified single-transformer approach is of particular interest as it is increasingly popular in the literature \citep{zhang2025unified}, being applied beyond spatial control to multimodal generation, understanding and editing. Notably, the recent foundation model Omnigen \citep{omnigen} adopts this prefilling approach for spatial control.


We investigate two recent transformer-based approaches for image generation, to cover both diffusion-based and (comparatively underexplored) autoregressive methods: 1) SiT \citep{SiT}, which trains a diffusion transformer (DiT) architecture \citep{dit} using flow matching, and 2) VAR, which autoregressively generates image representations of increasing spatial scale \citep{var}. \cref{fig:prefill_illust} illustrates our prefill baseline for both cases.
\paragraph{Implementation and experimental details.}
To implement the above prefilling approach (\cref{fig:prefill_illust}), we simply fine-tune the existing architectures with additional conditioning tokens, which the generation tokens attend to via a block-causal mask. We consider VAR-d16 and SiT-XL/2, initialising both models with ImageNet-pretrained checkpoints.\footnote{VAR: \url{https://github.com/FoundationVision/VAR},  SiT: \url{https://github.com/sihyun-yu/REPA} \citep{repa}} The base architectures are left unchanged. For VAR we add new learnable positional embeddings for the control tokens and a shared additive ``level'' embedding for all control tokens, whilst for SiT we only add a new ``level'' embedding for the control tokens, reusing the original sinusoidal positional embeddings. For SiT, we set diffusion time $t=0$ (clean data) for control tokens. We finetune for 10 epochs with control conditioning with batch size 256 ($\sim$50K iterations) using the original optimisers and hyperparameters. Following the original papers we linearly increase guidance scale $\gamma_y$  from zero over generation scales for VAR, whilst keeping it constant for SiT. CFG is performed by only omitting class conditioning and not control conditioning. VAR and SiT use different image crops, leading to small differences in visualisations.
\paragraph{Results.}
\begin{figure}[]
 \vspace{-4mm}
\begin{minipage}{0.42\linewidth}
    \includegraphics[width=\linewidth]{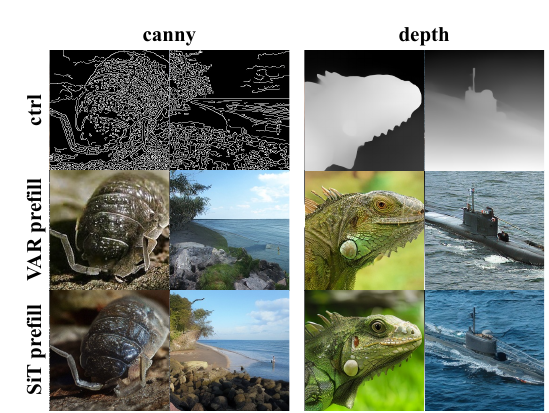}
\end{minipage}
\hfill
\begin{minipage}{0.56\textwidth}

\colorlet{CompareArrow}{Dandelion!90!black}

\resizebox{\textwidth}{!}{%
\begin{tabular}{@{}lll llll ll ll@{}}
\toprule
\multirow{4}{*}{\begin{tabular}[c]{@{}l@{}}model\\ \scriptsize(sampling params)\end{tabular}} &
\multirow{4}{*}{\begin{tabular}[c]{@{}l@{}}control\\ conditioning\end{tabular}} &
\multirow{4}{*}{\#params} &
\multicolumn{4}{c}{gen. quality} &
\multicolumn{2}{c}{ctrl consistency} &
\multicolumn{2}{c}{inf. cost} \\
\cmidrule(lr){4-7} \cmidrule(lr){8-9} \cmidrule(lr){10-11}
 &  &  &
  \begin{tabular}[c]{@{}l@{}}FID$\downarrow$\\ 50K\end{tabular} &
  \begin{tabular}[c]{@{}l@{}}IS$\uparrow$\\ 50K\end{tabular} &
  \begin{tabular}[c]{@{}l@{}}FID$\downarrow$\\ 10K\end{tabular} &
  \begin{tabular}[c]{@{}l@{}}IS$\uparrow$\\ 10K\end{tabular} &
  F1$\uparrow$ &
  RMSE$\downarrow$ &
  \begin{tabular}[c]{@{}l@{}}lat$\downarrow$ \\ (s) \\ bs=1\end{tabular} &
  \begin{tabular}[c]{@{}l@{}}TP$\uparrow$ \\ (img/s)\\ bs=16\end{tabular} \\
\midrule
\multicolumn{2}{l}{{\color[HTML]{656565} \begin{tabular}[c]{@{}l@{}}ImageNet-val \\ (ctrl source)\end{tabular}}} &
   &
  {\color[HTML]{656565} 1.77} &
  {\color[HTML]{656565} 237} &
  {\color[HTML]{656565} 4.44} &
  {\color[HTML]{656565} 187} &
   &
   &
   &
   \\ \midrule
 &
  {\color[HTML]{656565} no ctrl} &
   &
  {\color[HTML]{656565} 3.36} &
  {\color[HTML]{656565} 277} &
  {\color[HTML]{656565} 5.71} &
  {\color[HTML]{656565} 214} &
   &
   &
  {\color[HTML]{656565} 0.26} &
  {\color[HTML]{656565} 17} \\
 &
  canny prefill &
   &
  3.94~\tikzmark{aTop}&
  190 &
  6.60 &
  158 &
  34.6~\tikzmark{dTop}\tikzmark{bTop}&
   &
   &
   \\
 &
  {\color[HTML]{840101} + ctrl quant} &
   &
  {\color[HTML]{840101} 3.95} &
  {\color[HTML]{840101} 193} &
  {\color[HTML]{840101} 6.65} &
  {\color[HTML]{840101} 156} &
  {\color[HTML]{840101} 30.9~\tikzmark{dBot}} &
   &
   &
   \\
\multirow{-4}{*}{\begin{tabular}[c]{@{}l@{}}VAR-d16\\ \scriptsize CFG=2.5, temp=1, \\ \scriptsize top-$p$=0.96, top-$k$=900\end{tabular}} &
  depth prefill &
  \multirow{-4}{*}{310M} &
  3.30 &
  207&
  5.95 &
  165 &
   &
  31.3 &
  \multirow{-3}{*}{0.31} &
  \multirow{-3}{*}{12} \\ \midrule
 &
  {\color[HTML]{656565} no ctrl} &
   &
  {\color[HTML]{656565} 2.01} &
  {\color[HTML]{656565} 300} &
  {\color[HTML]{656565} 4.47} &
  {\color[HTML]{656565} 228} &
   &
   &
  {\color[HTML]{656565} 3.0} &
  \tikzmark{cTop}{\color[HTML]{656565} 0.39} \\
 &
  canny prefill &
   &
  3.04 &
  205 &
  5.69 &
  161 &
  39.3 &
   &
   &
   \\
\multirow{-3}{*}{\begin{tabular}[c]{@{}l@{}}SiT-XL/2\\ \scriptsize CFG=1.5, Euler ODE,\\ \scriptsize steps=64\end{tabular}} &
  depth prefill &
  \multirow{-3}{*}{675M} &
  2.78 &
  218 &
  5.43 &
  170 &
   &
  29.3 &
  \multirow{-2}{*}{3.4} &
  \multirow{-2}{*}{\tikzmark{cBot}0.35} \\ \midrule
{\color[HTML]{656565} } &
  {\color[HTML]{656565} canny} &
  {\color[HTML]{656565} } &
  {\color[HTML]{656565} 7.85} &
  {\color[HTML]{656565} 162} &
   &
   &
   &
   &
   &
   \\
\multirow{-2}{*}{{\color[HTML]{656565} \begin{tabular}[c]{@{}l@{}}ControlVAR*\\ VAR-d30\end{tabular}}} &
  {\color[HTML]{656565} depth} &
  \multirow{-2}{*}{{\color[HTML]{656565} 2.0B}} &
  {\color[HTML]{656565} 6.50} &
  {\color[HTML]{656565} 182} &
   &
   &
   &
   &
   &
   \\ \midrule
{\color[HTML]{656565} } &
  {\color[HTML]{656565} canny} &
  {\color[HTML]{656565} } &
  {\color[HTML]{656565} 7.69}~\tikzmark{aBot}&
   &
   &
   &
  {\color[HTML]{656565} 34.9}~\tikzmark{bBot}&
   &
   &
   \\
\multirow{-2}{*}{{\color[HTML]{656565} \begin{tabular}[c]{@{}l@{}}ControlAR*\\ LlamaGen-L\end{tabular}}} &
  {\color[HTML]{656565} depth} &
  \multirow{-2}{*}{{\color[HTML]{656565} 343M}} &
  {\color[HTML]{656565} 4.19} &
   &
   &
   &
   &
  {\color[HTML]{656565} 31.1} &
   &
   \\ \bottomrule
\end{tabular}
\begin{tikzpicture}[remember picture,overlay]
  \tikzstyle{CompareLine}=[draw=CompareArrow,line width=2pt,<->] 
  \tikzstyle{CompareLabel}=[font=\bfseries, text=CompareArrow, anchor=west]

\newcommand{\CompareV}[6][]{%
  \draw[CompareLine,#1] 
    ($ (pic cs:#2) + (#5,#6) $) -- ($ (pic cs:#3) + (#5,0) $);
  \node[CompareLabel,#1] 
    at ($($(pic cs:#2)+(#5,#6)$)!0.5!($(pic cs:#3)+(#5,0)$) + (-0.2em,0)$) {#4};
}

\CompareV[]{aTop}{aBot}{(a)}{0.0ex}{1.2ex} 
\CompareV[]{bTop}{bBot}{(b)}{8ex}{1.2ex}
\CompareV[]{cTop}{cBot}{(c)}{5ex}{1.2ex}
  \CompareV[]{dTop}{dBot}{(d)}{0.5ex}{1.2ex}
\end{tikzpicture}

}

\end{minipage}

\caption{\textbf{Left:} Examples of conditional generations using prefilling. Additional examples can be found in \cref{app:example-gens}. \textbf{Right:} Results for our prefill baseline where each model is finetuned from an ImageNet-pretrained base model. ``*'' indicates results copied from \cite{controlar,controlvar}. With default sampling parameters, prefilling is able to \textcolor{Dandelion!90!black}{\textbf{(a)}} generate higher quality images than recent transformer-based approaches with \textcolor{Dandelion!90!black}{\textbf{(b)}} comparable control consistency. We note the \textcolor{Dandelion!90!black}{\textbf{(c)}} relatively low overhead for SiT, as KV-caching means the control tokens use up only a single forward pass. \textcolor{Dandelion!90!black}{\textbf{(d)}} Quantising the control input with VQVAE hurts consistency for VAR, without affecting generation quality.}
\label{fig:baseline_results}
\vspace{-3mm}
\end{figure}
\cref{fig:baseline_results} shows results for our prefill baseline using default sampling parameters from the original papers.\footnote{Although we reduce the number of sampling steps for SiT to 64 to reduce evaluation time.}\footnote{We note that in the VAR paper CFG is defined with $\hat\gamma=\gamma + 1$, so in our convention the CFG scale is higher by 1 compared to \citet{var}.} We find that prefilling is able to achieve better image generation quality (FID$\downarrow$) compared to existing transformer-based approaches \citep{controlvar,controlar} in the literature whilst achieving comparable control consistency (much better in the case of SiT). Notably our VAR-d16 considerably outperforms ControlVAR d-30, suggesting that the image-control joint modelling proposed by \citep{controlvar} is suboptimal for conditional generation. We note that control-conditioned IS is lower than for generations without control; however this is to be expected since the control inputs are sourced from the ImageNet validation dataset, which naturally has lower IS/more ambiguous samples.
We also find that the use of KV-caching means that the latency and throughput overheads of prefilling compared to no-ctrl generation are quite reasonable, especially for SiT where the cache is calculated only once compared to the 64 denoising steps.
\begin{takeawaybox}
    Prefilling (\cref{fig:prefill_illust}) is a simple, general and effective baseline for spatial control with transformers.
\end{takeawaybox}
We note that our result goes against the results in ControlAR  \citep{controlar} that suggest that prefilling is a poor choice (they discard this design choice and opt for additive injection of control information, which is what we report in \cref{fig:baseline_results}). We hypothesise that \citet{controlar}'s choices to 1) vector \textit{quantise} the control input, destroying information, and 2) enforce a triangular causal attention mask, limiting attention between control tokens, may have hurt performance. We perform our own experiment with VAR's VQVAE (\cref{fig:baseline_results}) confirming the negative effect of quantising the control input. Although avoiding quantisation may sound obvious, AR models are often trained using token indices as inputs by default \citep{llamagen}, meaning that ``turning off'' quantisation may lead to additional implementation complexity.
\begin{takeawaybox}
    Vector quantisation of the control input hurts generation-control consistency.
\end{takeawaybox}
\section{The Impact of Sampling Parameters}\label{sec:samp_params}
For our intial experiments we left sampling/inference hyperparameters untouched compared to the original VAR \citep{var} and SiT \citep{SiT} papers, however, is it natural to ask whether adjusting sampling parameters at generation time can provide improved performance ``for free''. To this end, we consider, classifer-free guidance (CFG), control guidance (ctrl-G) and sampling distribution truncation.  We note that this aspect of generation has been surprisingly neglected in the literature, with research tending to focus on new architectures \citep{controlar} or improved training losses \citep{controlnet_plus_plus,ctrlu}, often presenting evaluations with a single hyperparameter setting. 
\subsection{Control Guidance (ctrl-G)}\label{sec:ctrlg}

Consider the idea of CFG introduced in \cref{eq:cfg}; it can be easily extended to the spatial control condition $\b c$, to concentrate sampling density in regions where $p(\b c|\b z)$ is implicitly higher, potentially improving control consistency,
\begin{align}\label{eq:ctrl-g}
    p^{\text{guid}}_{\b \theta}(\b z) = \frac{1}{Z}p_{\b \theta}(\b z| \b y, \b c)&{\underbrace{\left[\frac{p_{\b \theta}(\b z| \b y, \b c)}{p_{\b \theta}(\b z|\b c)}\right]}_\text{class/prompt}}^{\gamma_y-1}{\underbrace{\left[\frac{p_{\b \theta}(\b z| \b c)}{p_{\b \theta}(\b z)}\right]}_\text{spatial control}}^{\gamma_c-1}, \quad\quad p(\b y|\b z, \b c) \propto \frac{p(\b z| \b y, \b c)}{p(\b z|\b c)},\quad p(\b c|\b z) \propto \frac{p(\b z| \b c)}{p(\b z)}~,
\end{align}
giving updated guidance equations for diffusion score estimates $\b s$ and VAR logits $\b v$,
\begin{align}
        \b s^{\text{guid}}_{\b \theta}(\b z_t|\b y, \b c) = &\b s_{\b \theta}(\b z_t|\b y,\b c) \quad+( \gamma_y-1)\left[\b s_{\b \theta}(\b z_t|\b y, \b c)-\b s_{\b \theta}(\b z_t| \b c) \right]
        +( \gamma_c-1)\left[\b s_{\b \theta}(\b z_t|\b c)-\b s_{\b \theta}(\b z_t) \right]\label{eq:score-ctrl-guid}\\
        \b v^{\text{guid}}_{\b \theta}(\b r|\b y, \b c) = &\b v_{\b \theta}(\b r|\b y, \b c) \quad+(\gamma_y-1)\left[\b v_{\b \theta}(\b r|\b y, \b c)-\b v_{\b \theta}(\b r|\b c) \right] +(\gamma_c-1)\left[\b v_{\b \theta}(\b r|\b c)-\b v_{\b \theta}(\b r) \right] ~.
\end{align}
In the above equations, $\gamma_y$ controls the extent of CFG, whilst $\gamma_c$ influences the control guidance (ctrl-G). Extending CFG to additional conditions has been proposed in prior work \citep{pix2pix,controlvar}, however, its effect on spatial control has not been well explored. For our application, prior work does not vary over $(\gamma_y,\gamma_c)$ independently for evaluation, \eg \citet{omnigen} report a single parameter setting whilst \citet{controlvar} set $\gamma_y = \gamma_c$. We also note that using ctrl-G incurs extra cost, requiring three model inferences rather than the two needed for CFG.
\paragraph{Oversaturation for diffusion/flow-based generation.} 
\begin{figure}
    \vspace{-8mm}
    \centering
    \includegraphics[width=\linewidth]{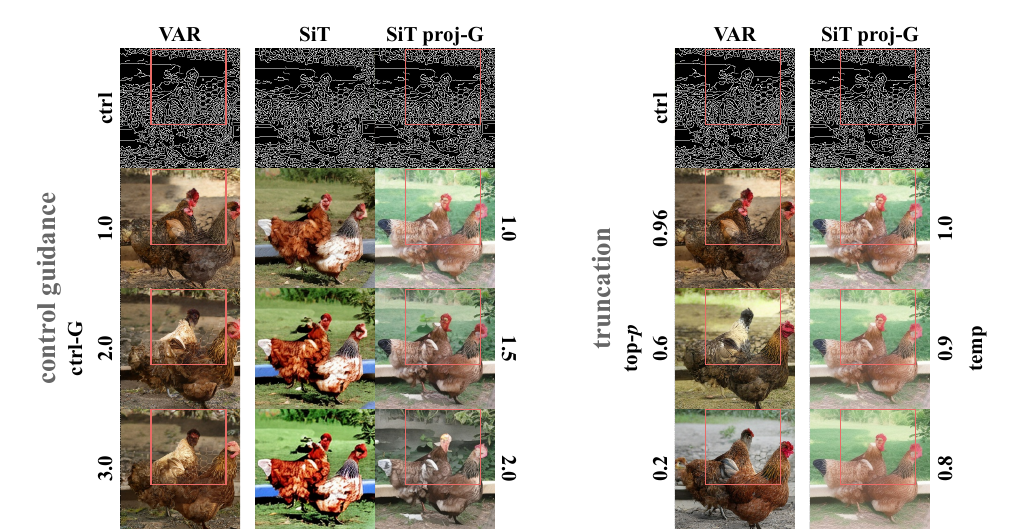}
    \vspace{-3mm}\caption{Visualisation of the effect of adjusting sampling parameters, CFG=3.0 (please zoom in for details). \textbf{Left (ctrl-G):} consistency to the canny map is visibly improved, but this may come at the cost of image quality, \eg over-emphasised edges or incongruous shapes, \eg note the treatment of what was originally a hexagonal metal wire mesh in the original control image. SiT suffers from saturation artefacts; applying projected guidance (proj-G) ameloriates this issue. \textbf{Right (distribution truncation):} Top-$p$ softmax truncation visibly improves VAR consistency without hurting generation quality. Temperature scaling the score has little visible effect on the generation.}
    \label{fig:vis_sampling_params}
\end{figure}
Applying ctrl-G to SiT quickly runs into an issue -- generations become highly saturated, shown in \cref{fig:vis_sampling_params}.
Thankfully, \citet{projg} recently propose \textit{projected} guidance (proj-G) to deal with contrast/saturation issues arising in large CFG scenarios. Applying their approach to ctrl-G, we get,
\begin{equation}
            \b s^{\text{proj-guid}}_{\b \theta}(\b z_t|\b y, \b c) = [\alpha_t \b z^{\text{proj-guid}}_{\b \theta}(\b z_t|\b y,\b c) -\b z_t]/\sigma_t^2,\quad \b z^{\text{proj-guid}}_{\b \theta}(\b z_t|\b y,\b c)= \b z_{\b \theta}(\b z_t|\b y,\b c) + [\b g-(\b g\cdot\hat{\b z})\hat{\b z}],\label{eq:proj-g} 
\end{equation}
\vspace{-4mm}
\begin{equation}
    \text{where}\quad\b g =  ( \gamma_y-1)\left[\b z_{\b \theta}(\b z_t|\b y, \b c)-\b z_{\b \theta}(\b z_t| \b c) \right]
        + ( \gamma_c-1)\left[\b z_{\b \theta}(\b z_t|\b c)-\b z_{\b \theta}(\b z_t) \right],\quad\quad \hat{\b z}=\frac{\b z_{\b \theta}(\b z_t|\b y,\b c)}{||\b z_{\b \theta}(\b z_t|\b y,\b c)||}
\end{equation}
The model is reparameterised to \textit{data} ($\b z_0$) prediction and only the component of guidance that is \textit{orthogonal} to the fully conditioned data estimate $\b z_{\b \theta}(\b z_t|\b y,\b c)$ is used. \citet{projg} show that this orthogonal component is chiefly responsible for quality improvements, whilst the other (parallel) component increases saturation, as it intuitively extends the data prediction $\b z_{\b \theta}$. We find this successfully fixes the saturation issue, without otherwise negatively affecting generations, hence we adopt proj-G for all our following SiT experiments. We also note that the \textit{constrained} distribution over the discretised latent space of a VQVAE modelled by VAR is not susceptible to these saturation issues.
\begin{takeawaybox}
    For flow/diffusion models, control guidance (\cref{eq:score-ctrl-guid}) may lead to oversaturation. Applying projected guidance \citep{projg} (\ie only using the orthogonal component \cref{eq:proj-g}) assuages this issue. 
\end{takeawaybox}
\paragraph{Results.}
\begin{figure}[t]
 \vspace{-4mm}
    \centering
    \includegraphics[width=.49\linewidth]{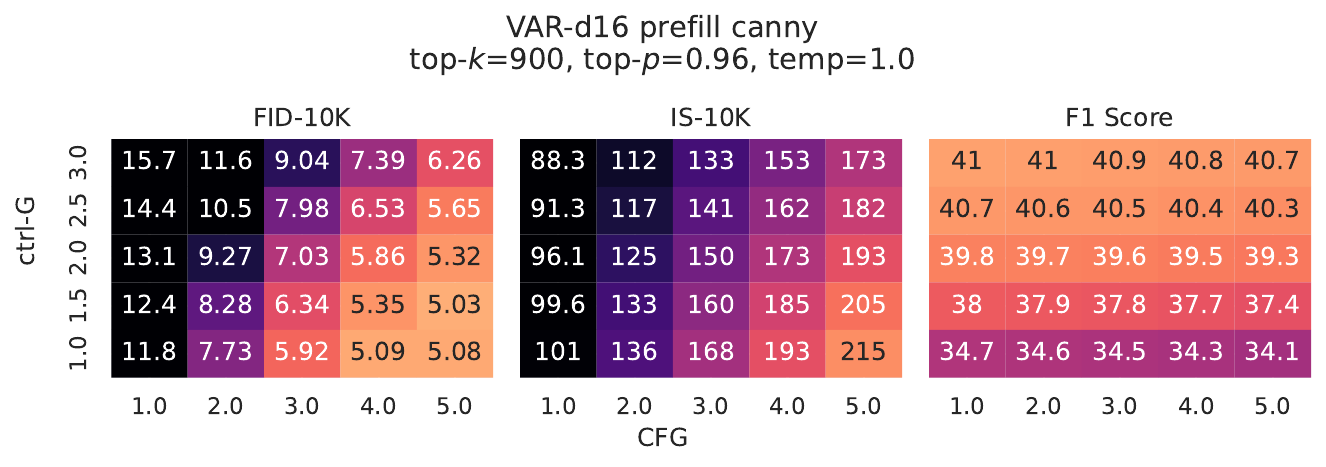}
    \includegraphics[width=.49\linewidth]{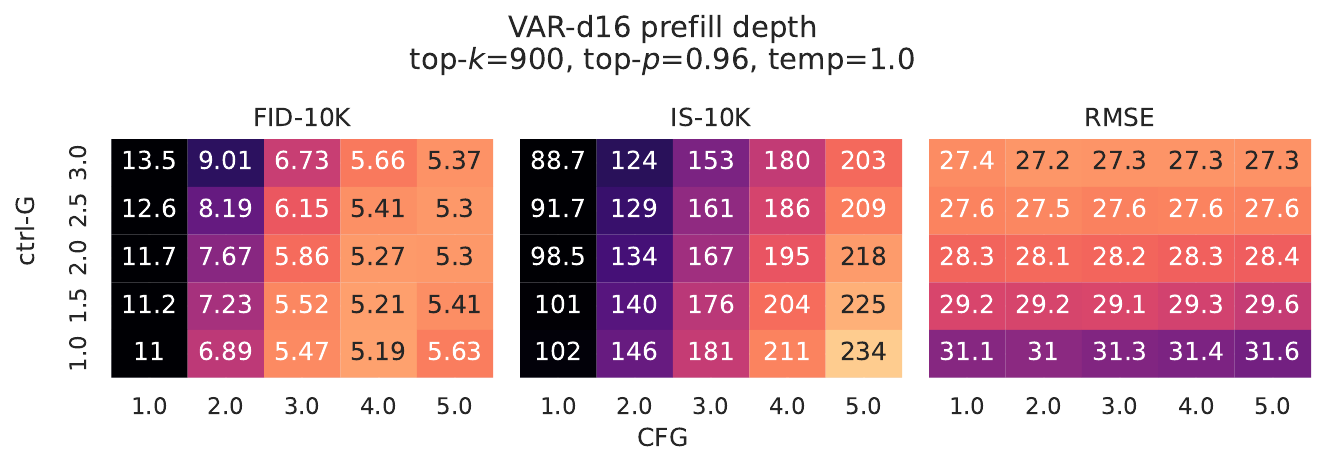}

    \includegraphics[width=.49\linewidth]{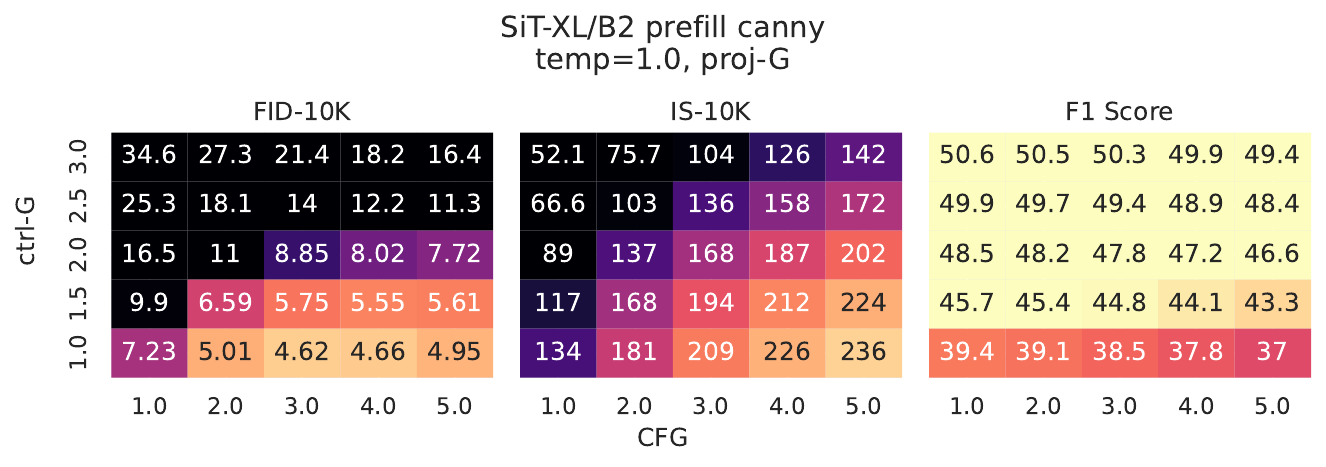}
    \includegraphics[width=.49\linewidth]{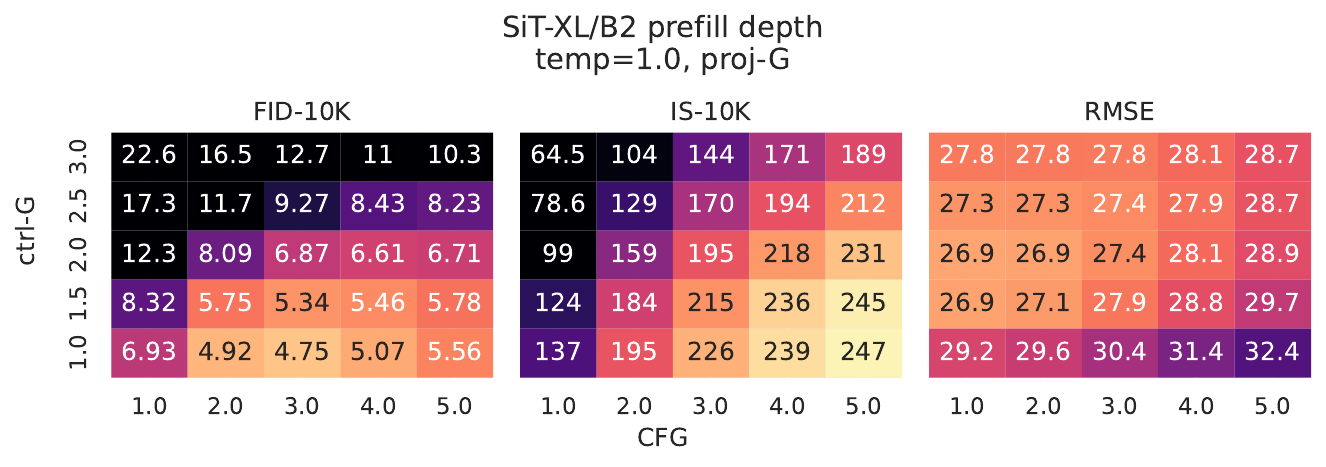}
    \vspace{-3mm}\caption{The effect of CFG and Ctrl-G on conditional generation. Brighter means better. CFG generally improves generation quality according to FID$\downarrow$ and IS$\uparrow$, with a slight decrease in control consistency (F1$\uparrow$, RMSE$\downarrow$). \textbf{Ctrl-G significantly improves consistency, but introduces a trade-off against generation quality.} }
    \label{fig:ctrl-g}
    \vspace{-3mm}
\end{figure}
\cref{fig:ctrl-g} shows how image generation quality and control consistency vary with both CFG ($\gamma_y$) and ctrl-G ($\gamma_c$). Compared to the default settings used in \cref{fig:baseline_results}, we find that increasing  CFG generally improves both IS and FID (although too high a scale may hurt FID), but slightly decreases consistency. Interestingly, in the case of VAR, higher CFG \textit{improves} FID over the no-control model ($\sim5.7$ to $\sim 5.1$), demonstrating that providing a weaker model with spatial cues can boost generation quality. On the other hand, ctrl-G is able to drastically improve consistency, but this is traded off against worse generation quality. This is visualised in the left of \cref{fig:vis_sampling_params}, where we see that the canny edge map is better conformed to, but at the cost of over-emphasised edges and incongruous generation for larger ctrl-G. Intuitively, ctrl-G only targets the probability of the \textit{control} given the generation $p(\b c|\b z)$ (\cref{eq:ctrl-g}), which may not align with image \textit{quality}. This trade-off aligns with the results reported for image \textit{editing} in \citep{pix2pix}. We note that the F1 values of $\sim$50 for SiT are \textit{significantly higher} than the values commonly found in the literature \citep{controlar,omnigen}, demonstrating the large effect that this \textit{generation-time} adjustment has on consistency. 
Additionally, ctrl-G incurs the extra inference cost of needing to run the no-control model $p_{\b \theta}(\b z)$, reducing latency and throughput:\footnote{We use the pretrained ImageNet model for $p_{\b \theta}(\b z)$ \textit{in series} with the control-prefill model. Better latency and parameter efficiency could be achieved by training the control-prefill model with empty control $\b c=\varnothing$ and parallelising like CFG, although the FLOPs increase would still remain.}
\begin{center}
\footnotesize
\begin{tabular}{@{}lllll@{}}
\toprule
 & VAR-d16 prefill + CFG & VAR-d16 prefill + CFG + ctrl-G & SiT-XL/2 prefill + CFG & SiT-XL/2 prefill + CFG + ctrl-G \\ \midrule
\#params                   & 310M & 310M (+ 310M) & 675M & 675M (+ 675M) \\
lat$\downarrow$ (s) bs=1   &   0.31   &  0.55             &  3.4    &     5.9          \\
TP$\uparrow$ (img/s) bs=16 &   12   &   8.4            &  0.35    &     0.24          \\ \bottomrule
\end{tabular}%
\end{center}
\begin{takeawaybox}
    Control guidance greatly improves control consistency, but there is a trade-off against image quality and inference cost. CFG has a small negative effect on consistency but generally improves generation quality.
\end{takeawaybox}
Finally we note that in ControlVAR \citep{controlvar}, when performing guidance (with class and control conditioning), all guidance scales $\gamma$ are set to be equal. However, \cref{fig:ctrl-g} shows that this setting may lead to excessive degradation in generation quality, demonstrating the importance of \textit{separately tuning} $\gamma_y$ (CFG) and $\gamma_c$ (ctrl-G).

\subsection{Sampling Distribution Truncation}

Another intuitive, but unexplored way of potentially improving control consistency, is to directly truncate/concentrate the sampling distribution. This intuitively assumes that the high density/probablity regions predicted by the model $p_{\theta}(\b z|\b y, \b c)$ will most conform to the input spatial control. We select one method for each of VAR and SiT.

\paragraph{Top-$\b p$ sampling for VAR.} As VAR samples from a \textit{softmax} output $\b \pi$, there are a number of options to concentrate probability mass (temperature, top-$p$, top-$k$). In this work, we focus on top-$p$ sampling \citep{topp}, which has been shown to effectively improve the coherence of language generation. It directly truncates the softmax distribution $\b \pi$,
\begin{equation}
\pi^{\text{top-}p}_k =
\frac{\pi_k}{\sum_{i\in V_p}\pi_i} \text{ if } k \in V_p \text{,~~otherwise } 0
\quad \quad 
\text{where } V_p = \text{the smallest set of indices } V \text{ s.t. } \sum\nolimits_{i \in V} \pi_i \ge p ~,
\end{equation}
effectively discarding the tail $1-p$ of the softmax. Intuitively, one would expect the most probable vector codes, predicted by the model's softmax at each token location, to best conform to the spatial control condition.
\paragraph{Score temperature scaling for diffusion/flow models.} In the case of diffusion, temperature scaling the score \citep{ADM,score_anneal,autoguidance}, can be intuited as raising the marginal distributions to power $1/\tau$, also focusing sampling on higher density regions,
\begin{equation}
    \frac{1}{\tau_t}\nabla_{\b z_t}\log\; p(\b z_t) = \nabla_{\b z_t}\log\; \left[p(\b z_t)^{\frac{1}{\tau_t}}\right], \quad \text{ where }\quad     1/\tau_t = \frac{(\alpha_t^2+\sigma_t^2)\frac{1}{\tau_0}}{\alpha^2 + \frac{\sigma^2}{\tau_0}}.
\end{equation}
We adopt the annealing schedule from \citet{score_anneal} for $1/\tau_t$ that samples from the temperature-scaled data distribution \textit{assuming it is Gaussian}, where $\tau_0$ is the temperature at $t=0$ (clean data), labelled ``temp'' in our experiments. This choice is motivated by the poor generation results from \citet{ADM} for constant $\tau$.
\begin{figure}[t]
 \vspace{-4mm}
    \centering
    \includegraphics[width=.49\linewidth]{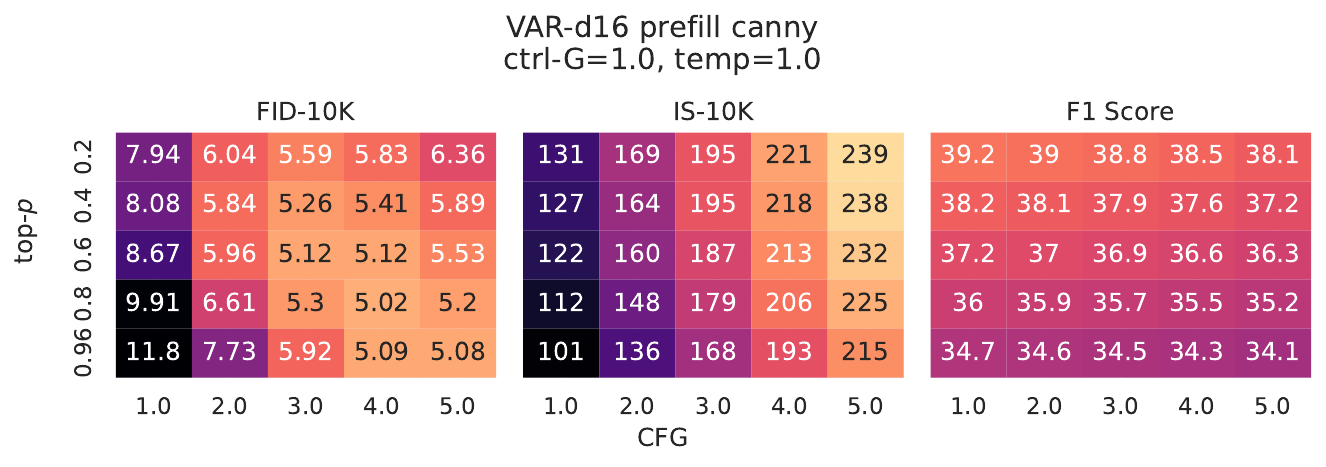}
    \includegraphics[width=.49\linewidth]{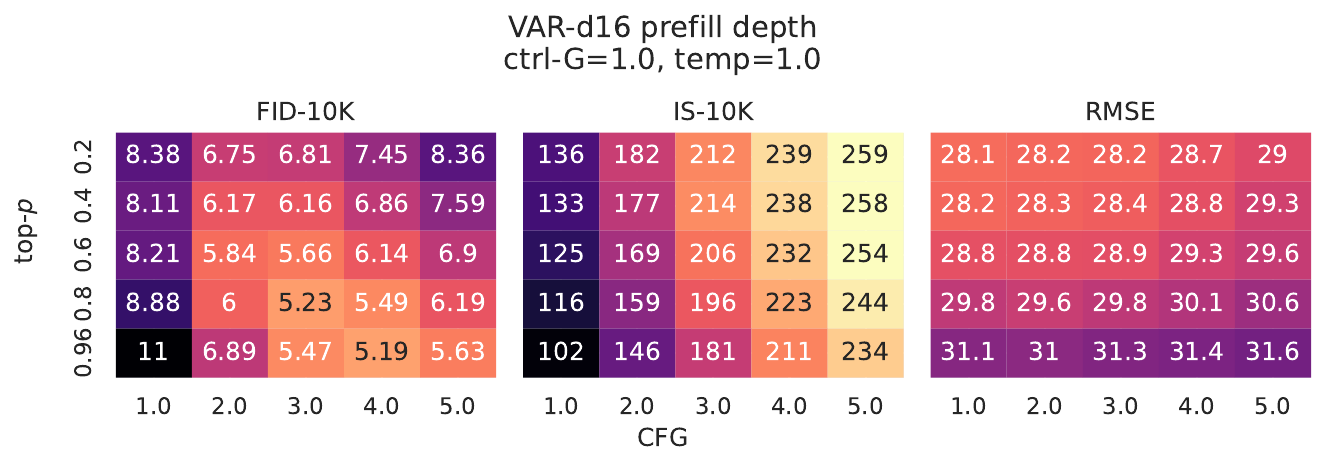}

    \includegraphics[width=.49\linewidth]{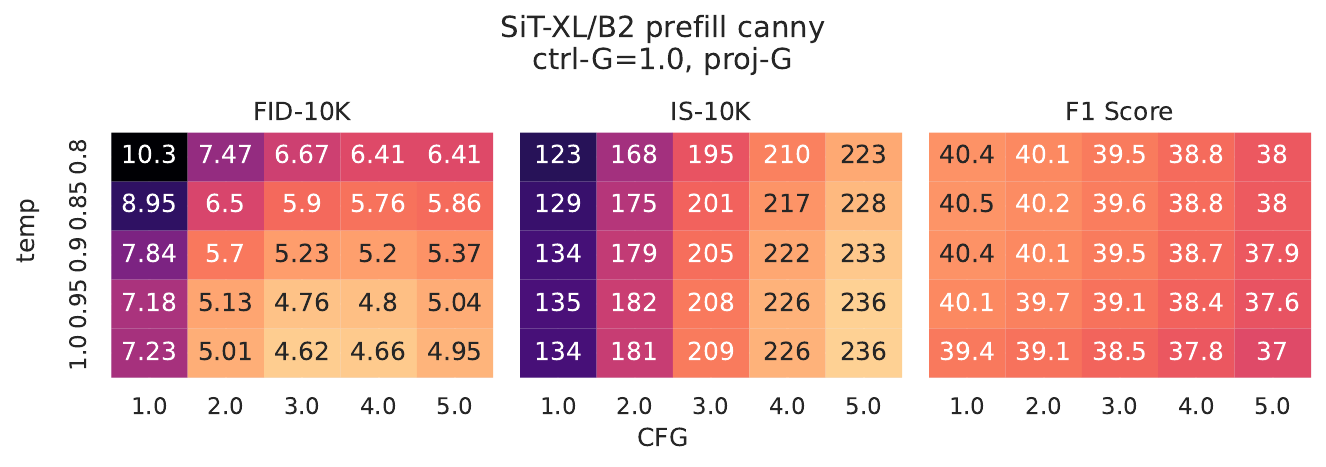}
    \includegraphics[width=.49\linewidth]{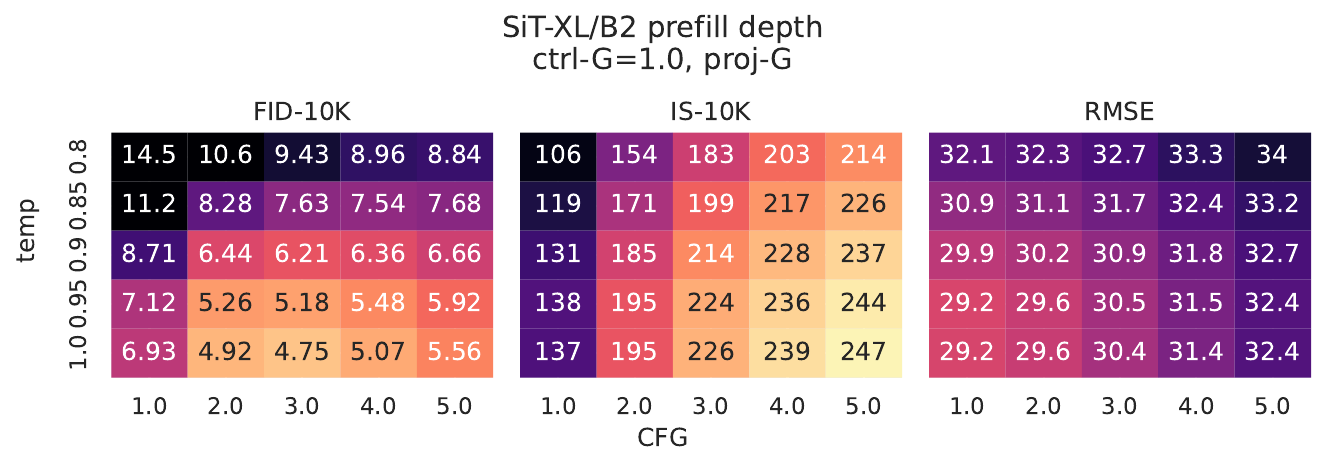}
    \vspace{-3mm}\caption{Effect of CFG and distribution truncation on conditional generation. Brighter means better. Again, CFG generally improves generation quality, with a slight decrease in consistency. \textbf{Top-$\b p$ softmax truncation improves both generation quality and consistency for VAR}, although aggressive truncation may reduce diversity and hurt FID. Score temperature scaling does not produce any meaningful benefit for SiT.}
    \label{fig:trunc}
    \vspace{-3mm}
\end{figure}
\paragraph{Results.}
\cref{fig:vis_sampling_params,fig:trunc,fig:trunc-ctrlg} show how top-$p$ and temperature scaling affect generation quality and control consistency alongside CFG. We see that CFG behaves similarly to \cref{fig:ctrl-g}, generally improving generation quality and slightly hurting consistency. Notably, truncation via top-$p$ is able to improve \textit{both} generation quality and control consistency (although excessive truncation seems to hurt FID likely due to reduced diversity). \cref{fig:trunc-ctrlg} (\cref{app:exp}) further shows that the consistency improvements can be \textit{stacked} with ctrl-G. This demonstrates that top-$p$ is an important parameter to adjust for softmax-based generative models, providing improvements in all fronts \textit{without any additional cost}. In contrast, temperature scaling is unable to meaningfully improve consistency, leading to a small increase for canny, and no improvement for depth, whilst simultaneously hurting generation quality. Compared to the results for ctrl-G, this suggests that naively scaling the score (without changing its direction) is insufficient to improve consistency. One possible intuition, is that score temperature scaling occurs \textit{uniformly} over the spatial dimensions, \ie it is a global transformation, and so it may have limited \textit{spatial/local} influence, whilst ctrl-G varies in space. On the other hand, top-$p$ explicitly truncates at each token \textit{location}, allowing it to influence spatial consistency.

\begin{takeawaybox}
    Distributional truncation of the softmax (top-$p$ sampling), can improve both generation quality and control consistency for VAR at inference time for no extra cost. Conversely, truncation via temperature scaling of the score does not meaningfully benefit the consistency or generation quality of diffusion/flow models.
\end{takeawaybox}
We finally note that \cref{fig:ctrl-g,fig:trunc,fig:trunc-ctrlg} demonstrate that \textit{meaningfully improved performance} can be easily obtained by the sampling adjustments explore in this section (compared to the default settings used in \cref{fig:baseline_results}). 
\begin{takeawaybox}
    Classifier-free guidance, control guidance and softmax truncation offer meaningful additional ``knobs'' that practictioners can adjust to improve the characteristics of their generations \textit{at sampling time}.
\end{takeawaybox}
\section{Adapters for Spatial Control}\label{sec:adapters}
\begin{figure}
\vspace{-4mm}
    \centering
    \includegraphics[width=\linewidth]{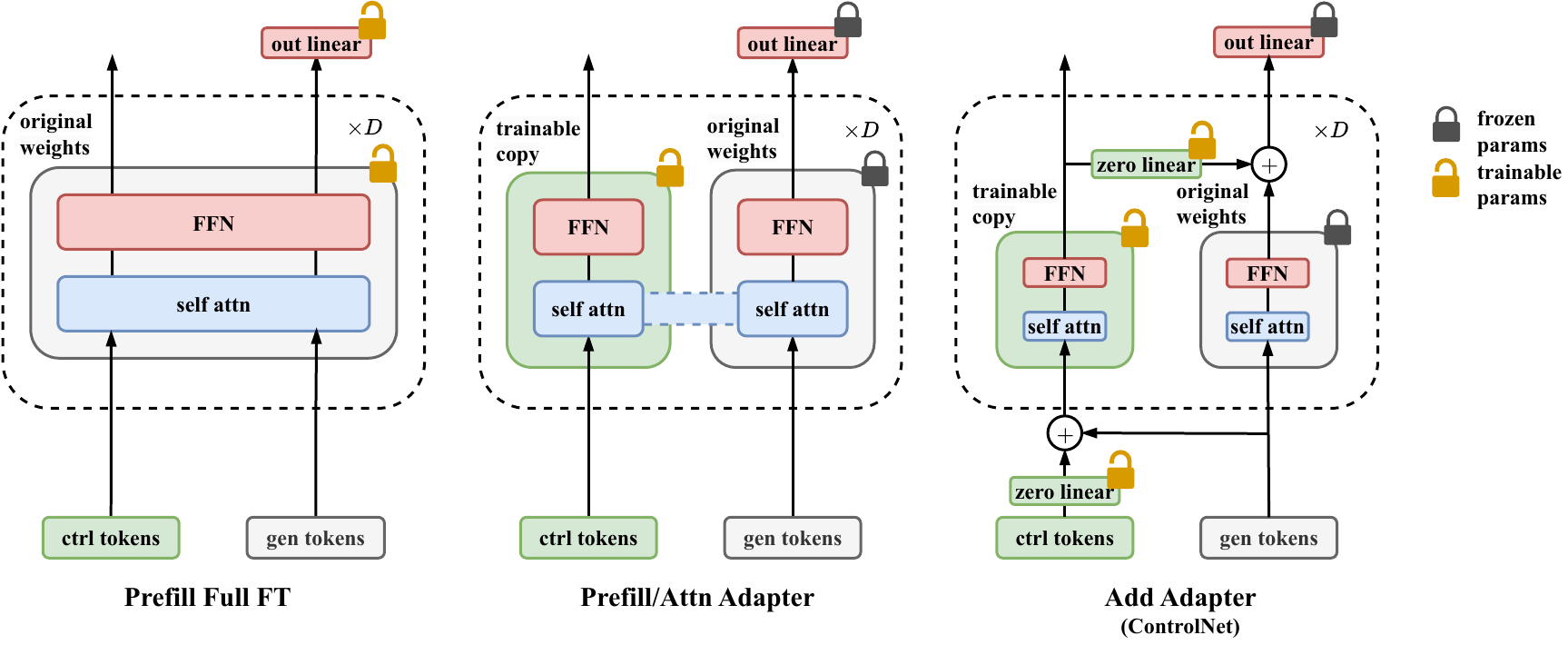}
    \vspace{-7mm}\caption{Illustration of adapter approaches considered in this work. Compared to full finetuning with prefill (left), adapters (centre, right) freeze the original weights, using a trainable copy to process control tokens and communicate spatial conditioning to generation tokens at each layer. The prefill/attention adapter communicates flexibly via sharing key-value pairs with the generation tokens, whilst the addition adapter directly injects information via zero-initialised linear layers \citep{controlnet}, requiring each generation token to be spatially matched to a control token.}
    \vspace{-2mm}
    \label{fig:adapter-illust}
\end{figure}

In the previous sections, we did not consider adapter-style approaches, typified by ControlNet \citep{controlnet}, where a pretrained image generation model's weights are \textit{frozen} and an additional module is trained for spatial control. In this section, we aim to clarify and empirically demonstrate to readers the motivation for adapters, positioning them relative to approaches where all parameters are trained.
Adapter-based approaches are motivated by the problem setting of \textit{adding} spatial control to existing powerful pre-trained foundation models such as Stable Diffusion \citep{LDM,sdxl,sd3} and Flux \citep{flux2024} whilst mitigating the potential that fine-tuning on limited/undiverse downstream control would cause degradation in general generation ability and ``forgetting'' of concepts in the pretraining data \citep{controlnet}. However, we highlight that this problem setting may not apply to all practitioners. 
For example, an adapter-based approach may not actually be the most suitable if the aim is to endow a single generative foundation model with many abilities from the ground up \citep{chen2024UniReal,omnigen,zhang2025unified}, or if fine-tuning data is rich and diverse enough that forgetting is not a concern.
We note that this nuance may become lost in the literature, \eg \citet{controlar} benchmark their approach, where all parameters are trained, against adapters without discussing the above.
The following experiments aim to highlight the differences between training all parameters versus using an adapter, so that practitioners can have a clearer view of which style of approach is more suitable for their use-case.
\vspace{-2mm}\subsection{Adapter Architectures}
We consider two different adapter architectures, that supply spatial control information to the base model via two different mechanisms. We note that in both cases, we train copies of the base model's transformer blocks, only fine-tuning the FFN, self-attention modules and learned positional embeddings, leaving adaptive layer norm frozen.

\textbf{Add adapter.} Illustrated in the right of \cref{fig:adapter-illust}, this approach adapts the ControlNet \citep{controlnet} architecture to transformers, like in \cite{chen2024pixartdelta}, training a copy of the base model to additively inject spatial conditioning information at each layer, via zero-initialised linear layers, whilst the original weights are performing generation. Inference for each generation token is \textit{matched} by inference on a corresponding control token.

\textbf{Attention/prefill adapter.} Illustrated in the centre of \cref{fig:adapter-illust}, this approach retains the prefill paradigm of conditioning from the baseline, and trains a separate copy of the transformer blocks to be used during condition prefilling, leaving generation to the original model weights. Although the adapter and base model have different attention weight matrices, generation tokens are able to attend back to the keys and values produced by the control adapter.

These two approaches aim to broadly represent two distinct potential options for adapter-based approaches to spatial conditioning, i.e., attention and additive injection. Our experiments are of course not exhaustive, and there is an exponentially large space of specific design choices that could be explored: selective freezing, the use of low-rank adaptation (LoRA) \citep{lora}, the depth of the adapter, the location and direction of the exchange of conditioning information \etc. However, we hope that our experiments can still provide practical high-level research takeaways to the community.
\subsection{Results}
\begin{figure}[t]
 \vspace{-4mm}
\centering
    \begin{minipage}{0.56\textwidth}
        \includegraphics[width=\textwidth]{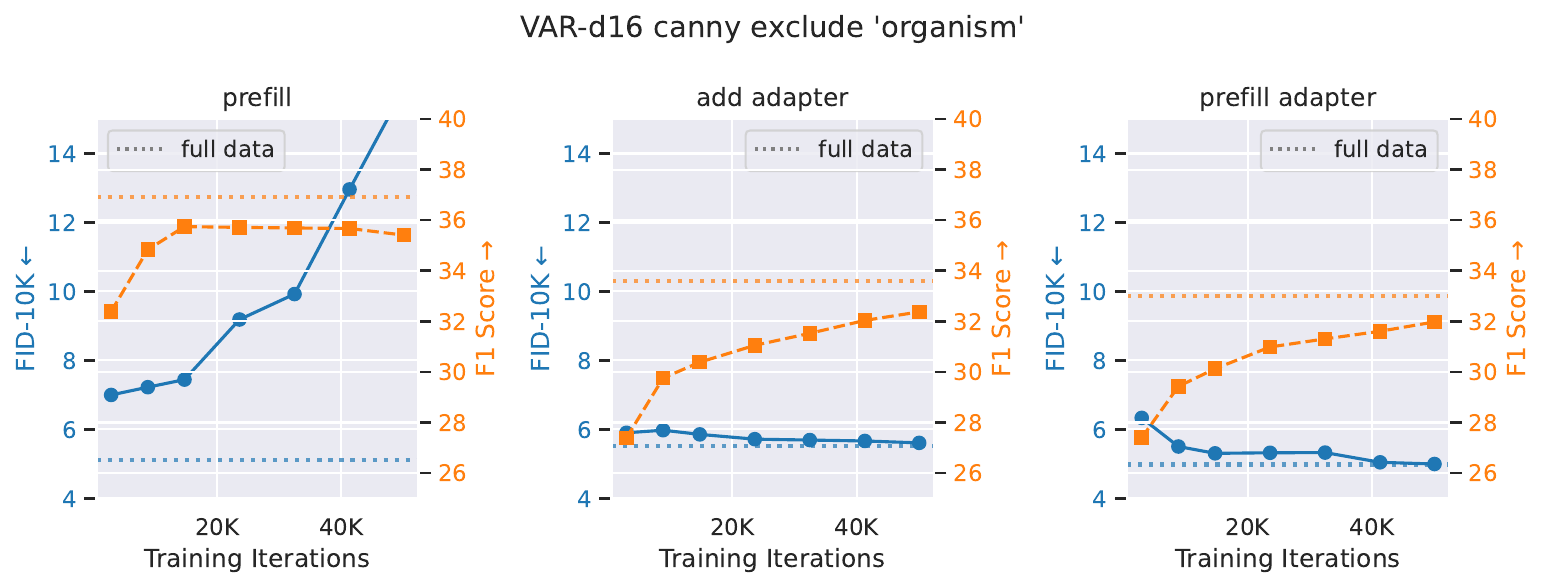}

        \includegraphics[width=\textwidth]{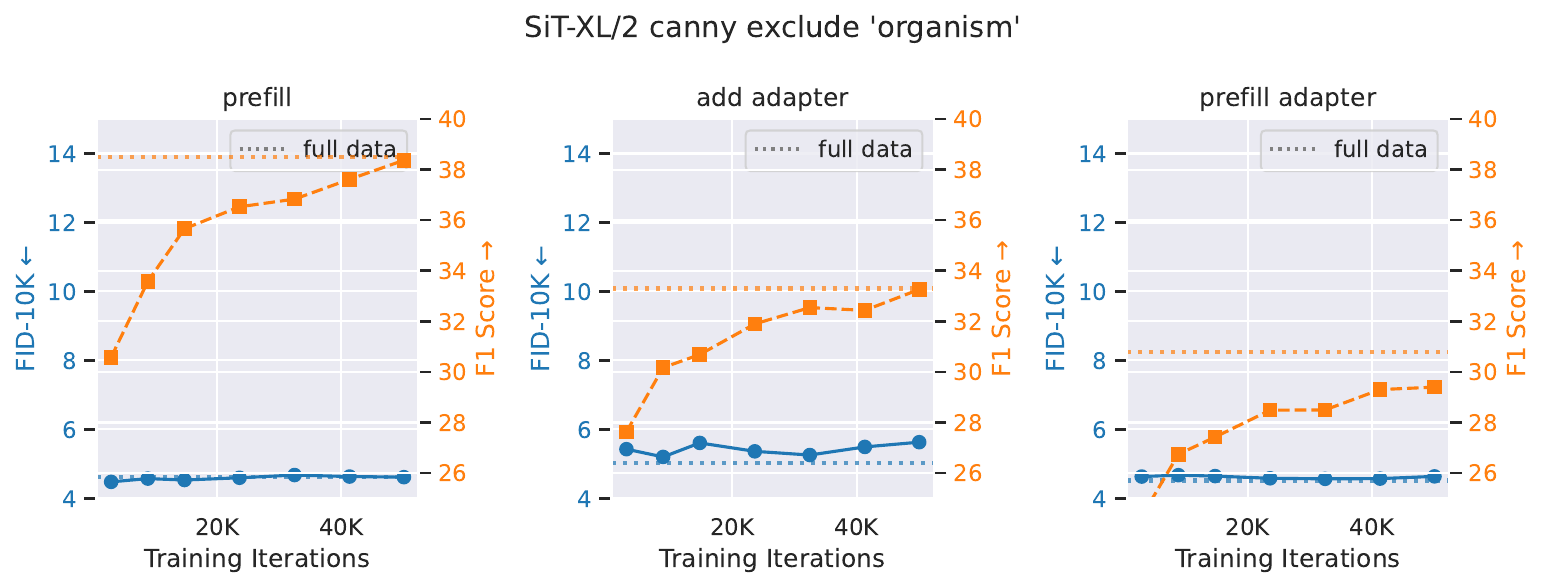}
    \end{minipage}
    \hspace{-2mm}
    \begin{minipage}{0.44\textwidth}
        \includegraphics[width=\linewidth]{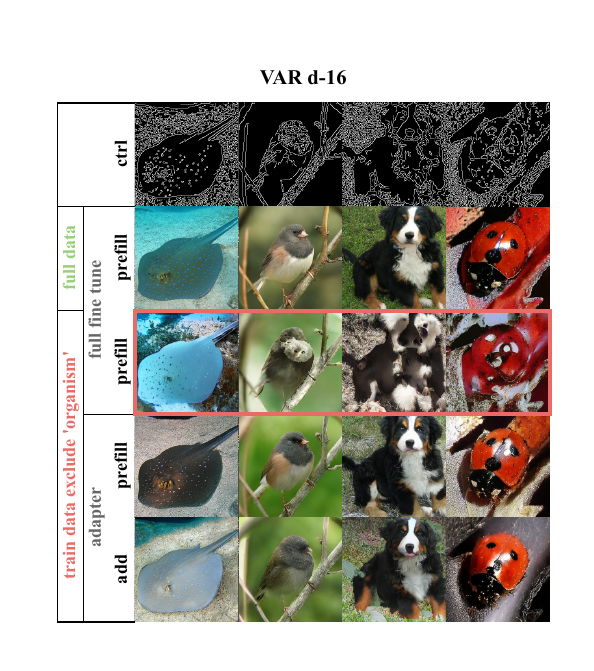}
    \end{minipage}
    \vspace{-3mm}\caption{\textbf{Left:} Generation quality (FID$\downarrow$) and control consistency (F1$\uparrow$) of Canny-conditioned VAR trained on control data that excludes classes that are ``organisms'' according to WordNet, evaluated for the \textit{whole} ImageNet distribution. Fully finetuning VAR (prefill) leads to a collapse in generation quality, whilst using adapters mitigates this. On the other hand, the larger SiT model manages to maintain generation quality over the whole distribution. See \cref{fig:forgor2} in \cref{app:exp} for depth results. \textbf{Right:} Visualisation of generations -- full finetuning exhibits catastrophic forgetting, whilst adapters are able to generalise control-conditioned generation to ``organism'' classes. }
    \label{fig:forgor}
\end{figure}

Based on the exploration of the previous section, we select a set of sampling parameters that aim to balance generation quality with control consistency for our remaining experiments. They are specified in \cref{tab:adapter,fig:forgor}. Additional generation examples for the following experiments can be found in \cref{app:example-gens}.

\begin{table}[]
\begin{minipage}{0.5\textwidth}

\colorlet{CompareArrow}{Dandelion!90!black}
\resizebox{\textwidth}{!}{%
\setlength{\extrarowheight}{0pt}
\renewcommand{\arraystretch}{1.1}
\begin{tabular}{@{}lllllllllll@{}}
\toprule
\multicolumn{2}{l}{\multirow{4}{*}{\begin{tabular}[c]{@{}l@{}}model\\ \scriptsize(sampling params)\end{tabular}}} &
\multirow{4}{*}{method} &
\multirow{4}{*}{ctrl} &
\multirow{4}{*}{\begin{tabular}[c]{@{}l@{}}\#params\\ base\\ +adapter\end{tabular}} &
\multicolumn{2}{c}{gen. quality} &
\multicolumn{2}{c}{ctrl consistency} &
\multicolumn{2}{c}{inf. cost} \\
\cmidrule(lr){6-7} \cmidrule(lr){8-9} \cmidrule(lr){10-11}
& & & & &
\begin{tabular}[c]{@{}l@{}}FID$\downarrow$\\ 10K\end{tabular} &
\begin{tabular}[c]{@{}l@{}}IS$\uparrow$\\ 10K\end{tabular} &
F1$\uparrow$ &
RMSE$\downarrow$ &
\begin{tabular}[c]{@{}l@{}}lat$\downarrow$\\(s)\\bs=1\end{tabular} &
\begin{tabular}[c]{@{}l@{}}TP$\uparrow$\\(img/s)\\bs=16\end{tabular} \\
\midrule
\multicolumn{2}{l}{{\color[HTML]{656565} \begin{tabular}[c]{@{}l@{}}ImageNet-val \\ (ctrl source)\end{tabular}}} &
  {\color[HTML]{656565} } &
  {\color[HTML]{656565} } &
  {\color[HTML]{656565} } &
  {\color[HTML]{656565} 4.44} &
  {\color[HTML]{656565} 187} &
  {\color[HTML]{656565} } &
  {\color[HTML]{656565} } &
  {\color[HTML]{656565} } &
  {\color[HTML]{656565} } \\ \midrule
\multicolumn{3}{l}{{\color[HTML]{656565} \begin{tabular}[c]{@{}l@{}}VAR-d16 CFG=2.5, temp=1, \\ top-$p$=0.96, top-$k$=900\end{tabular}}} &
  {\color[HTML]{656565} none} &
  {\color[HTML]{656565} 310M} &
  {\color[HTML]{656565} 5.71} &
  {\color[HTML]{656565} 214} &
  {\color[HTML]{656565} } &
  {\color[HTML]{656565} } &
  {\color[HTML]{656565} 0.26} &
  {\color[HTML]{656565} 17} \\ \midrule
 &
   &
  prefill &
  canny &
  310M &
  5.12 &
  187 &
  36.9\tikzmark{ct}&
   &
   &
   \\
 &
   &
   &
  depth &
   &
  5.66 &
  207\tikzmark{at}&
   &
  28.9 &
  \multirow{-2}{*}{0.31} &
  \multirow{-2}{*}{12} \\ \cmidrule(l){3-11} 
 &
   &
  {\color[HTML]{030191} } &
  {\color[HTML]{030191} canny} &
  {\color[HTML]{030191} } &
  {\color[HTML]{030191} 5.51} &
  {\color[HTML]{030191} 192} &
  {\color[HTML]{030191} 33.6\tikzmark{bt}} &
  {\color[HTML]{030191} } &
  {\color[HTML]{030191} } &
  {\color[HTML]{030191} } \\
 &
   &
  \multirow{-2}{*}{{\color[HTML]{030191} \begin{tabular}[c]{@{}l@{}}add adapter\\ (CtrlNet)\end{tabular}}} &
  {\color[HTML]{030191} depth} &
  \multirow{-2}{*}{{\color[HTML]{030191} \begin{tabular}[c]{@{}l@{}}310M\\ + 220M\tikzmark{et}\end{tabular}}} &
  {\color[HTML]{030191} 6.08} &
  {\color[HTML]{030191} 217} &
  {\color[HTML]{030191} } &
  {\color[HTML]{030191} 31.1} &
  \multirow{-2}{*}{{\color[HTML]{030191} 0.55\tikzmark{dt}}} &
  \multirow{-2}{*}{{\color[HTML]{030191} 7.9}} \\ \cmidrule(l){3-11} 
 &
   &
  {\color[HTML]{030191} } &
  {\color[HTML]{030191} canny} &
  {\color[HTML]{030191} } &
  {\color[HTML]{030191} 4.98} &
  {\color[HTML]{030191} 201\tikzmark{ab}} &
  {\color[HTML]{030191} 33.0\tikzmark{bb}} &
  {\color[HTML]{030191} } &
  {\color[HTML]{030191} } &
  {\color[HTML]{030191} } \\
 &
  \multirow{-6}{*}{\begin{tabular}[c]{@{}l@{}}ctrl-G\\ =1.0\end{tabular}} &
  \multirow{-2}{*}{{\color[HTML]{030191} \begin{tabular}[c]{@{}l@{}}prefill\\ adapter\end{tabular}}} &
  {\color[HTML]{030191} depth} &
  \multirow{-2}{*}{{\color[HTML]{030191} \begin{tabular}[c]{@{}l@{}}310M\\ + 202M\end{tabular}}} &
  {\color[HTML]{030191} 5.64} &
  {\color[HTML]{030191} 208} &
  {\color[HTML]{030191} } &
  {\color[HTML]{030191} 29.7} &
  \multirow{-2}{*}{{\color[HTML]{030191} 0.31\tikzmark{db}}} &
  \multirow{-2}{*}{{\color[HTML]{030191} 12}} \\ \cmidrule(l){2-11} 
 &
   &
   &
  canny &
   &
  5.40 &
  180 &
  39.9 &
   &
   &
   \\
 &
   &
  \multirow{-2}{*}{prefill} &
  depth &
  \multirow{-2}{*}{310M$\times 2$} &
  5.59 &
  199 &
   &
  27.4 &
  \multirow{-2}{*}{0.55} &
  \multirow{-2}{*}{8.4} \\ \cmidrule(l){3-11} 
 &
   &
  {\color[HTML]{030191} } &
  {\color[HTML]{030191} canny} &
  {\color[HTML]{030191} } &
  {\color[HTML]{030191} 6.10} &
  {\color[HTML]{030191} 174} &
  {\color[HTML]{030191} 37.3} &
  {\color[HTML]{030191} } &
  {\color[HTML]{030191} } &
  {\color[HTML]{030191} } \\
 &
   &
  \multirow{-2}{*}{{\color[HTML]{030191} \begin{tabular}[c]{@{}l@{}}add adapter\\ (CtrlNet)\end{tabular}}} &
  {\color[HTML]{030191} depth} &
  \multirow{-2}{*}{{\color[HTML]{030191} \begin{tabular}[c]{@{}l@{}}310M\\ + 220M\tikzmark{eb}\end{tabular}}} &
  {\color[HTML]{030191} 5.86} &
  {\color[HTML]{030191} 207} &
  {\color[HTML]{030191} } &
  {\color[HTML]{030191} 29.3} &
  \multirow{-2}{*}{{\color[HTML]{030191} 0.80}} &
  \multirow{-2}{*}{{\color[HTML]{030191} 6.3}} \\ \cmidrule(l){3-11} 
 &
   &
  {\color[HTML]{030191} } &
  {\color[HTML]{030191} canny} &
  {\color[HTML]{030191} } &
  {\color[HTML]{030191} 5.15} &
  {\color[HTML]{030191} 190} &
  {\color[HTML]{030191} 35.9\tikzmark{cb}} &
  {\color[HTML]{030191} } &
  {\color[HTML]{030191} } &
  {\color[HTML]{030191} } \\
\multirow{-12}{*}{\begin{tabular}[c]{@{}l@{}}VAR-d16\\\scriptsize  CFG=3.0, \\ \scriptsize temp=1, \\ \scriptsize top-$p$=0.6, \\ \scriptsize top-$k$=900\end{tabular}} &
  \multirow{-6}{*}{\begin{tabular}[c]{@{}l@{}}ctrl-G\\ =1.5\end{tabular}} &
  \multirow{-2}{*}{{\color[HTML]{030191} \begin{tabular}[c]{@{}l@{}}prefill\\ adapter\end{tabular}}} &
  {\color[HTML]{030191} depth} &
  \multirow{-2}{*}{{\color[HTML]{030191} \begin{tabular}[c]{@{}l@{}}310M\\ + 202M\end{tabular}}} &
  {\color[HTML]{030191} 5.47} &
  {\color[HTML]{030191} 201} &
  {\color[HTML]{030191} } &
  {\color[HTML]{030191} 27.9} &
  \multirow{-2}{*}{{\color[HTML]{030191} 0.56}} &
  \multirow{-2}{*}{{\color[HTML]{030191} 8.5}} \\ \bottomrule
\end{tabular}

\begin{tikzpicture}[remember picture,overlay]
  \tikzstyle{CompareLine}=[draw=CompareArrow,line width=2pt,<->] 
  \tikzstyle{CompareLabel}=[font=\bfseries, text=CompareArrow, anchor=west]

\newcommand{\CompareV}[6][]{%
  \draw[CompareLine,#1] 
    ($ (pic cs:#2) + (#5,#6) $) -- ($ (pic cs:#3) + (#5,0) $);
  \node[CompareLabel,#1] 
    at ($($(pic cs:#2)+(#5,#6)$)!0.5!($(pic cs:#3)+(#5,0)$) + (-0.2em,0)$) {#4};
}

\CompareV[]{at}{ab}{(a)}{0.5ex}{1.2ex} 
\CompareV[]{bt}{bb}{(b)}{1ex}{0.3ex}
\CompareV[]{ct}{cb}{(c)}{8ex}{1.2ex}
  \CompareV[]{dt}{db}{(d)}{1ex}{3ex}
  \CompareV[]{et}{eb}{(e)}{1.5ex}{2ex}
\end{tikzpicture}

}
\end{minipage}
\hfill
\begin{minipage}{0.5\textwidth}
\colorlet{CompareArrow}{Dandelion!90!black}
\setlength{\extrarowheight}{0pt}
\renewcommand{\arraystretch}{1.08}
\resizebox{\textwidth}{!}{%

\begin{tabular}{@{}lllllllllll@{}}
\toprule
\multicolumn{2}{l}{\multirow{4}{*}{\begin{tabular}[c]{@{}l@{}}model\\ \scriptsize(sampling params)\end{tabular}}} &
\multirow{4}{*}{method} &
\multirow{4}{*}{ctrl} &
\multirow{4}{*}{\begin{tabular}[c]{@{}l@{}}\#params\\ base\\ +adapter\end{tabular}} &
\multicolumn{2}{c}{gen. quality} &
\multicolumn{2}{c}{ctrl consistency} &
\multicolumn{2}{c}{inf. cost} \\
\cmidrule(lr){6-7} \cmidrule(lr){8-9} \cmidrule(lr){10-11}
& & & & &
\begin{tabular}[c]{@{}l@{}}FID$\downarrow$\\ 10K\end{tabular} &
\begin{tabular}[c]{@{}l@{}}IS$\uparrow$\\ 10K\end{tabular} &
F1$\uparrow$ &
RMSE$\downarrow$ &
\begin{tabular}[c]{@{}l@{}}lat$\downarrow$\\(s)\\bs=1\end{tabular} &
\begin{tabular}[c]{@{}l@{}}TP$\uparrow$\\(img/s)\\bs=16\end{tabular} \\
\midrule
\multicolumn{2}{l}{{\color[HTML]{656565} \begin{tabular}[c]{@{}l@{}}ImageNet-val \\ (ctrl source)\end{tabular}}} &
  {\color[HTML]{656565} } &
  {\color[HTML]{656565} } &
  {\color[HTML]{656565} } &
  {\color[HTML]{656565} 4.44} &
  {\color[HTML]{656565} 187} &
  {\color[HTML]{656565} } &
  {\color[HTML]{656565} } &
  {\color[HTML]{656565} } &
  {\color[HTML]{656565} } \\ \midrule
\multicolumn{3}{l}{{\color[HTML]{656565} \begin{tabular}[c]{@{}l@{}}SiT-XL/2 CFG=1.5, \\ Euler ODE, steps=64\end{tabular}}} &
  {\color[HTML]{656565} none} &
  {\color[HTML]{656565} } &
  {\color[HTML]{656565} 4.47} &
  {\color[HTML]{656565} 228} &
  {\color[HTML]{656565} } &
  {\color[HTML]{656565} } &
  {\color[HTML]{656565} 3.0} &
  {\color[HTML]{656565} 0.39} \\ \midrule
 &
   &
  prefill &
  canny &
   &
  4.62 &
  209 &
  38.5\tikzmark{ct2} &
   &
   &
   \\
 &
   &
   &
  depth &
  \multirow{-2}{*}{675M} &
  4.75 &
  226\tikzmark{at2}&
   &
  30.4 &
  \multirow{-2}{*}{3.4} &
  \multirow{-2}{*}{0.35} \\ \cmidrule(l){3-11} 
 &
   &
  {\color[HTML]{030191} } &
  {\color[HTML]{030191} canny} &
  {\color[HTML]{030191} } &
  {\color[HTML]{030191} 5.02} &
  {\color[HTML]{030191} 226} &
  {\color[HTML]{030191} 33.3\tikzmark{bt2}} &
  {\color[HTML]{030191} } &
  {\color[HTML]{030191} } &
  {\color[HTML]{030191} } \\
 &
   &
  \multirow{-2}{*}{{\color[HTML]{030191} \begin{tabular}[c]{@{}l@{}}add adapter\\ (CtrlNet)\end{tabular}}} &
  {\color[HTML]{030191} depth} &
  \multirow{-2}{*}{{\color[HTML]{030191} \begin{tabular}[c]{@{}l@{}}675M\\ + 485M\tikzmark{eb2}\end{tabular}}} &
  {\color[HTML]{030191} 4.84} &
  {\color[HTML]{030191} 241} &
  {\color[HTML]{030191} } &
  {\color[HTML]{030191} 33.5} &
  \multirow{-2}{*}{{\color[HTML]{030191} 6.2\tikzmark{dt2}}} &
  \multirow{-2}{*}{{\color[HTML]{030191} 0.19}} \\ \cmidrule(l){3-11} 
 &
   &
  {\color[HTML]{030191} } &
  {\color[HTML]{030191} canny} &
  {\color[HTML]{030191} } &
  {\color[HTML]{030191} 4.52} &
  {\color[HTML]{030191} 235\tikzmark{ab2}} &
  {\color[HTML]{030191} 30.8\tikzmark{bb2}} &
  {\color[HTML]{030191} } &
  {\color[HTML]{030191} } &
  {\color[HTML]{030191} } \\
 &
  \multirow{-6}{*}{\begin{tabular}[c]{@{}l@{}}ctrl-G\\ =1.0\end{tabular}} &
  \multirow{-2}{*}{{\color[HTML]{030191} \begin{tabular}[c]{@{}l@{}}prefill\\ adapter\end{tabular}}} &
  {\color[HTML]{030191} depth} &
  \multirow{-2}{*}{{\color[HTML]{030191} \begin{tabular}[c]{@{}l@{}}675M\\ + 446M\end{tabular}}} &
  {\color[HTML]{030191} 4.62} &
  {\color[HTML]{030191} 251} &
  {\color[HTML]{030191} } &
  {\color[HTML]{030191} 32.7} &
  \multirow{-2}{*}{{\color[HTML]{030191} 3.4\tikzmark{db2}}} &
  \multirow{-2}{*}{{\color[HTML]{030191} 0.35}} \\ \cmidrule(l){2-11} 
 &
   &
   &
  canny &
   &
  5.75 &
  194 &
  44.8 &
   &
   &
   \\
 &
   &
  \multirow{-2}{*}{prefill} &
  depth &
  \multirow{-2}{*}{675M$\times 2$} &
  5.34 &
  215 &
   &
  27.9 &
  \multirow{-2}{*}{5.9} &
  \multirow{-2}{*}{0.24} \\ \cmidrule(l){3-11} 
 &
   &
  {\color[HTML]{030191} } &
  {\color[HTML]{030191} canny} &
  {\color[HTML]{030191} } &
  {\color[HTML]{030191} 6.00} &
  {\color[HTML]{030191} 210} &
  {\color[HTML]{030191} 40.5} &
  {\color[HTML]{030191} } &
  {\color[HTML]{030191} } &
  {\color[HTML]{030191} } \\
 &
   &
  \multirow{-2}{*}{{\color[HTML]{030191} \begin{tabular}[c]{@{}l@{}}add adapter\\ (CtrlNet)\end{tabular}}} &
  {\color[HTML]{030191} depth} &
  \multirow{-2}{*}{{\color[HTML]{030191} \begin{tabular}[c]{@{}l@{}}675M\\ + 485M\tikzmark{et2}\end{tabular}}} &
  {\color[HTML]{030191} 5.51} &
  {\color[HTML]{030191} 228} &
  {\color[HTML]{030191} } &
  {\color[HTML]{030191} 30.5} &
  \multirow{-2}{*}{{\color[HTML]{030191} 8.1}} &
  \multirow{-2}{*}{{\color[HTML]{030191} 0.15}} \\ \cmidrule(l){3-11} 
 &
   &
  {\color[HTML]{030191} } &
  {\color[HTML]{030191} canny} &
  {\color[HTML]{030191} } &
  {\color[HTML]{030191} 5.39} &
  {\color[HTML]{030191} 218} &
  {\color[HTML]{030191} 37.1\tikzmark{cb2}} &
  {\color[HTML]{030191} } &
  {\color[HTML]{030191} } &
  {\color[HTML]{030191} } \\
\multirow{-12}{*}{\begin{tabular}[c]{@{}l@{}}SiT-XL/2\\ \scriptsize CFG=3.0, \\ \scriptsize Euler \\\scriptsize  -ODE,\\ \scriptsize steps=64,\\ \scriptsize proj-G\end{tabular}} &
  \multirow{-6}{*}{\begin{tabular}[c]{@{}l@{}}ctrl-G\\ =1.5\end{tabular}} &
  \multirow{-2}{*}{{\color[HTML]{030191} \begin{tabular}[c]{@{}l@{}}prefill\\ adapter\end{tabular}}} &
  {\color[HTML]{030191} depth} &
  \multirow{-2}{*}{{\color[HTML]{030191} \begin{tabular}[c]{@{}l@{}}675M\\ + 446M\end{tabular}}} &
  {\color[HTML]{030191} 5.17} &
  {\color[HTML]{030191} 227} &
  {\color[HTML]{030191} } &
  {\color[HTML]{030191} 29.7} &
  \multirow{-2}{*}{{\color[HTML]{030191} 5.8}} &
  \multirow{-2}{*}{{\color[HTML]{030191} 0.24}} \\ \bottomrule
\end{tabular}

\begin{tikzpicture}[remember picture,overlay]
  \tikzstyle{CompareLine}=[draw=CompareArrow,line width=2pt,<->] 
  \tikzstyle{CompareLabel}=[font=\bfseries, text=CompareArrow, anchor=west]

\newcommand{\CompareV}[6][]{%
  \draw[CompareLine,#1] 
    ($ (pic cs:#2) + (#5,#6) $) -- ($ (pic cs:#3) + (#5,0) $);
  \node[CompareLabel,#1] 
    at ($($(pic cs:#2)+(#5,#6)$)!0.5!($(pic cs:#3)+(#5,0)$) + (-0.2em,0)$) {#4};
}

\CompareV[]{at2}{ab2}{(a)}{0.5ex}{1.2ex} 
\CompareV[]{bt2}{bb2}{(b)}{1ex}{0.3ex}
\CompareV[]{ct2}{cb2}{(c)}{8ex}{1.2ex}
  \CompareV[]{dt2}{db2}{(d)}{1ex}{3ex}
  \CompareV[]{et2}{eb2}{(e)}{1.5ex}{2ex}
\end{tikzpicture}
}

\end{minipage}

\caption{Performance of {\color[HTML]{030191}adapter-based} approaches compared to the prefill baseline. \textcolor{Dandelion!90!black}{\textbf{(a)}} {\color[HTML]{030191}Adapters} have worse control consistency (F1,RMSE) compared to prefill+finetune, although generation quality is comparable. \textcolor{Dandelion!90!black}{\textbf{(b)}} The addition adapter is better for canny conditioning, whilst the prefill adapter is better for depth conditioning. \textcolor{Dandelion!90!black}{\textbf{(c)}} Adjusting sampling parameters (such as ctrl-G) can make up some of the difference in control consistency. \textcolor{Dandelion!90!black}{\textbf{(d)}} The prefill adapter is a little more parameter efficient (no zero layers) and considerably more inference efficient than the addition adapter. \textcolor{Dandelion!90!black}{\textbf{(e)}} Adapters enjoy no parameter overhead for ctrl-G, as the control-free base model weights are frozen and available.}
\label{tab:adapter}
\end{table}
\paragraph{Adapters vs prefill baseline (training on full data).} \cref{tab:adapter} shows results comparing the performance of the two adapter approaches to the prefill + finetuning baseline. We find that generally speaking adapters have \textit{worse} control consistency, but have comparable generation quality. Intuitively, an adapter is less \textit{flexible} to influence the generation process, limiting its ability to closely follow conditioning, but also retaining the original model's generation capabilities. This aligns with existing research on LoRA in language modelling \citep{lora_forgor}. The addition adapter has the advantage for canny conditioning, whilst the prefill adapter is superior for depth conditioning, although both are worse than full finetuning. We also demonstrate that adjusting the sampling parameters of an adapter (in this case ctrl-G) can make up some of difference in control consistency, although the associated tradeoffs discussed in \cref{sec:samp_params} still exist. We note that this suggests the comparisons made in ControlAR \citep{controlar} and Omnigen \citep{omnigen} may not be entirely fair, as their approaches train all parameters, but they compare against adapter-based approaches. 
\begin{takeawaybox}
    Adapters, where the generative model is frozen, can learn spatial conditioning. Control consistency is consistently worse than prefill + finetuning all parameters, although good generation quality is maintained. 
\end{takeawaybox}
We find that the prefill adapter is slightly more parameter efficient (due to not needing any additional zero-initialised layers), although it is of course possible to explore other ways of improving parameter efficiency, such as reducing the depth/number of blocks in the adapter or using LoRA \citep{lora}.\footnote{\cref{app:lora} contains a supplementary experiment that shows that LoRA can be used to improve parameter efficiency at the cost of performance.} We note that prefilling (adapter or not) is considerably more inference efficient compared to the addition adapter based on ControlNet \citep{controlnet}. The former is able to utilise a KV-cache, meaning that inference over control tokens only happens once per generation, whilst in the latter case, each generation token forward pass is paired with a spatially aligned control token forward pass, significantly increasing inference costs. This difference is especially apparent for SiT, where the KV-cache can be re-used for all (64) denoising steps. We note that \citet{tan2024ominicontrol,tan2025ominicontrol2} make a similar observation when designing their method for conditional generation built around FLUX \citep{flux2024}.
\begin{takeawaybox}
    Prefilling (adapter or not) with control tokens is considerably more inference efficient than adapters that are run each generation step (\eg ControlNet \citep{controlnet}) due to the re-use of the control KV-cache.
\end{takeawaybox}
We also note that since adapters keep the original (no control) model's weights intact, there is no \textit{parameter} overhead for ctrl-G. As such it can be readily used and considered for many existing adapter-based systems. In \cref{app:ctrlnet} we demonstrate this, showing that applying ctrl-G to ControlNet++ \citep{controlnet_plus_plus} out of the box results in similar qualitative and quantitative trade offs to \cref{sec:ctrlg} without any need for additional training (\cref{fig:t2i-tradeoff}).
\begin{takeawaybox}
    Ctrl-G can be used out-of-the-box with no parameter overhead for adapter-based approaches, meaning it can currently be readily applied to existing models/adapters to adjust generation behaviour.
\end{takeawaybox}
\paragraph{Forgetting.}
In order to investigate the phenomenon of ``forgetting'' we finetune and train adapters on a semantically pruned version of ImageNet where all samples with labels that are children of the WordNet \citep{miller-1990-wordnet} node ``organism'' are removed. We train for the same number of iterations as previously ($\sim$ 50K). Results are shown in \cref{fig:forgor,fig:forgor2}. We find that training all VAR parameters exhibits a collapse in generation quality, where the model ``forgets'' how to generate ``organism'' classes outside of the control finetuning data. Conversely, we observe that adapters are able to mitigate this effect, successfully generalising control-conditioned generation to classes unseen in the control data (although present in the base model's pretraining data). SiT interestingly is able to maintain its generation quality compared to finetuning on the full dataset. We hypothesise this robustness may be due to the larger size (675M vs 310M parameters) and longer pretraining (800 vs 200) of the SiT model \citep{repa} compared to the VAR model \citep{var}. 
\begin{takeawaybox}
    Full finetuning on insufficient control data may lead to catastrophic forgetting. Adapters are more robust to this, generalising control to parts of the data distribution seen in pre-training but not in control finetuning.
\end{takeawaybox}
Finally, to summarise the position and practical applicability of adapter-based approaches:
\begin{takeawaybox}
Practitioners should consider the suitability of adapters for their use case. Adapters endow \textit{pretrained} generative models with spatial controllability, mitigate ``forgetting'' when fine-tuning on limited data, and are modularised as an additional set of weights. However, control consistency is \textit{worse} than when training all parameters. 
\end{takeawaybox}
\section{Concluding Remarks}
In this work, we investigate spatially controlled image generation with transformers, performing controlled experiments over diffusion/flow and autoregressive models. We demonstrate that control token prefilling is a simple and well-performing baseline that can be generally applied across generation paradigms. We then find that simple sampling enhancements, such as control guidance and softmax truncation can meaningfully improve generation-control consistency. Finally, we reaffirm the practical motivation for adapter-based approaches, demonstrating that they can add spatial control to pretrained generative models, whilst maintaining generation quality/mitigating ``forgetting'' when finetuned on limited control data. However, this comes at the cost of inferior generation-control consistency compared to training all parameters. We hope the takeaways in this work can be useful to practitioners and researchers in the future.

\bibliography{main}

\begin{thebibliography}{54}
\providecommand{\natexlab}[1]{#1}
\providecommand{\url}[1]{\texttt{#1}}
\expandafter\ifx\csname urlstyle\endcsname\relax
  \providecommand{\doi}[1]{doi: #1}\else
  \providecommand{\doi}{doi: \begingroup \urlstyle{rm}\Url}\fi

\bibitem[Biderman et~al.(2024)Biderman, Portes, Ortiz, Paul, Greengard, Jennings, King, Havens, Chiley, Frankle, Blakeney, and Cunningham]{lora_forgor}
Dan Biderman, Jacob Portes, Jose Javier~Gonzalez Ortiz, Mansheej Paul, Philip Greengard, Connor Jennings, Daniel King, Sam Havens, Vitaliy Chiley, Jonathan Frankle, Cody Blakeney, and John~Patrick Cunningham.
\newblock Lo{RA} learns less and forgets less.
\newblock \emph{Transactions on Machine Learning Research}, 2024.
\newblock ISSN 2835-8856.
\newblock URL \url{https://openreview.net/forum?id=aloEru2qCG}.
\newblock Featured Certification.

\bibitem[Brooks et~al.(2023)Brooks, Holynski, and Efros]{pix2pix}
Tim Brooks, Aleksander Holynski, and Alexei~A. Efros.
\newblock Instructpix2pix: Learning to follow image editing instructions.
\newblock In \emph{Proceedings of the IEEE/CVF Conference on Computer Vision and Pattern Recognition (CVPR)}, pp.\  18392--18402, June 2023.

\bibitem[Canny(1986)]{canny}
John Canny.
\newblock A computational approach to edge detection.
\newblock \emph{IEEE Transactions on Pattern Analysis and Machine Intelligence}, PAMI-8\penalty0 (6):\penalty0 679--698, 1986.
\newblock \doi{10.1109/TPAMI.1986.4767851}.

\bibitem[Chang et~al.(2022)Chang, Zhang, Jiang, Liu, and Freeman]{maskgit}
Huiwen Chang, Han Zhang, Lu~Jiang, Ce~Liu, and William~T. Freeman.
\newblock Maskgit: Masked generative image transformer.
\newblock In \emph{Proceedings of the IEEE/CVF Conference on Computer Vision and Pattern Recognition (CVPR)}, pp.\  11315--11325, June 2022.

\bibitem[Chen et~al.(2024{\natexlab{a}})Chen, Wu, Luo, Xie, Paul, Luo, Zhao, and Li]{chen2024pixartdelta}
Junsong Chen, Yue Wu, Simian Luo, Enze Xie, Sayak Paul, Ping Luo, Hang Zhao, and Zhenguo Li.
\newblock Pixart-{$\delta$}: Fast and controllable image generation with latent consistency models, 2024{\natexlab{a}}.
\newblock URL \url{https://arxiv.org/abs/2401.05252}.

\bibitem[Chen et~al.(2024{\natexlab{b}})Chen, YU, GE, Yao, Xie, Wang, Kwok, Luo, Lu, and Li]{pixartalpha}
Junsong Chen, Jincheng YU, Chongjian GE, Lewei Yao, Enze Xie, Zhongdao Wang, James Kwok, Ping Luo, Huchuan Lu, and Zhenguo Li.
\newblock Pixart-\${\textbackslash}alpha\$: Fast training of diffusion transformer for photorealistic text-to-image synthesis.
\newblock In \emph{The Twelfth International Conference on Learning Representations}, 2024{\natexlab{b}}.
\newblock URL \url{https://openreview.net/forum?id=eAKmQPe3m1}.

\bibitem[Chen et~al.(2025{\natexlab{a}})Chen, Cai, Chen, Xie, Yang, Tang, Li, and Han]{deepcomp}
Junyu Chen, Han Cai, Junsong Chen, Enze Xie, Shang Yang, Haotian Tang, Muyang Li, and Song Han.
\newblock Deep compression autoencoder for efficient high-resolution diffusion models.
\newblock In \emph{The Thirteenth International Conference on Learning Representations}, 2025{\natexlab{a}}.
\newblock URL \url{https://openreview.net/forum?id=wH8XXUOUZU}.

\bibitem[Chen et~al.(2025{\natexlab{b}})Chen, Zhang, Zhang, Zhou, Kim, Liu, Li, Zhang, Zhao, Wang, Ding, Lin, and Zhao]{chen2024UniReal}
Xi~Chen, Zhifei Zhang, He~Zhang, Yuqian Zhou, Soo~Ye Kim, Qing Liu, Yijun Li, Jianming Zhang, Nanxuan Zhao, Yilin Wang, Hui Ding, Zhe Lin, and Hengshuang Zhao.
\newblock Unireal: Universal image generation and editing via learning real-world dynamics.
\newblock In \emph{Proceedings of the Computer Vision and Pattern Recognition Conference (CVPR)}, pp.\  12501--12511, June 2025{\natexlab{b}}.

\bibitem[Deng et~al.(2009)Deng, Dong, Socher, Li, Li, and Fei-Fei]{imagenet}
Jia Deng, Wei Dong, Richard Socher, Li-Jia Li, Kai Li, and Li~Fei-Fei.
\newblock Imagenet: A large-scale hierarchical image database.
\newblock In \emph{2009 IEEE Conference on Computer Vision and Pattern Recognition}, pp.\  248--255, 2009.
\newblock \doi{10.1109/CVPR.2009.5206848}.

\bibitem[Dhariwal \& Nichol(2021)Dhariwal and Nichol]{ADM}
Prafulla Dhariwal and Alexander~Quinn Nichol.
\newblock Diffusion models beat {GAN}s on image synthesis.
\newblock In A.~Beygelzimer, Y.~Dauphin, P.~Liang, and J.~Wortman Vaughan (eds.), \emph{Advances in Neural Information Processing Systems}, 2021.
\newblock URL \url{https://openreview.net/forum?id=OU98jZWS3x_}.

\bibitem[Esser et~al.(2021)Esser, Rombach, and Ommer]{taming}
Patrick Esser, Robin Rombach, and Bjorn Ommer.
\newblock Taming transformers for high-resolution image synthesis.
\newblock In \emph{Proceedings of the IEEE/CVF Conference on Computer Vision and Pattern Recognition (CVPR)}, pp.\  12873--12883, June 2021.

\bibitem[Esser et~al.(2024)Esser, Kulal, Blattmann, Entezari, M\"{u}ller, Saini, Levi, Lorenz, Sauer, Boesel, Podell, Dockhorn, English, and Rombach]{sd3}
Patrick Esser, Sumith Kulal, Andreas Blattmann, Rahim Entezari, Jonas M\"{u}ller, Harry Saini, Yam Levi, Dominik Lorenz, Axel Sauer, Frederic Boesel, Dustin Podell, Tim Dockhorn, Zion English, and Robin Rombach.
\newblock Scaling rectified flow transformers for high-resolution image synthesis.
\newblock In Ruslan Salakhutdinov, Zico Kolter, Katherine Heller, Adrian Weller, Nuria Oliver, Jonathan Scarlett, and Felix Berkenkamp (eds.), \emph{Proceedings of the 41st International Conference on Machine Learning}, volume 235 of \emph{Proceedings of Machine Learning Research}, pp.\  12606--12633. PMLR, 21--27 Jul 2024.
\newblock URL \url{https://proceedings.mlr.press/v235/esser24a.html}.

\bibitem[Gao et~al.(2024)Gao, Hoogeboom, Heek, Bortoli, Murphy, and Salimans]{gao2025diffusionmeetsflow}
Ruiqi Gao, Emiel Hoogeboom, Jonathan Heek, Valentin~De Bortoli, Kevin~P. Murphy, and Tim Salimans.
\newblock Diffusion meets flow matching: Two sides of the same coin.
\newblock 2024.
\newblock URL \url{https://diffusionflow.github.io/}.

\bibitem[Heusel et~al.(2017)Heusel, Ramsauer, Unterthiner, Nessler, and Hochreiter]{IS}
Martin Heusel, Hubert Ramsauer, Thomas Unterthiner, Bernhard Nessler, and Sepp Hochreiter.
\newblock Gans trained by a two time-scale update rule converge to a local nash equilibrium.
\newblock In \emph{Advances in Neural Information Processing Systems}, volume~30, 2017.

\bibitem[Ho \& Salimans(2021)Ho and Salimans]{cfg}
Jonathan Ho and Tim Salimans.
\newblock Classifier-free diffusion guidance.
\newblock In \emph{NeurIPS 2021 Workshop on Deep Generative Models and Downstream Applications}, 2021.
\newblock URL \url{https://openreview.net/forum?id=qw8AKxfYbI}.

\bibitem[Holtzman et~al.(2020)Holtzman, Buys, Du, Forbes, and Choi]{topp}
Ari Holtzman, Jan Buys, Li~Du, Maxwell Forbes, and Yejin Choi.
\newblock The curious case of neural text degeneration.
\newblock In \emph{International Conference on Learning Representations (ICLR)}, 2020.

\bibitem[Hu et~al.(2022)Hu, yelong shen, Wallis, Allen-Zhu, Li, Wang, Wang, and Chen]{lora}
Edward~J Hu, yelong shen, Phillip Wallis, Zeyuan Allen-Zhu, Yuanzhi Li, Shean Wang, Lu~Wang, and Weizhu Chen.
\newblock Lo{RA}: Low-rank adaptation of large language models.
\newblock In \emph{International Conference on Learning Representations}, 2022.
\newblock URL \url{https://openreview.net/forum?id=nZeVKeeFYf9}.

\bibitem[Huang et~al.(2023)Huang, Sun, Xie, Li, and Liu]{huang2023t2icompbench}
Kaiyi Huang, Kaiyue Sun, Enze Xie, Zhenguo Li, and Xihui Liu.
\newblock T2i-compbench: A comprehensive benchmark for open-world compositional text-to-image generation.
\newblock \emph{Advances in Neural Information Processing Systems}, 36:\penalty0 78723--78747, 2023.

\bibitem[Ingraham et~al.(2023)Ingraham, Baranov, Costello, Barber, Wang, Ismail, Frappier, Lord, Ng-Thow-Hing, Van~Vlack, Tie, Xue, Cowles, Leung, Rodrigues, Morales-Perez, Ayoub, Green, Puentes, Oplinger, Panwar, Obermeyer, Root, Beam, Poelwijk, and Grigoryan]{score_anneal}
John~B. Ingraham, Max Baranov, Zak Costello, Karl~W. Barber, Wujie Wang, Ahmed Ismail, Vincent Frappier, Dana~M. Lord, Christopher Ng-Thow-Hing, Erik~R. Van~Vlack, Shan Tie, Vincent Xue, Sarah~C. Cowles, Alan Leung, Jo{\~a}o~V. Rodrigues, Claudio~L. Morales-Perez, Alex~M. Ayoub, Robin Green, Katherine Puentes, Frank Oplinger, Nishant~V. Panwar, Fritz Obermeyer, Adam~R. Root, Andrew~L. Beam, Frank~J. Poelwijk, and Gevorg Grigoryan.
\newblock Illuminating protein space with a programmable generative model.
\newblock \emph{Nature}, 623\penalty0 (7989):\penalty0 1070--1078, Nov 2023.
\newblock ISSN 1476-4687.
\newblock \doi{10.1038/s41586-023-06728-8}.
\newblock URL \url{https://doi.org/10.1038/s41586-023-06728-8}.

\bibitem[Jiao et~al.(2025)Jiao, Zhang, Qian, Huang, Zhao, Shi, Ma, Wei, and Jie]{flexvar}
Siyu Jiao, Gengwei Zhang, Yinlong Qian, Jiancheng Huang, Yao Zhao, Humphrey Shi, Lin Ma, Yunchao Wei, and Zequn Jie.
\newblock Flexvar: Flexible visual autoregressive modeling without residual prediction, 2025.
\newblock URL \url{https://arxiv.org/abs/2502.20313}.

\bibitem[Kalajdzievski(2023)]{rslora}
Damjan Kalajdzievski.
\newblock A rank stabilization scaling factor for fine-tuning with lora, 2023.
\newblock URL \url{https://arxiv.org/abs/2312.03732}.

\bibitem[Karras et~al.(2024)Karras, Aittala, Kynk{\"a}{\"a}nniemi, Lehtinen, Aila, and Laine]{autoguidance}
Tero Karras, Miika Aittala, Tuomas Kynk{\"a}{\"a}nniemi, Jaakko Lehtinen, Timo Aila, and Samuli Laine.
\newblock Guiding a diffusion model with a bad version of itself.
\newblock In \emph{The Thirty-eighth Annual Conference on Neural Information Processing Systems}, 2024.
\newblock URL \url{https://openreview.net/forum?id=bg6fVPVs3s}.

\bibitem[Labs(2024)]{flux2024}
Black~Forest Labs.
\newblock Flux.
\newblock \url{https://github.com/black-forest-labs/flux}, 2024.

\bibitem[Li et~al.(2024{\natexlab{a}})Li, Yang, Kuang, Wu, Wang, Xiao, and Chen]{controlnet_plus_plus}
Ming Li, Taojiannan Yang, Huafeng Kuang, Jie Wu, Zhaoning Wang, Xuefeng Xiao, and Chen Chen.
\newblock Controlnet++: Improving conditional controls with efficient consistency feedback.
\newblock In \emph{European Conference on Computer Vision (ECCV)}, 2024{\natexlab{a}}.

\bibitem[Li et~al.(2024{\natexlab{b}})Li, Qiu, Chen, Kuen, Lin, Singh, and Raj]{controlvar}
Xiang Li, Kai Qiu, Hao Chen, Jason Kuen, Zhe Lin, Rita Singh, and Bhiksha Raj.
\newblock Controlvar: Exploring controllable visual autoregressive modeling.
\newblock \emph{arXiv preprint arXiv:2406.09750}, 2024{\natexlab{b}}.

\bibitem[Li et~al.(2023)Li, Liu, Wu, Mu, Yang, Gao, Li, and Lee]{gligen}
Yuheng Li, Haotian Liu, Qingyang Wu, Fangzhou Mu, Jianwei Yang, Jianfeng Gao, Chunyuan Li, and Yong~Jae Lee.
\newblock Gligen: Open-set grounded text-to-image generation.
\newblock In \emph{Proceedings of the IEEE/CVF Conference on Computer Vision and Pattern Recognition (CVPR)}, pp.\  22511--22521, June 2023.

\bibitem[Li et~al.(2025)Li, Cheng, Chen, Sun, Shen, Ran, Chen, Liu, and Wang]{controlar}
Zongming Li, Tianheng Cheng, Shoufa Chen, Peize Sun, Haocheng Shen, Longjin Ran, Xiaoxin Chen, Wenyu Liu, and Xinggang Wang.
\newblock Control{AR}: Controllable image generation with autoregressive models.
\newblock In \emph{The Thirteenth International Conference on Learning Representations}, 2025.
\newblock URL \url{https://openreview.net/forum?id=BWuBDdXVnH}.

\bibitem[Lipman et~al.(2023)Lipman, Chen, Ben-Hamu, Nickel, and Le]{lipman2023flow}
Yaron Lipman, Ricky T.~Q. Chen, Heli Ben-Hamu, Maximilian Nickel, and Matthew Le.
\newblock Flow matching for generative modeling.
\newblock In \emph{The Eleventh International Conference on Learning Representations}, 2023.
\newblock URL \url{https://openreview.net/forum?id=PqvMRDCJT9t}.

\bibitem[Ma et~al.(2024)Ma, Goldstein, Albergo, Boffi, Vanden-Eijnden, and Xie]{SiT}
Nanye Ma, Mark Goldstein, Michael~S. Albergo, Nicholas~M. Boffi, Eric Vanden-Eijnden, and Saining Xie.
\newblock Sit: Exploring flow and diffusion-based generative models with scalable interpolant transformers.
\newblock In Ale{\v{s}} Leonardis, Elisa Ricci, Stefan Roth, Olga Russakovsky, Torsten Sattler, and G{\"u}l Varol (eds.), \emph{Computer Vision -- ECCV 2024}, pp.\  23--40, Cham, 2024. Springer Nature Switzerland.
\newblock ISBN 978-3-031-72980-5.

\bibitem[Miller et~al.(1990)Miller, Beckwith, Fellbaum, Gross, and Miller]{miller-1990-wordnet}
G.~A. Miller, R.~Beckwith, C.~Fellbaum, D.~Gross, and K.~J. Miller.
\newblock {Introduction to WordNet: an on‑line lexical database}.
\newblock \emph{International Journal of Lexicography}, 3\penalty0 (4):\penalty0 235--244, 1990.
\newblock URL \url{http://wordnetcode.princeton.edu/5papers.pdf}.

\bibitem[Mou et~al.(2024)Mou, Wang, Xie, Wu, Zhang, Qi, and Shan]{t2iadapter}
Chong Mou, Xintao Wang, Liangbin Xie, Yanze Wu, Jian Zhang, Zhongang Qi, and Ying Shan.
\newblock T2i-adapter: learning adapters to dig out more controllable ability for text-to-image diffusion models.
\newblock In \emph{Proceedings of the Thirty-Eighth AAAI Conference on Artificial Intelligence and Thirty-Sixth Conference on Innovative Applications of Artificial Intelligence and Fourteenth Symposium on Educational Advances in Artificial Intelligence}, AAAI'24/IAAI'24/EAAI'24. AAAI Press, 2024.
\newblock ISBN 978-1-57735-887-9.
\newblock \doi{10.1609/aaai.v38i5.28226}.
\newblock URL \url{https://doi.org/10.1609/aaai.v38i5.28226}.

\bibitem[Peebles \& Xie(2023)Peebles and Xie]{dit}
William Peebles and Saining Xie.
\newblock Scalable diffusion models with transformers.
\newblock In \emph{Proceedings of the IEEE/CVF International Conference on Computer Vision (ICCV)}, pp.\  4195--4205, October 2023.

\bibitem[Podell et~al.(2024)Podell, English, Lacey, Blattmann, Dockhorn, M{\"u}ller, Penna, and Rombach]{sdxl}
Dustin Podell, Zion English, Kyle Lacey, Andreas Blattmann, Tim Dockhorn, Jonas M{\"u}ller, Joe Penna, and Robin Rombach.
\newblock {SDXL}: Improving latent diffusion models for high-resolution image synthesis.
\newblock In \emph{The Twelfth International Conference on Learning Representations}, 2024.
\newblock URL \url{https://openreview.net/forum?id=di52zR8xgf}.

\bibitem[Qin et~al.(2023)Qin, Zhang, Yu, Feng, Yang, Zhou, Wang, Niebles, Xiong, Savarese, Ermon, Fu, and Xu]{unicontrol}
Can Qin, Shu Zhang, Ning Yu, Yihao Feng, Xinyi Yang, Yingbo Zhou, Huan Wang, Juan~Carlos Niebles, Caiming Xiong, Silvio Savarese, Stefano Ermon, Yun Fu, and Ran Xu.
\newblock Unicontrol: A unified diffusion model for controllable visual generation in the wild.
\newblock In \emph{Thirty-seventh Conference on Neural Information Processing Systems}, 2023.
\newblock URL \url{https://openreview.net/forum?id=v54eUIayFh}.

\bibitem[Ranftl et~al.(2021)Ranftl, Bochkovskiy, and Koltun]{depth}
René Ranftl, Alexey Bochkovskiy, and Vladlen Koltun.
\newblock Vision transformers for dense prediction.
\newblock In \emph{Proceedings of the IEEE/CVF International Conference on Computer Vision (ICCV)}, pp.\  12179--12188, 2021.
\newblock \doi{10.1109/ICCV48922.2021.01198}.

\bibitem[Ren et~al.(2025)Ren, Yu, He, Shen, Yuille, and Chen]{ren2025flowar}
Sucheng Ren, Qihang Yu, Ju~He, Xiaohui Shen, Alan Yuille, and Liang-Chieh Chen.
\newblock Flowar: Scale-wise autoregressive image generation meets flow matching.
\newblock In \emph{ICML}, 2025.

\bibitem[Rombach et~al.(2022)Rombach, Blattmann, Lorenz, Esser, and Ommer]{LDM}
Robin Rombach, Andreas Blattmann, Dominik Lorenz, Patrick Esser, and Bj\"orn Ommer.
\newblock High-resolution image synthesis with latent diffusion models.
\newblock In \emph{Proceedings of the IEEE/CVF Conference on Computer Vision and Pattern Recognition (CVPR)}, pp.\  10684--10695, June 2022.

\bibitem[Ronneberger et~al.(2015)Ronneberger, Fischer, and Brox]{unet}
Olaf Ronneberger, Philipp Fischer, and Thomas Brox.
\newblock U-net: Convolutional networks for biomedical image segmentation.
\newblock In Nassir Navab, Joachim Hornegger, William~M. Wells, and Alejandro~F. Frangi (eds.), \emph{Medical Image Computing and Computer-Assisted Intervention -- MICCAI 2015}, pp.\  234--241, Cham, 2015. Springer International Publishing.
\newblock ISBN 978-3-319-24574-4.

\bibitem[Sadat et~al.(2025)Sadat, Hilliges, and Weber]{projg}
Seyedmorteza Sadat, Otmar Hilliges, and Romann~M. Weber.
\newblock Eliminating oversaturation and artifacts of high guidance scales in diffusion models.
\newblock In \emph{The Thirteenth International Conference on Learning Representations}, 2025.
\newblock URL \url{https://openreview.net/forum?id=e2ONKX6qzJ}.

\bibitem[Salimans et~al.(2016)Salimans, Goodfellow, Zaremba, Cheung, Radford, and Chen]{FID}
Tim Salimans, Ian Goodfellow, Wojciech Zaremba, Vicki Cheung, Alec Radford, and Xi~Chen.
\newblock Improved techniques for training gans.
\newblock In \emph{Advances in Neural Information Processing Systems}, volume~29, 2016.

\bibitem[Song et~al.(2021)Song, Sohl-Dickstein, Kingma, Kumar, Ermon, and Poole]{scorebased}
Yang Song, Jascha Sohl-Dickstein, Diederik~P Kingma, Abhishek Kumar, Stefano Ermon, and Ben Poole.
\newblock Score-based generative modeling through stochastic differential equations.
\newblock In \emph{International Conference on Learning Representations}, 2021.
\newblock URL \url{https://openreview.net/forum?id=PxTIG12RRHS}.

\bibitem[Sun et~al.(2024)Sun, Jiang, Chen, Zhang, Peng, Luo, and Yuan]{llamagen}
Peize Sun, Yi~Jiang, Shoufa Chen, Shilong Zhang, Bingyue Peng, Ping Luo, and Zehuan Yuan.
\newblock Autoregressive model beats diffusion: Llama for scalable image generation.
\newblock \emph{arXiv preprint arXiv:2406.06525}, 2024.

\bibitem[Tan et~al.(2025{\natexlab{a}})Tan, Liu, Yang, Xue, and Wang]{tan2024ominicontrol}
Zhenxiong Tan, Songhua Liu, Xingyi Yang, Qiaochu Xue, and Xinchao Wang.
\newblock Ominicontrol: Minimal and universal control for diffusion transformer.
\newblock 2025{\natexlab{a}}.

\bibitem[Tan et~al.(2025{\natexlab{b}})Tan, Xue, Yang, Liu, and Wang]{tan2025ominicontrol2}
Zhenxiong Tan, Qiaochu Xue, Xingyi Yang, Songhua Liu, and Xinchao Wang.
\newblock Ominicontrol2: Efficient conditioning for diffusion transformers.
\newblock \emph{arXiv preprint arXiv:2503.08280}, 2025{\natexlab{b}}.

\bibitem[Tian et~al.(2024)Tian, Jiang, Yuan, PENG, and Wang]{var}
Keyu Tian, Yi~Jiang, Zehuan Yuan, BINGYUE PENG, and Liwei Wang.
\newblock Visual autoregressive modeling: Scalable image generation via next-scale prediction.
\newblock In \emph{The Thirty-eighth Annual Conference on Neural Information Processing Systems}, 2024.
\newblock URL \url{https://openreview.net/forum?id=gojL67CfS8}.

\bibitem[van~den Oord et~al.(2017)van~den Oord, Vinyals, and kavukcuoglu]{vqvae}
Aaron van~den Oord, Oriol Vinyals, and koray kavukcuoglu.
\newblock Neural discrete representation learning.
\newblock In I.~Guyon, U.~Von Luxburg, S.~Bengio, H.~Wallach, R.~Fergus, S.~Vishwanathan, and R.~Garnett (eds.), \emph{Advances in Neural Information Processing Systems}, volume~30. Curran Associates, Inc., 2017.
\newblock URL \url{https://proceedings.neurips.cc/paper_files/paper/2017/file/7a98af17e63a0ac09ce2e96d03992fbc-Paper.pdf}.

\bibitem[Vaswani et~al.(2017)Vaswani, Shazeer, Parmar, Uszkoreit, Jones, Gomez, Kaiser, and Polosukhin]{transformer}
Ashish Vaswani, Noam Shazeer, Niki Parmar, Jakob Uszkoreit, Llion Jones, Aidan~N Gomez, \L~ukasz Kaiser, and Illia Polosukhin.
\newblock Attention is all you need.
\newblock In I.~Guyon, U.~Von Luxburg, S.~Bengio, H.~Wallach, R.~Fergus, S.~Vishwanathan, and R.~Garnett (eds.), \emph{Advances in Neural Information Processing Systems}, volume~30. Curran Associates, Inc., 2017.
\newblock URL \url{https://proceedings.neurips.cc/paper_files/paper/2017/file/3f5ee243547dee91fbd053c1c4a845aa-Paper.pdf}.

\bibitem[Xiao et~al.(2025)Xiao, Wang, Zhou, Yuan, Xing, Yan, Li, Wang, Huang, and Liu]{omnigen}
Shitao Xiao, Yueze Wang, Junjie Zhou, Huaying Yuan, Xingrun Xing, Ruiran Yan, Chaofan Li, Shuting Wang, Tiejun Huang, and Zheng Liu.
\newblock Omnigen: Unified image generation.
\newblock In \emph{Proceedings of the Computer Vision and Pattern Recognition Conference (CVPR)}, pp.\  13294--13304, June 2025.

\bibitem[Yu et~al.(2024)Yu, Lezama, Gundavarapu, Versari, Sohn, Minnen, Cheng, Gupta, Gu, Hauptmann, Gong, Yang, Essa, Ross, and Jiang]{magvit2}
Lijun Yu, Jose Lezama, Nitesh~Bharadwaj Gundavarapu, Luca Versari, Kihyuk Sohn, David Minnen, Yong Cheng, Agrim Gupta, Xiuye Gu, Alexander~G Hauptmann, Boqing Gong, Ming-Hsuan Yang, Irfan Essa, David~A Ross, and Lu~Jiang.
\newblock Language model beats diffusion - tokenizer is key to visual generation.
\newblock In \emph{The Twelfth International Conference on Learning Representations}, 2024.
\newblock URL \url{https://openreview.net/forum?id=gzqrANCF4g}.

\bibitem[Yu et~al.(2025)Yu, Kwak, Jang, Jeong, Huang, Shin, and Xie]{repa}
Sihyun Yu, Sangkyung Kwak, Huiwon Jang, Jongheon Jeong, Jonathan Huang, Jinwoo Shin, and Saining Xie.
\newblock Representation alignment for generation: Training diffusion transformers is easier than you think.
\newblock In \emph{The Thirteenth International Conference on Learning Representations}, 2025.
\newblock URL \url{https://openreview.net/forum?id=DJSZGGZYVi}.

\bibitem[Zhang et~al.(2025{\natexlab{a}})Zhang, ang Gao, Jiang, Zhao, and Zheng]{ctrlu}
Guiyu Zhang, Huan ang Gao, Zijian Jiang, Hao Zhao, and Zhedong Zheng.
\newblock Ctrl-u: Robust conditional image generation via uncertainty-aware reward modeling.
\newblock In \emph{The Thirteenth International Conference on Learning Representations}, 2025{\natexlab{a}}.
\newblock URL \url{https://openreview.net/forum?id=eC2ICbECNM}.

\bibitem[Zhang et~al.(2023)Zhang, Rao, and Agrawala]{controlnet}
Lvmin Zhang, Anyi Rao, and Maneesh Agrawala.
\newblock Adding conditional control to text-to-image diffusion models.
\newblock In \emph{Proceedings of the IEEE/CVF International Conference on Computer Vision (ICCV)}, pp.\  3836--3847, October 2023.

\bibitem[Zhang et~al.(2025{\natexlab{b}})Zhang, Guo, Zhao, Fu, Duan, Wang, Chen, Xu, Luo, and Zhang]{zhang2025unified}
Xinjie Zhang, Jintao Guo, Shanshan Zhao, Minghao Fu, Lunhao Duan, Guo-Hua Wang, Qing-Guo Chen, Zhao Xu, Weihua Luo, and Kaifu Zhang.
\newblock Unified multimodal understanding and generation models: Advances, challenges, and opportunities.
\newblock \emph{arXiv preprint arXiv:2505.02567}, 2025{\natexlab{b}}.

\bibitem[Zhao et~al.(2023)Zhao, Chen, Chen, Bao, Hao, Yuan, and Wong]{unicontrolnet}
Shihao Zhao, Dongdong Chen, Yen-Chun Chen, Jianmin Bao, Shaozhe Hao, Lu~Yuan, and Kwan-Yee~K. Wong.
\newblock Uni-controlnet: All-in-one control to text-to-image diffusion models.
\newblock In \emph{Thirty-seventh Conference on Neural Information Processing Systems}, 2023.
\newblock URL \url{https://openreview.net/forum?id=VgQw8zXrH8}.

\end{thebibliography}
\bibliographystyle{tmlr}
\newpage
\appendix

\section{Additional Experiments}\label{app:exp}
\subsection{Distribution Truncation}
\begin{figure}
    \centering
    \includegraphics[width=.49\linewidth]{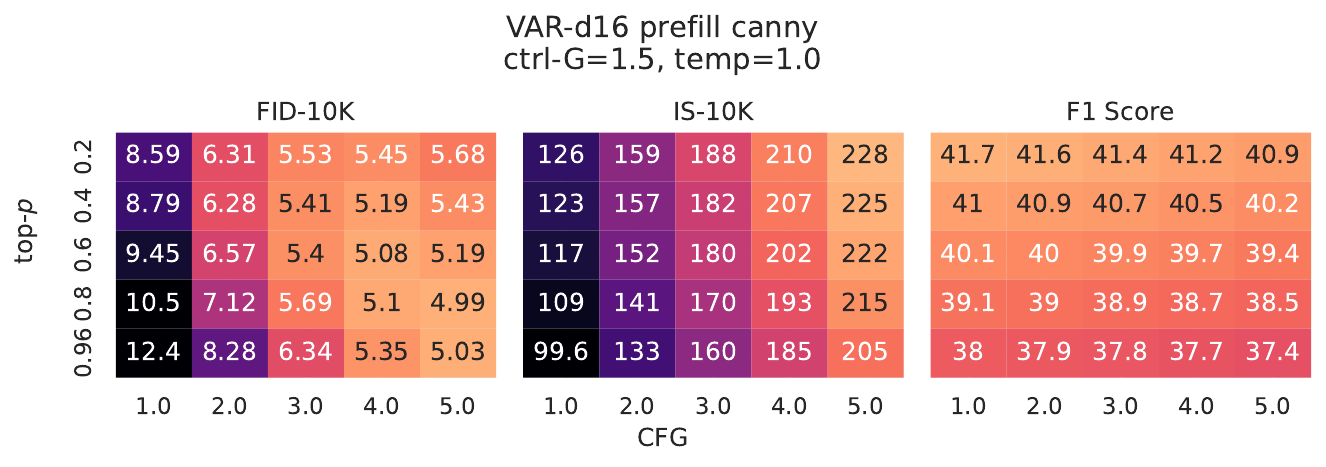}
    \includegraphics[width=.49\linewidth]{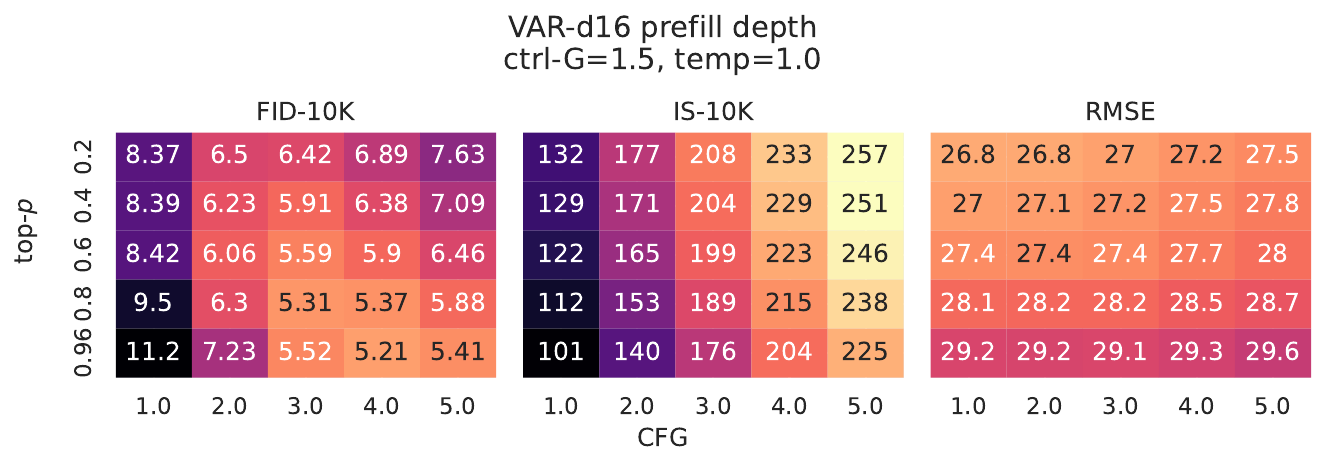}

    \includegraphics[width=.49\linewidth]{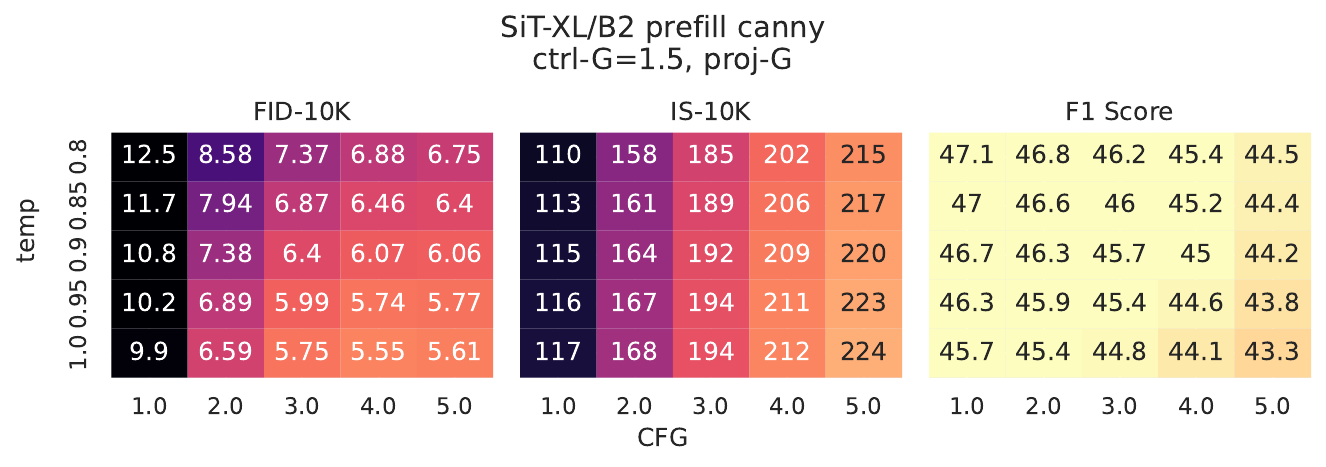}
    \includegraphics[width=.49\linewidth]{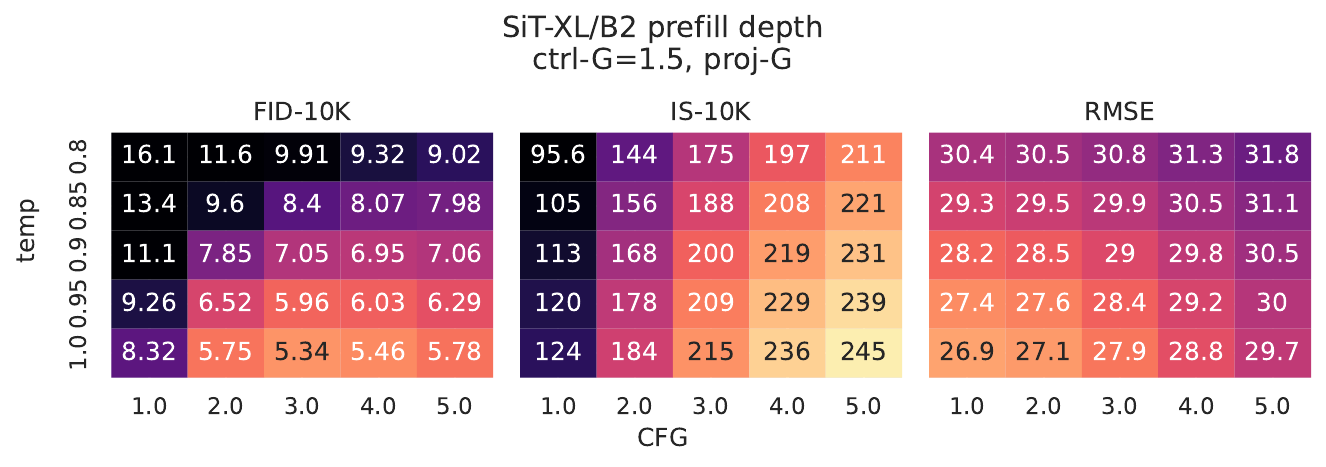}
    \vspace{-3mm}\caption{Effect of CFG and distribution truncation on conditional generation when ctrl-G=1.5. Brighter means better. The improvements in control consistency from softmax truncation via top-$p$ sampling stack with ctrl-G.}
    \label{fig:trunc-ctrlg}
\end{figure}

\cref{fig:trunc-ctrlg} shows how distribution truncation and CFG interact when ctrl-G=1.5. It tells a similar story to \cref{fig:ctrl-g} and also illustrates how both distribution truncation and control guidance can be adjusted together at inference time to improve control consistency. For VAR, the control consistency improvements from ctrl-G and top-$p$ stack together.

\subsection{Low Rank Adaptation (LoRA)}\label{app:lora}
\begin{table}[]
    \centering

\begin{tabular}{@{}lllllll@{}}
\toprule
\multirow{2}{*}{\begin{tabular}[c]{@{}l@{}}model \\ \scriptsize(sampling params)\end{tabular}} &
\multirow{2}{*}{method} &
\multirow{2}{*}{ctrl} &
\multirow{2}{*}{\begin{tabular}[c]{@{}l@{}}\#params \\ base + adapter\end{tabular}} &
\multicolumn{3}{c}{metrics} \\
& & & & FID-10K$\downarrow$ & IS-10K$\uparrow$ & F1$\uparrow$ \\
\midrule
\multirow{4}{*}{\begin{tabular}[c]{@{}l@{}}VAR-d16 \\ \scriptsize CFG=3.0, temp=1, \\ \scriptsize top-$p$=0.6, top-$k$=900, \\ \scriptsize ctrl-G=1.0\end{tabular}} &
prefill & canny & 310M & 5.12 &187 &36.9\\
& prefill + LoRA (r=64) & canny & 310M + 10M & 5.57 & 185 & 33.2 \\
\cmidrule(l){2-7}
& prefill adapter & canny & 310M + 202M &4.98& 201 &33.0 \\
& prefill adapter + LoRA (r=64) & canny & 310M + 10M & 5.59 & 182 & 28.9 \\
\bottomrule
\end{tabular}
    \caption{Experiments with Low-Rank Adaptation (LoRA). We use rsLoRA \citep{rslora} and set $r=\alpha=64$. We attach LoRA adapters to the $QKV$ matrices in each attention layer as well as the weight matrices in the feed-forward network. For prefill + LoRA the trainable weights see all tokens, whilst for prefill adapter + LoRA the trainable weights only see control tokens (\cref{fig:adapter-illust} where the trainable parameters are replaced with LoRA). LoRA enables much lower parameter overhead, but the reduced model flexibility comes at the cost of generation quality and control consistency.}
    \label{tab:lora}
\end{table}

\cref{tab:lora} shows an additional ablation, replacing the trainable parameters in \cref{fig:adapter-illust} with LoRA \citep{lora} adapters. We find that LoRA can considerably improve the parameter efficiency of learning a spatial control; however, it comes at the cost of generation quality as well as control consistency. This aligns with the behaviour reported in \citet{controlar}.

\subsection{Learning with Limited Data}
\begin{figure}
\centering

        \includegraphics[width=\textwidth]{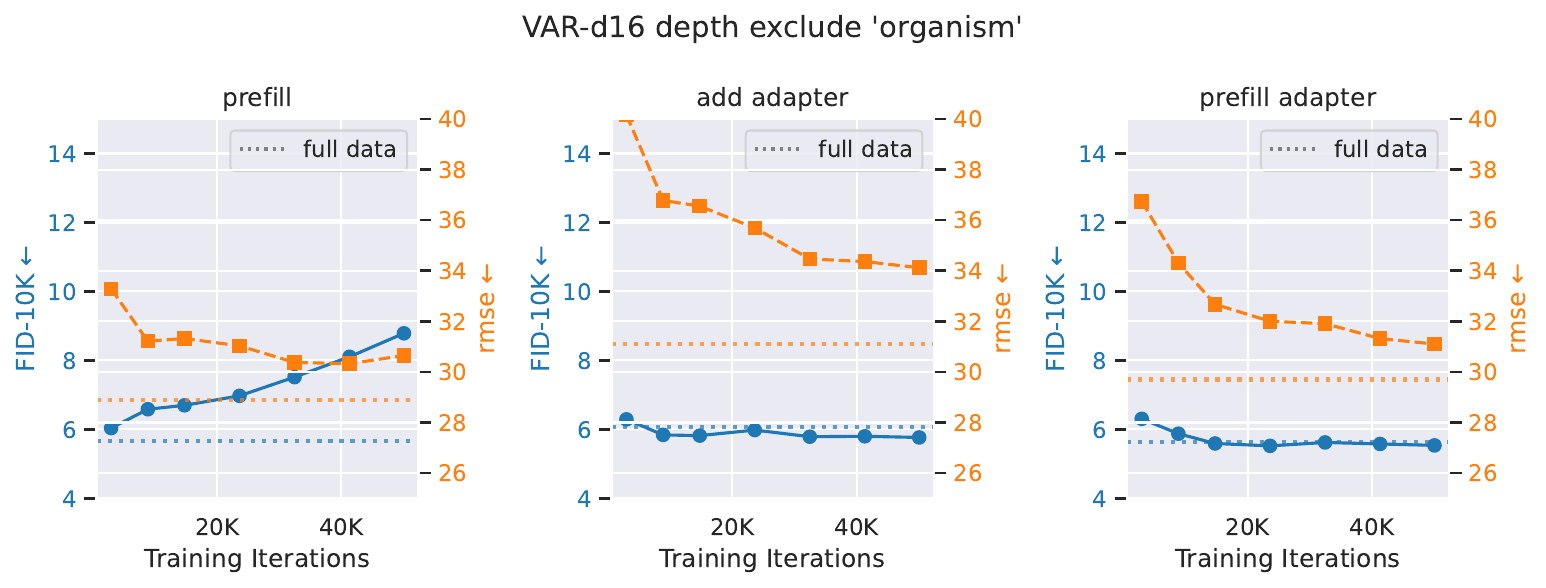}

        \includegraphics[width=\textwidth]{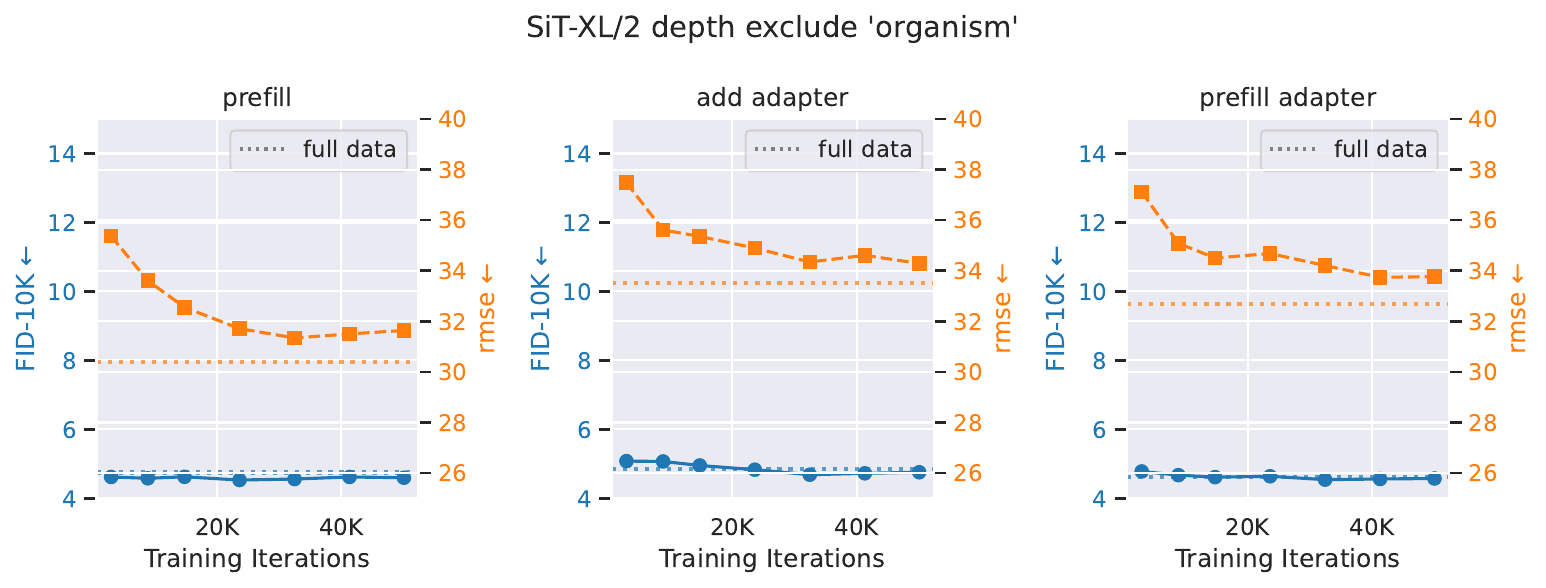}

    \vspace{-3mm}\caption{Generation quality (FID$\downarrow$) and control consistency (RMSE$\downarrow$), evaluated for the whole ImageNet distribution. Fully finetuning VAR (prefill) leads to a collapse in generation quality, whilst using adapters mitigates this. On the other hand the larger SiT model manages to maintain generation quality over the whole distribution.}
    \label{fig:forgor2}
\end{figure}

\cref{fig:forgor2} mirrors the left of \cref{fig:forgor} but for depth conditioning. We note that excluding ``organism'' classes from the finetuning data consistently hurts depth consistency (RMSE$\downarrow$). We hypothesise that this may be due to distributional shift between the depth maps of organism classes present at evaluation and non-organism classes used in training. 
\section{Example Generations}\label{app:example-gens}
\begin{figure}
\vspace{-3mm}
    \centering
    prefill, ctrl-G=1.0
    
    \includegraphics[width=0.49\linewidth]{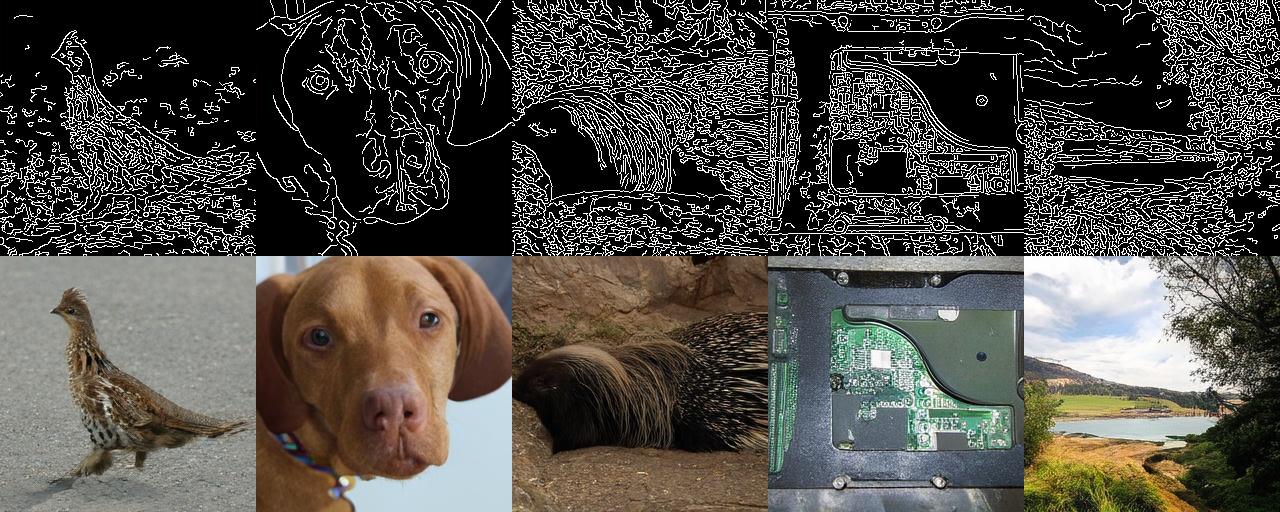}\hfill
    \includegraphics[width=0.49\linewidth]{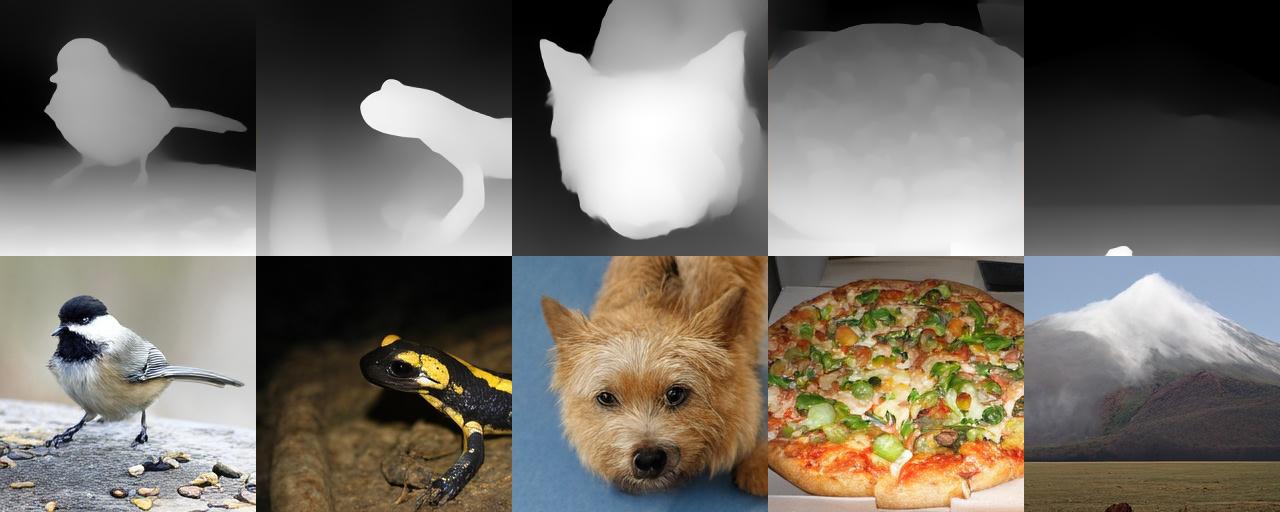} 
    
    prefill, ctrl-G=1.5
    
    \includegraphics[width=0.49\linewidth]{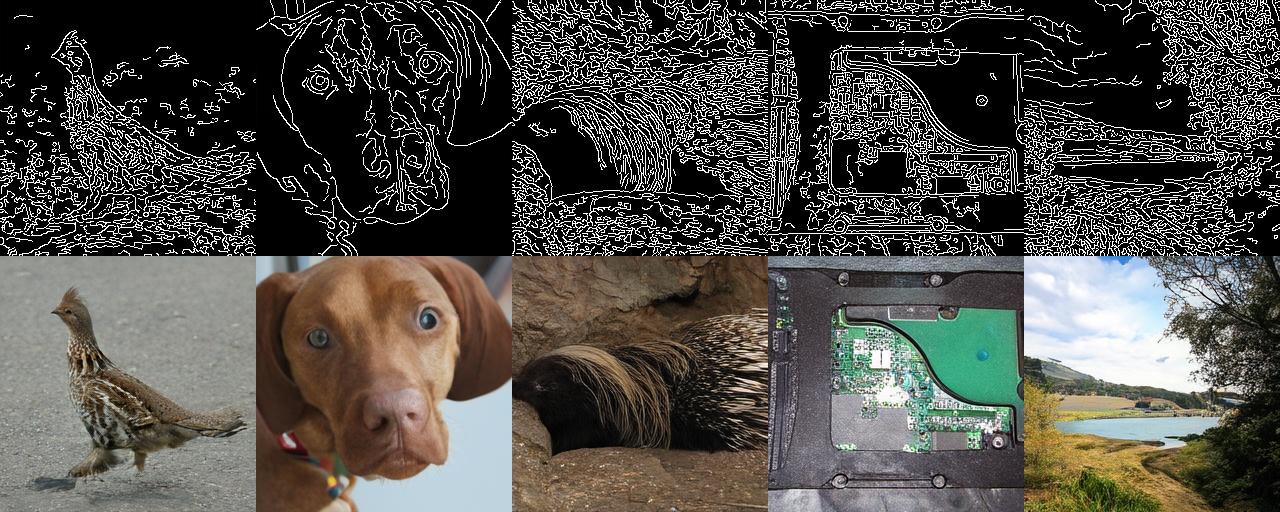}\hfill
    \includegraphics[width=0.49\linewidth]{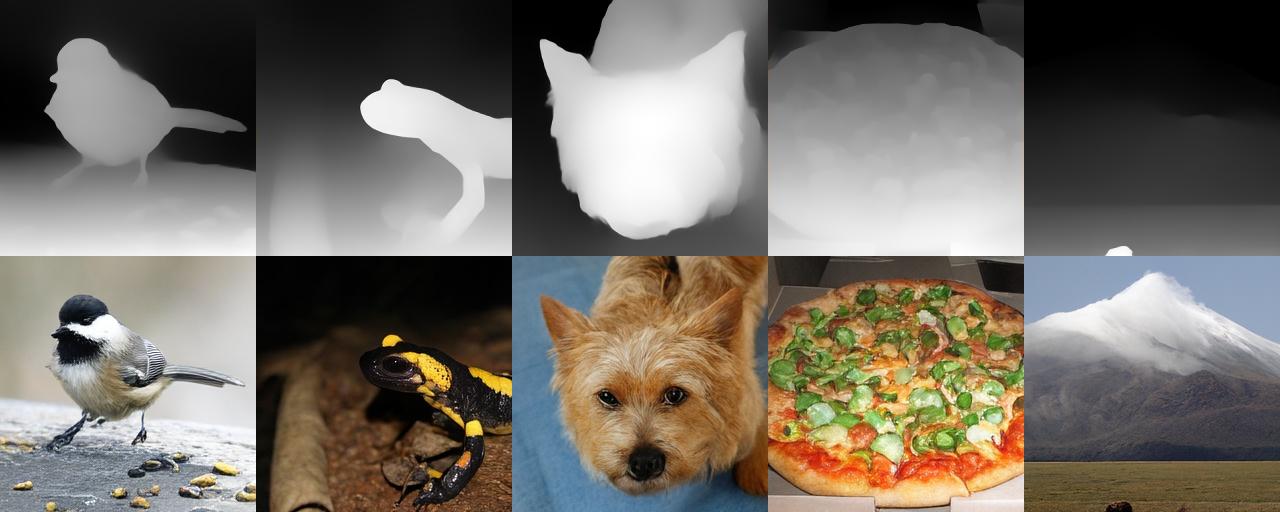}
    
    add adapter, ctrl-G=1.0 
    
    \includegraphics[width=0.49\linewidth]{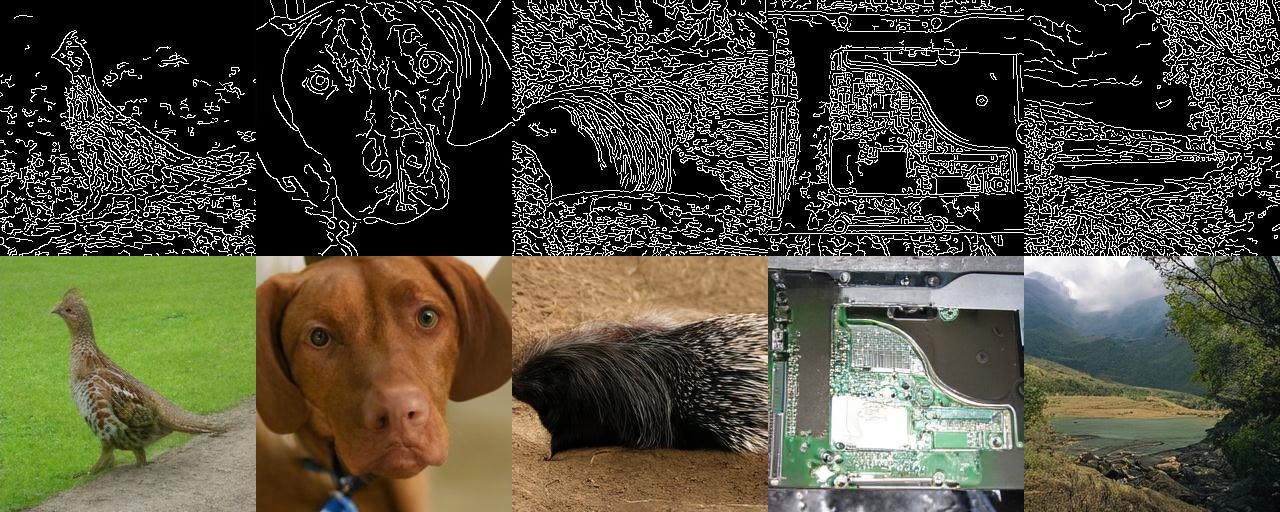}\hfill
    \includegraphics[width=0.49\linewidth]{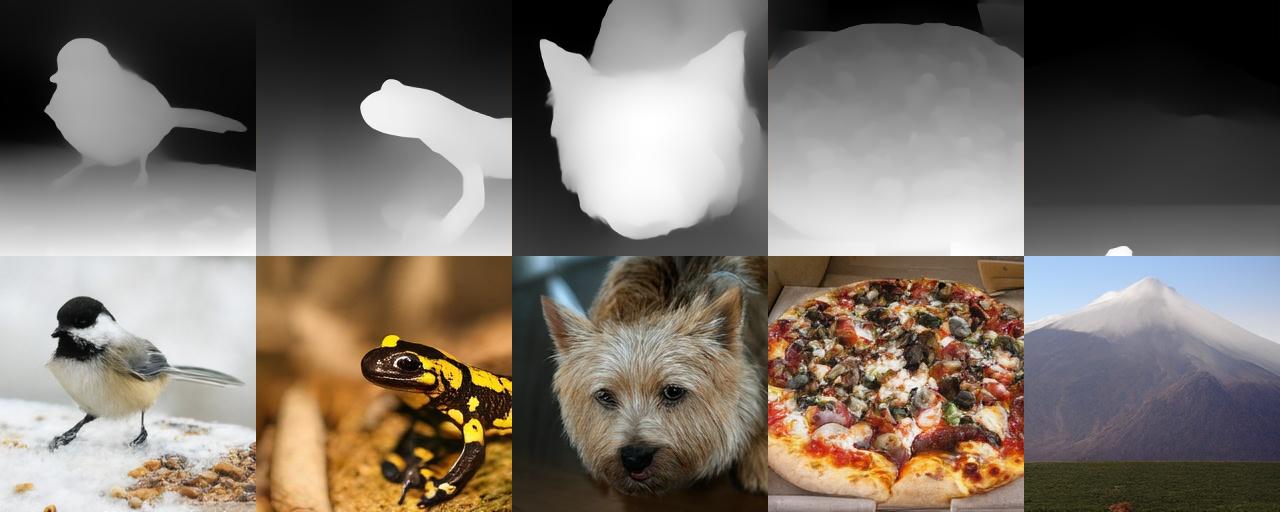}
    
    add adapter, ctrl-G=1.5 
    
    \includegraphics[width=0.49\linewidth]{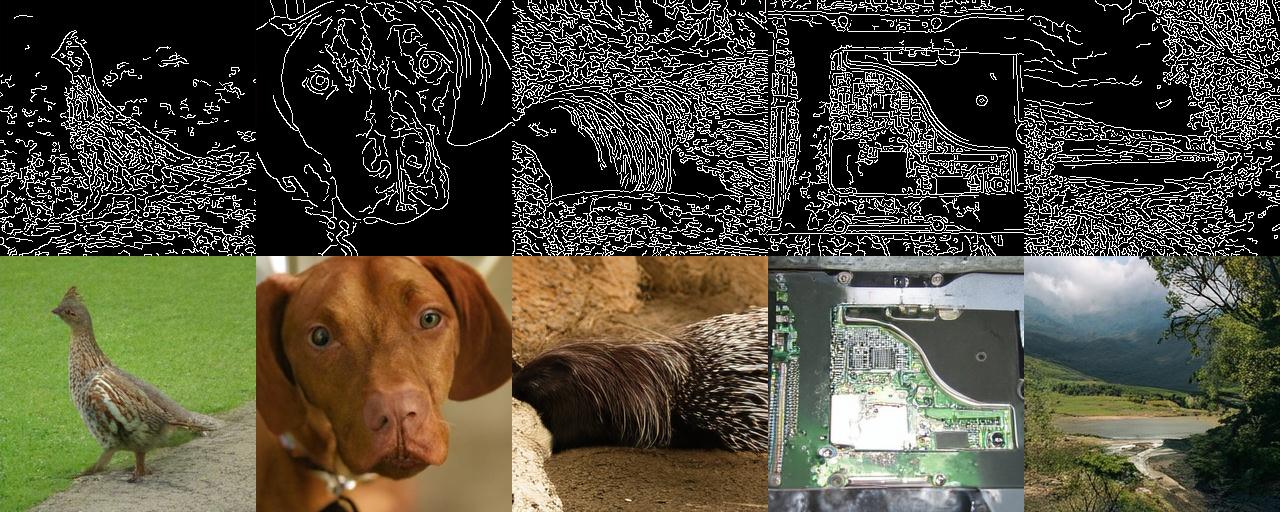}\hfill
    \includegraphics[width=0.49\linewidth]{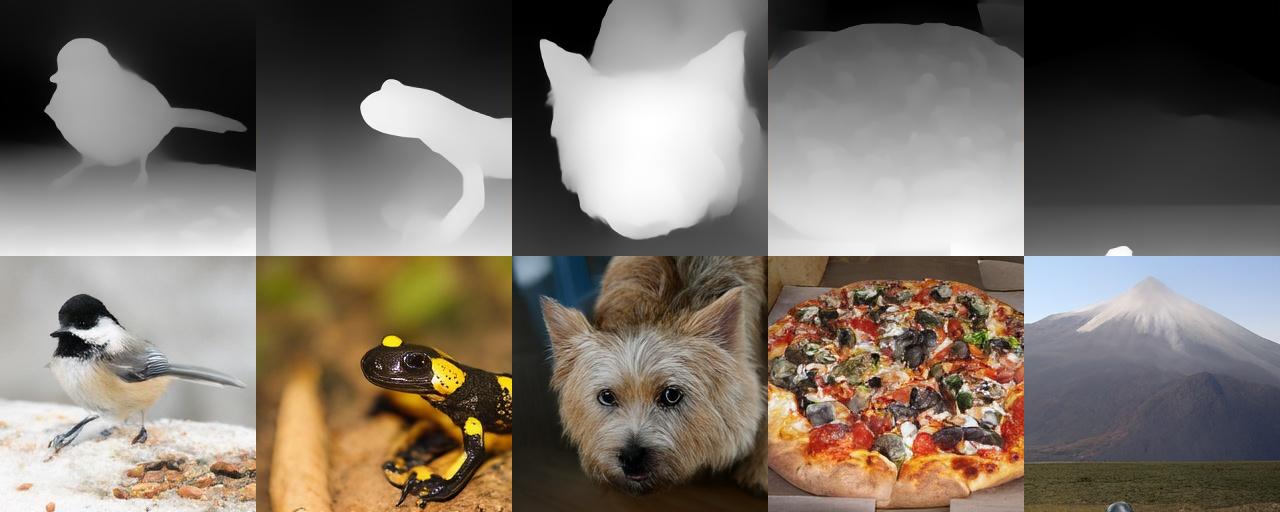}
    
    prefill adapter, ctrl-G=1.0
    
    \includegraphics[width=0.49\linewidth]{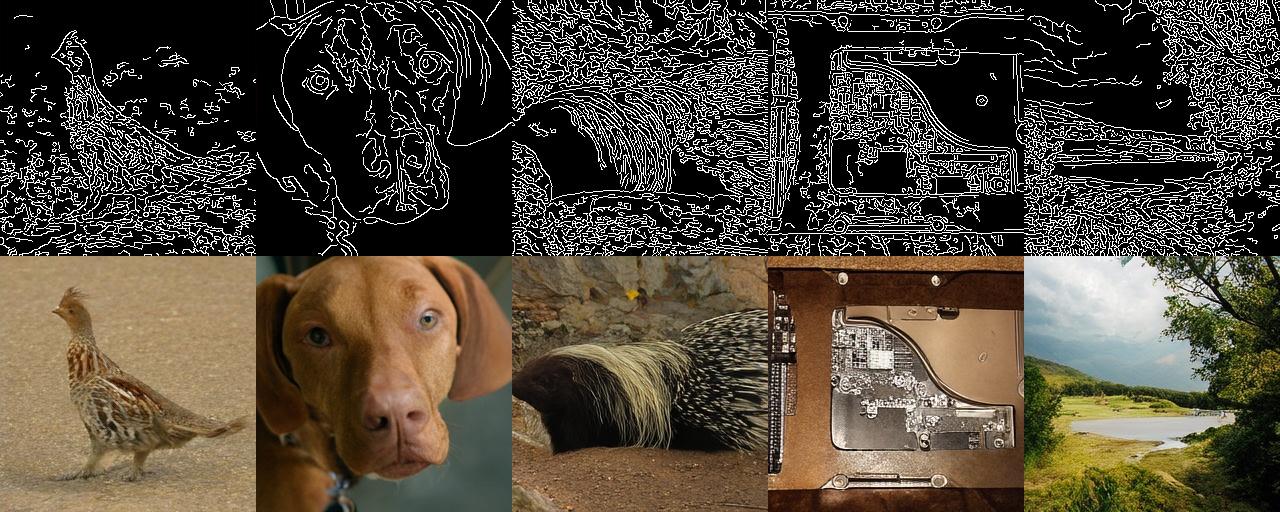}\hfill
    \includegraphics[width=0.49\linewidth]{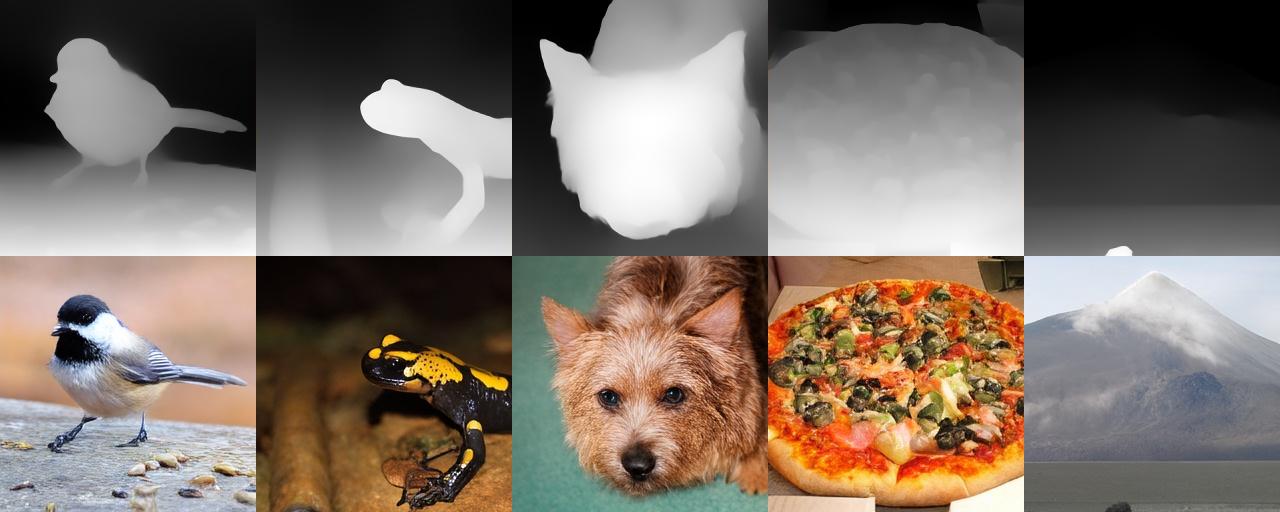}
    
    prefill adapter, ctrl-G=1.5
    
    \includegraphics[width=0.49\linewidth]{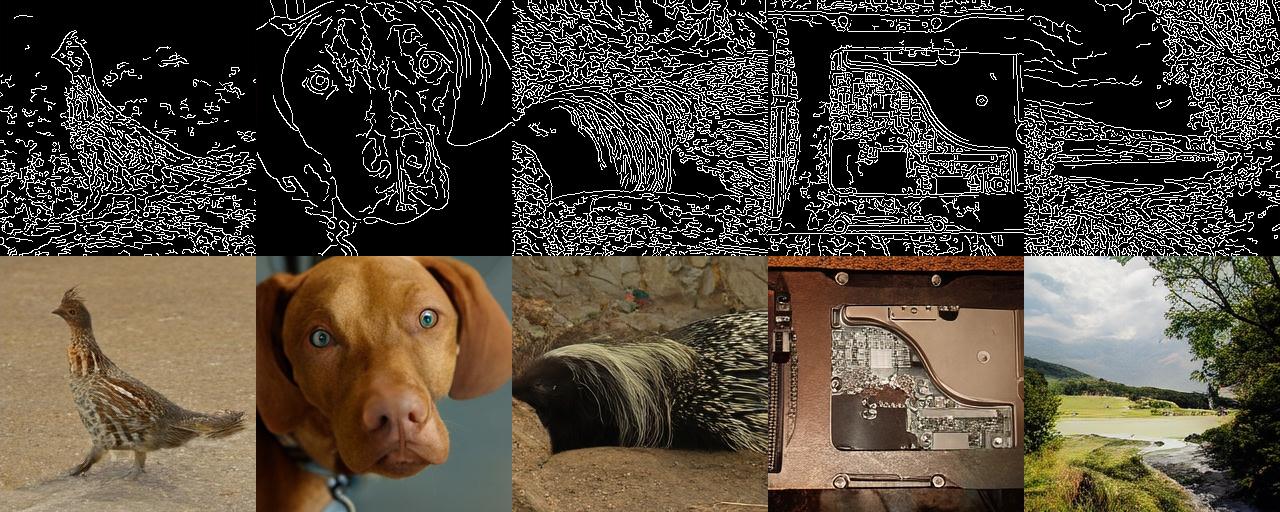}\hfill
    \includegraphics[width=0.49\linewidth]{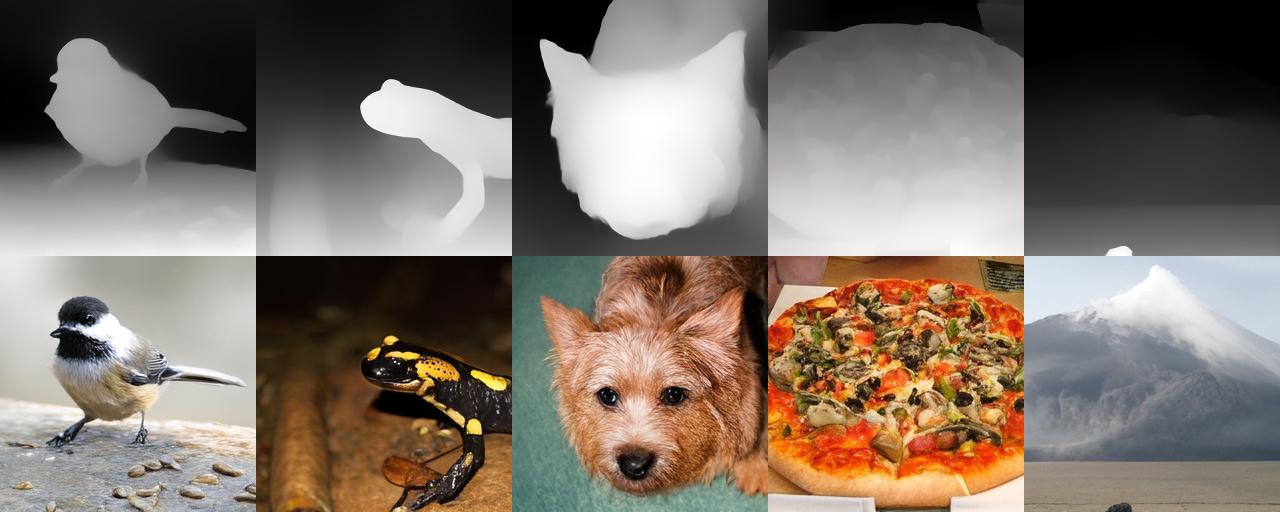}
    \caption{Additional examples of VAR generations. \\(CFG=3.0, temp=1, top-$p$=0.6, top-$k$=900)}
    \label{fig:var-ex1}
\end{figure}

\begin{figure}
    \centering
    prefill
    
    \includegraphics[width=0.49\linewidth]{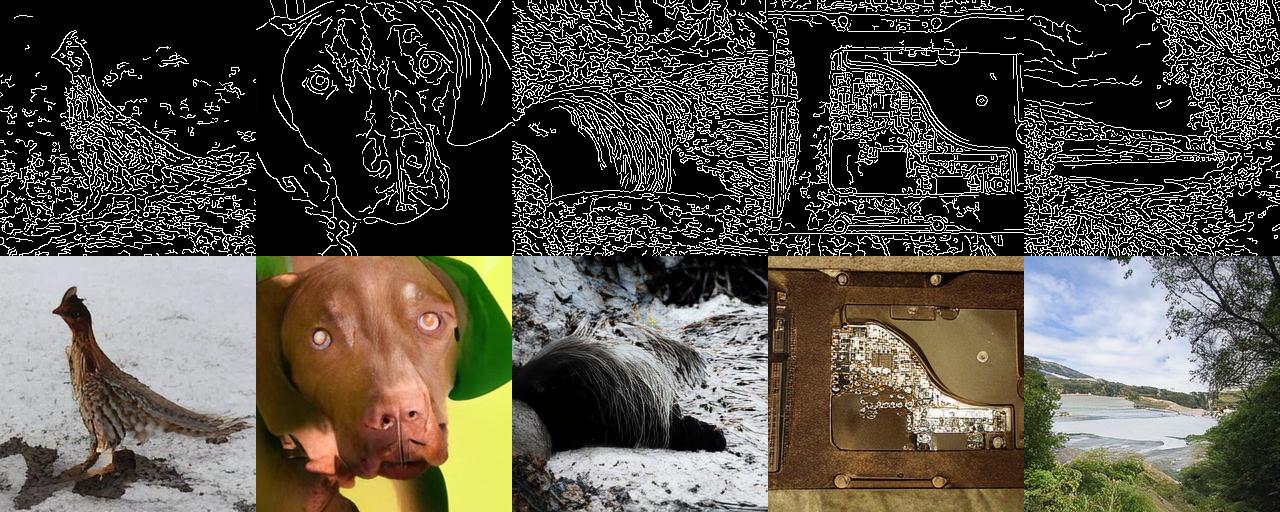}\hfill
    \includegraphics[width=0.49\linewidth]{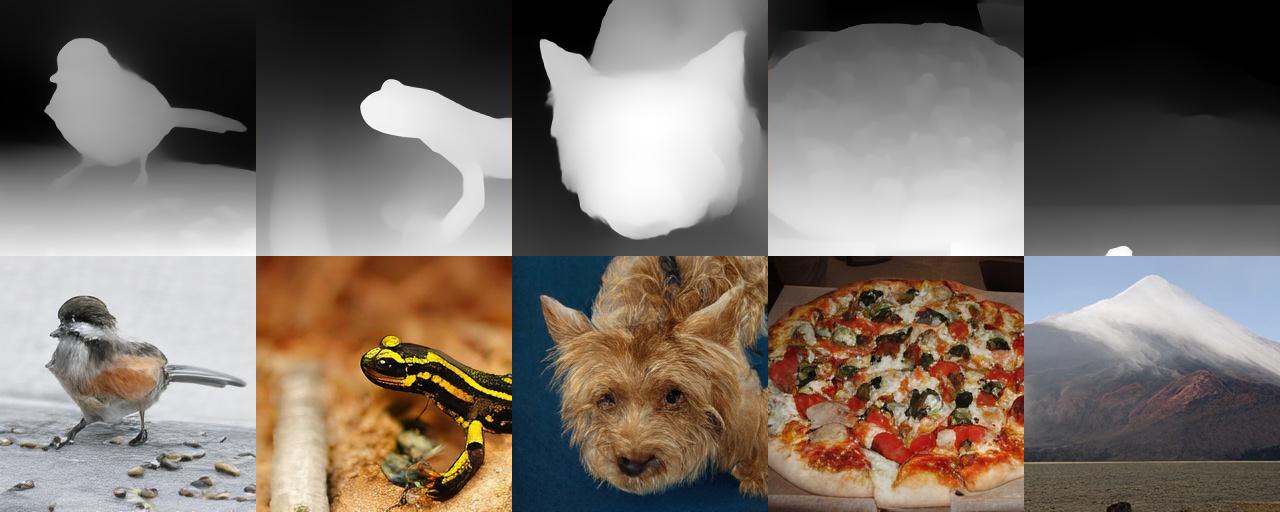}

  add adapter

    \includegraphics[width=0.49\linewidth]{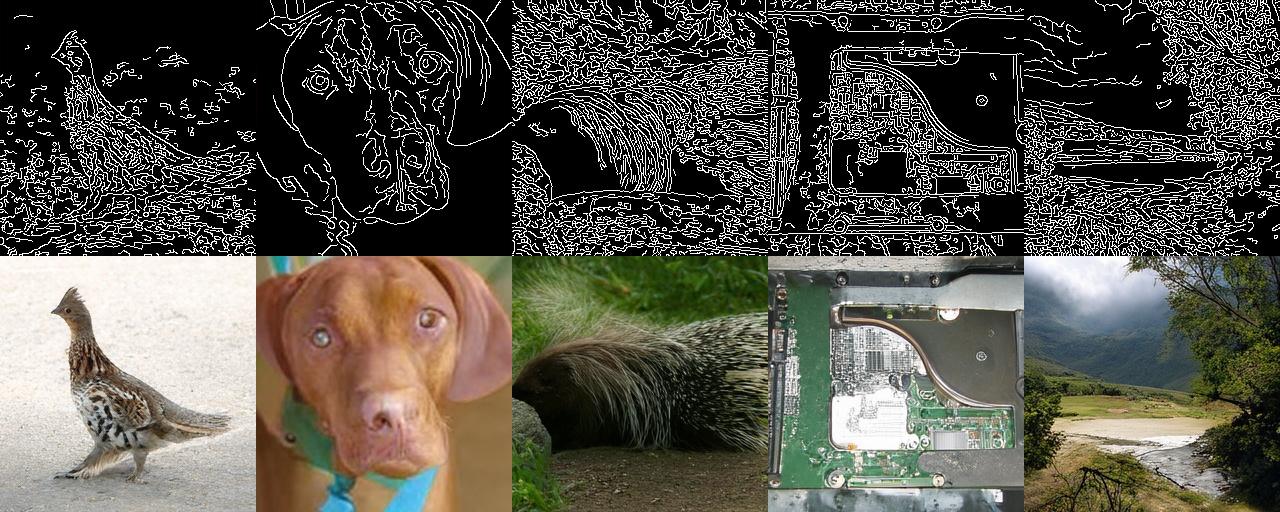}\hfill
    \includegraphics[width=0.49\linewidth]{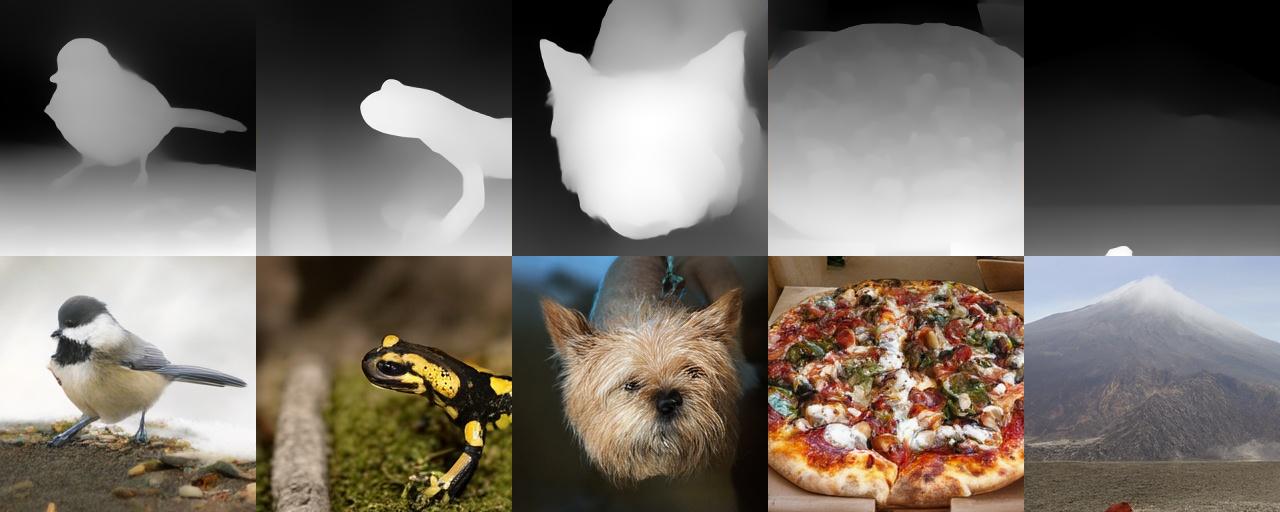}
    
    prefill adapter

    \includegraphics[width=0.49\linewidth]{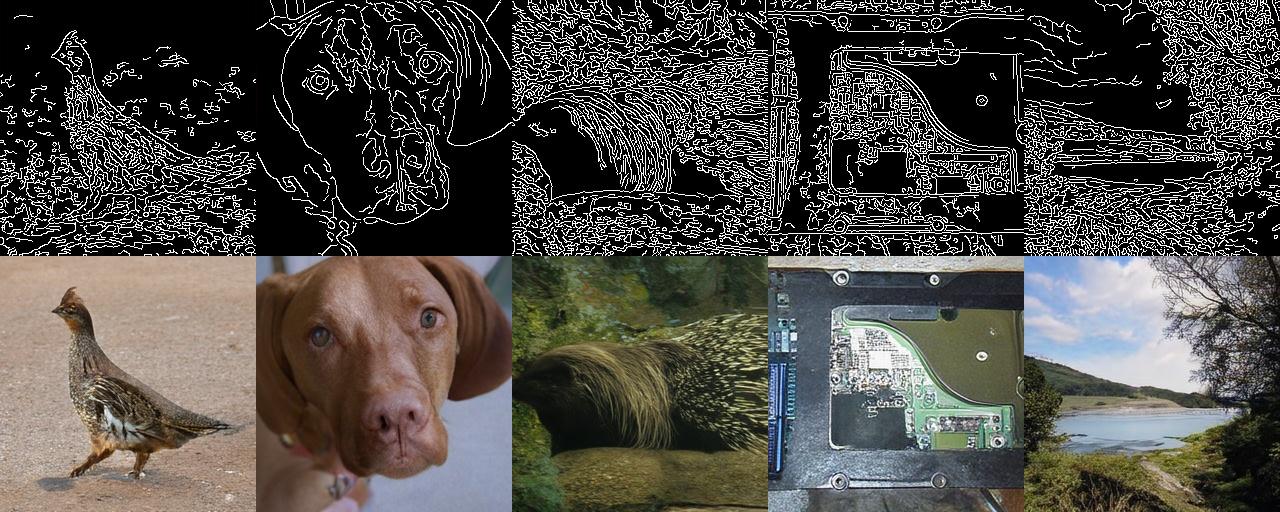}\hfill
    \includegraphics[width=0.49\linewidth]{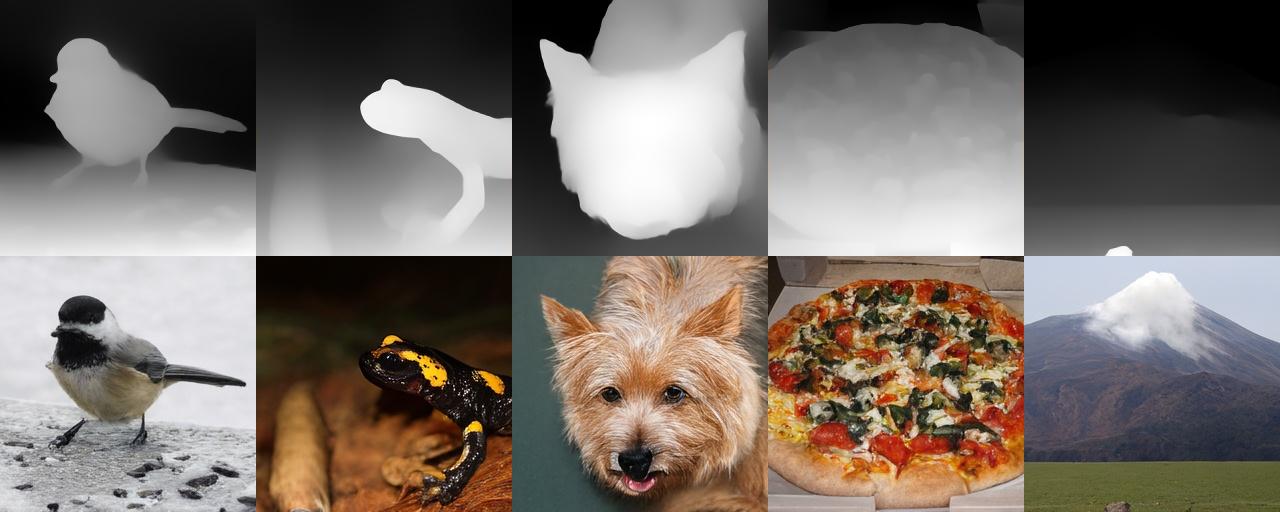}
    \caption{Additional examples of VAR generations with limited control data -- ``organism'' classes excluded. (CFG=3.0, temp=1, top-$p$=0.6, top-$k$=900, ctrl-G=1.0)}
    \label{fig:var-ex2}
\end{figure}

\begin{figure}
    \centering
    prefill, ctrl-G=1.0
    
    \includegraphics[width=0.49\linewidth]{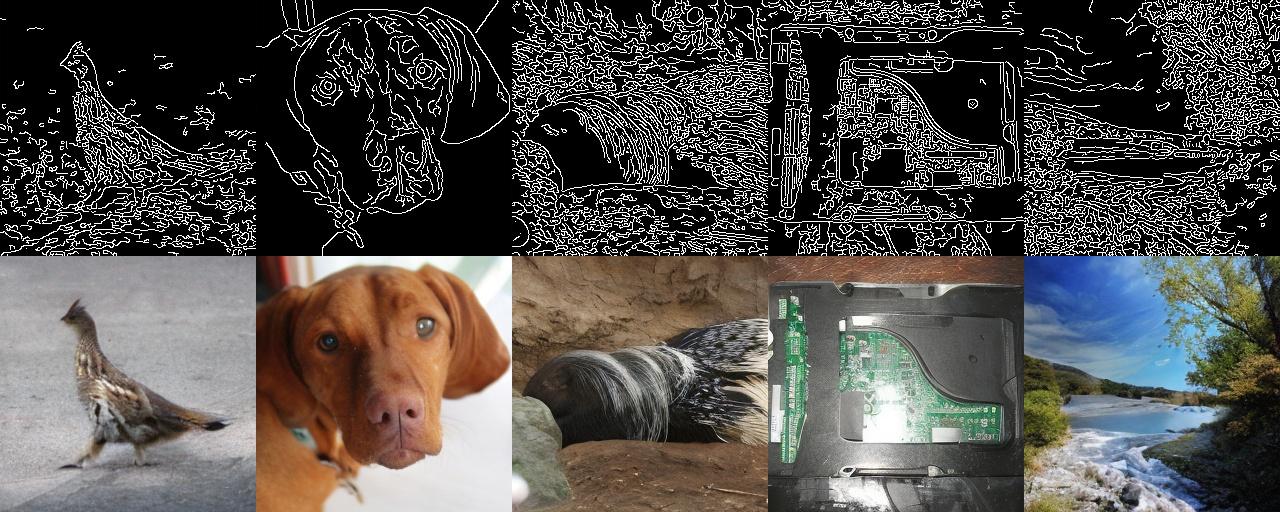}\hfill
    \includegraphics[width=0.49\linewidth]{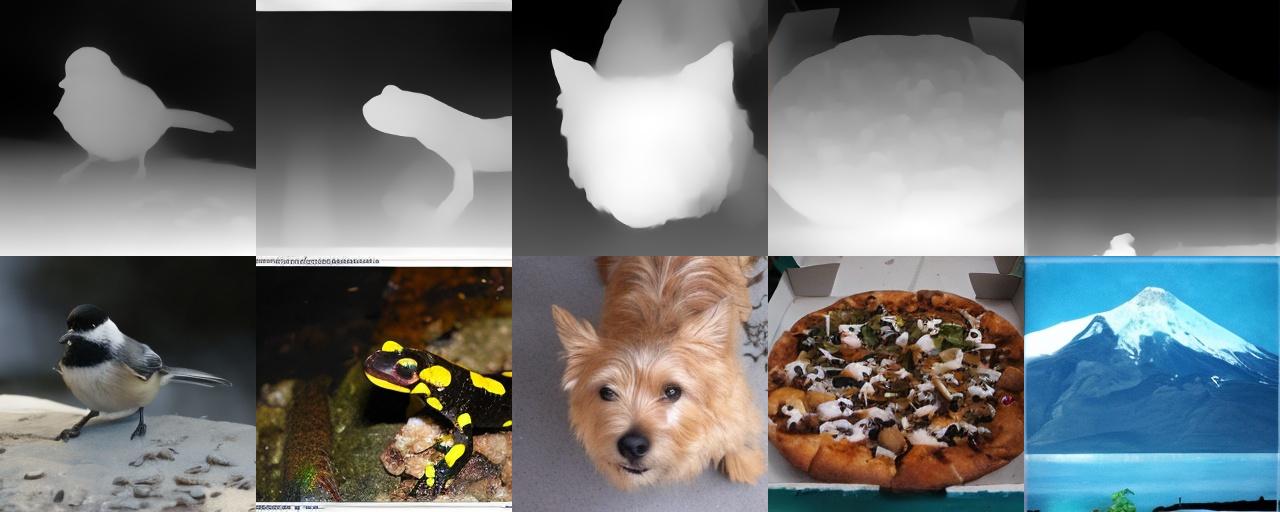}
    
    prefill, ctrl-G=1.5 
    
    \includegraphics[width=0.49\linewidth]{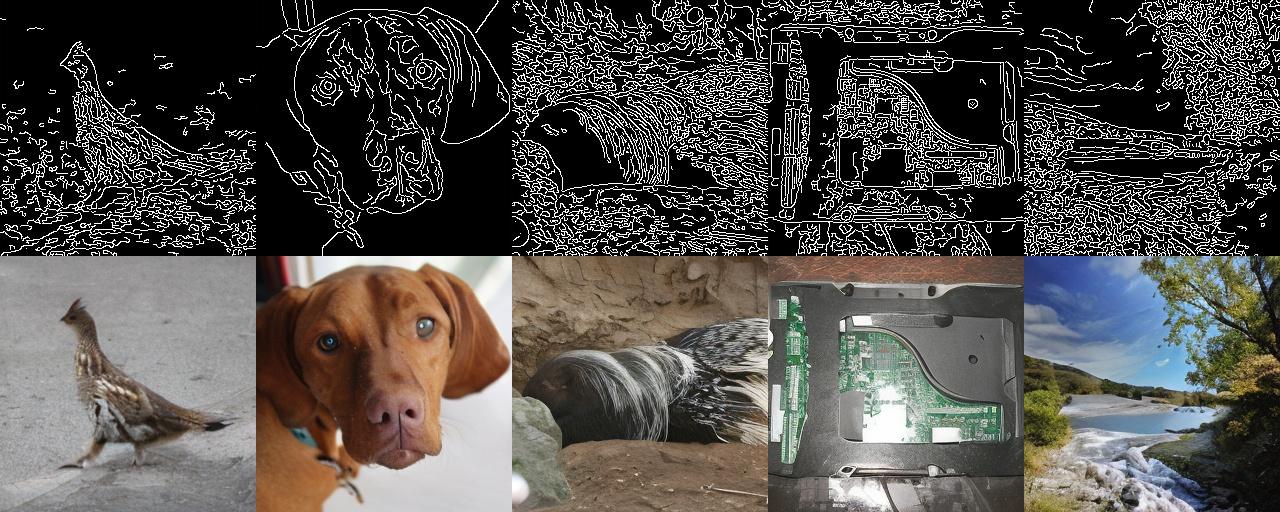}\hfill
    \includegraphics[width=0.49\linewidth]{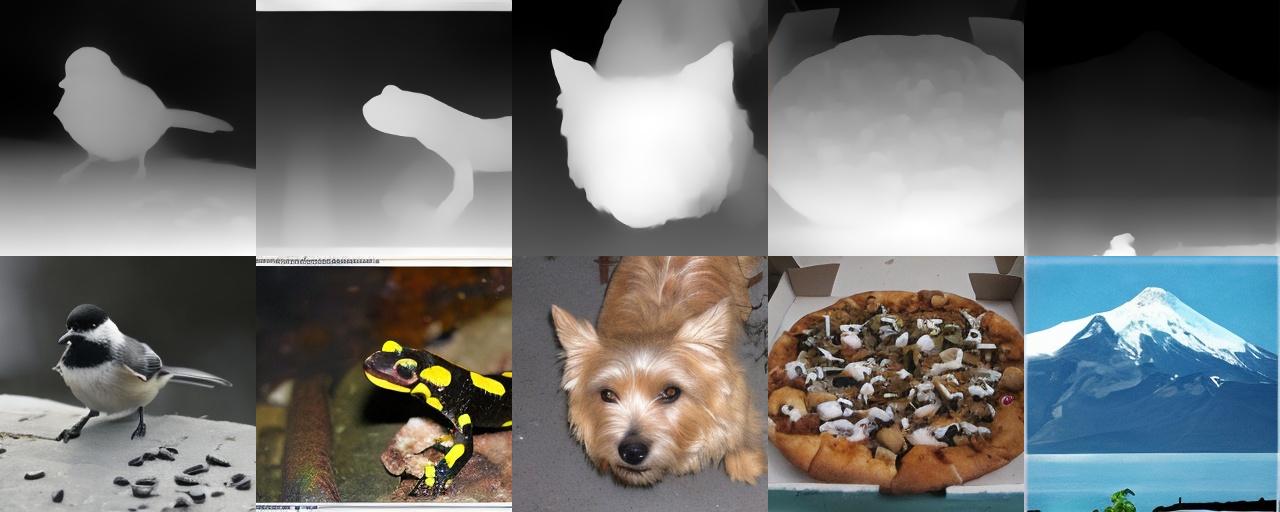}
    
    add adapter, ctrl-G=1.0

    \includegraphics[width=0.49\linewidth]{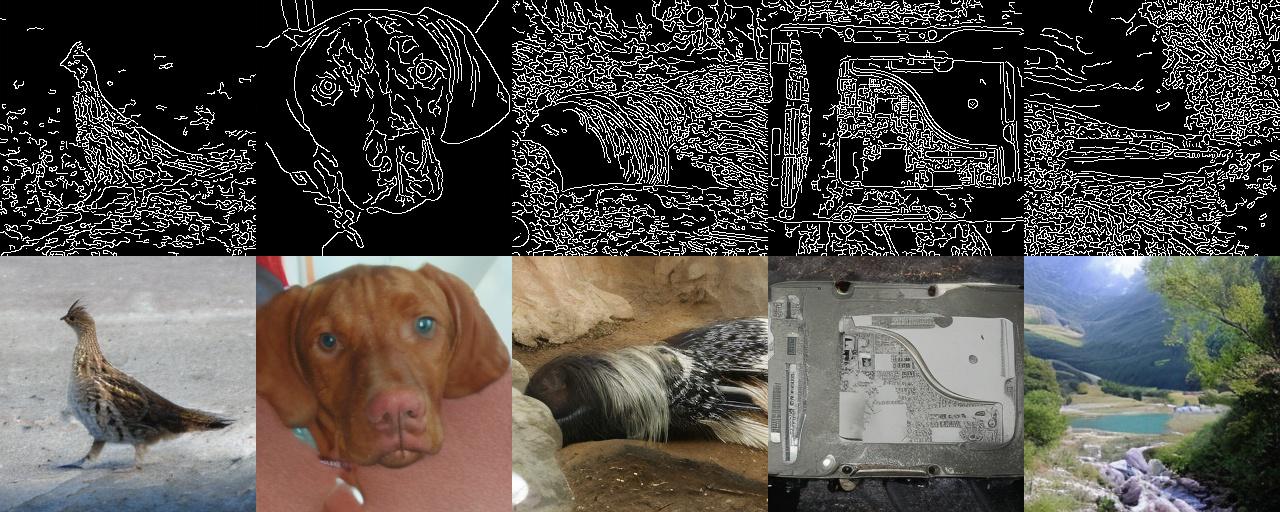}\hfill
    \includegraphics[width=0.49\linewidth]{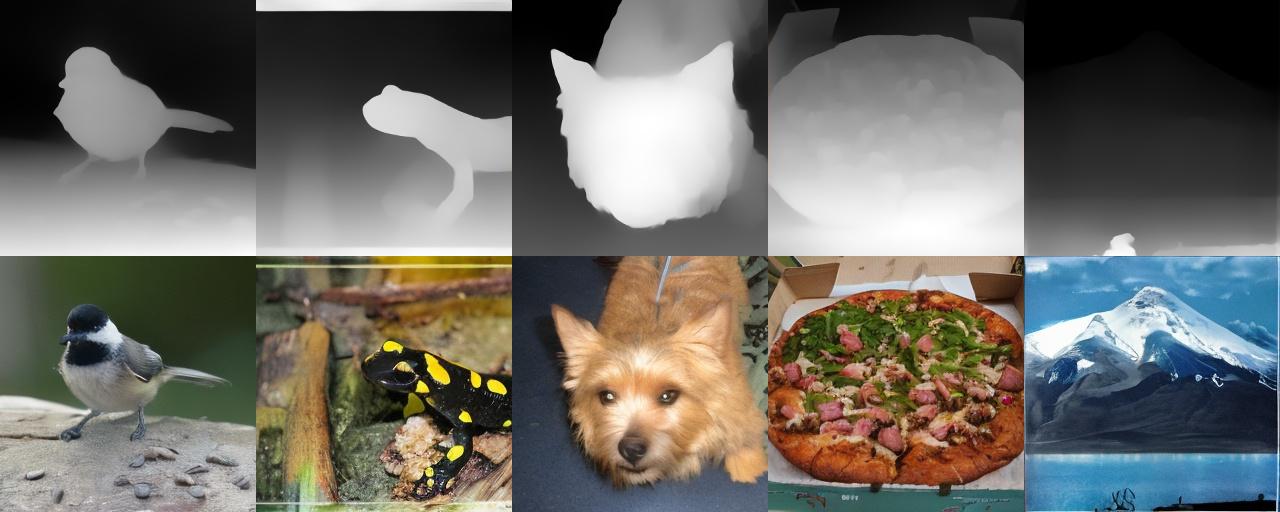}
        
    add adapter, ctrl-G=1.5

    \includegraphics[width=0.49\linewidth]{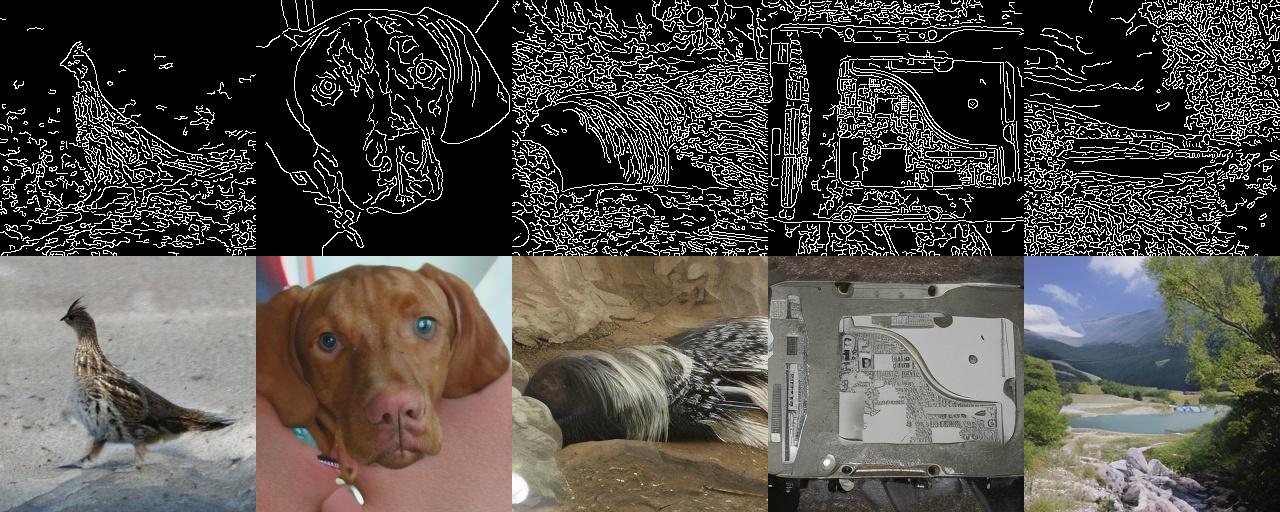}\hfill
    \includegraphics[width=0.49\linewidth]{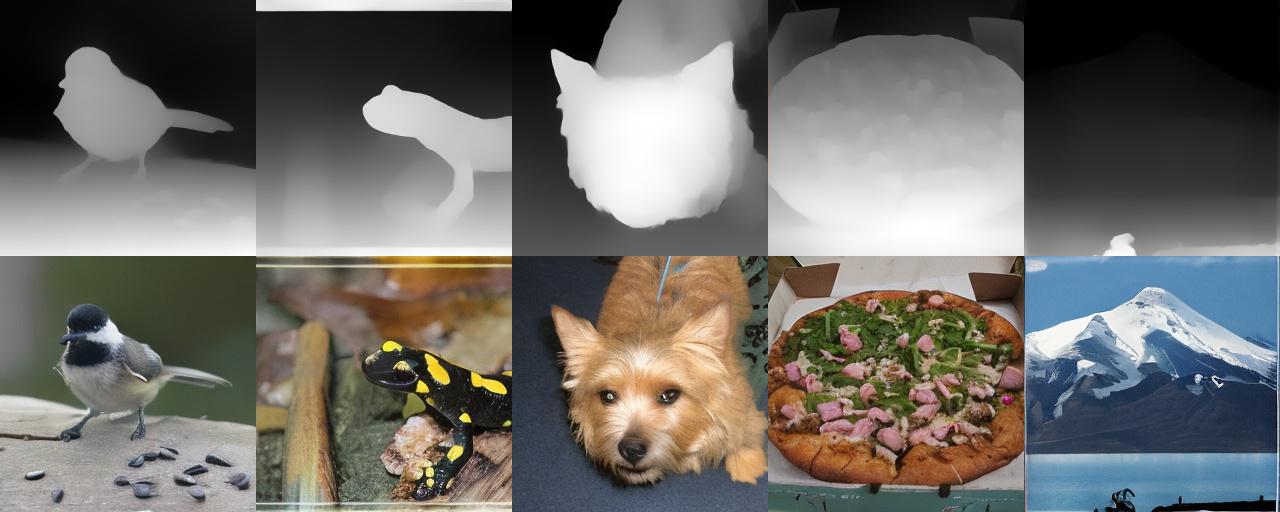}
    
    prefill adapter, ctrl-G=1.0 

    \includegraphics[width=0.49\linewidth]{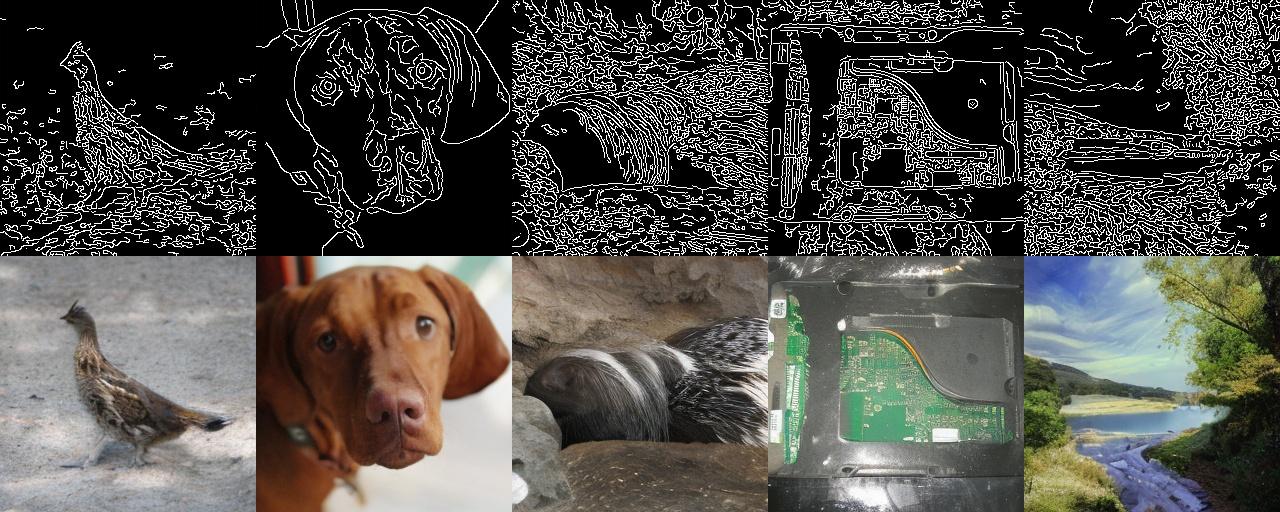}\hfill
    \includegraphics[width=0.49\linewidth]{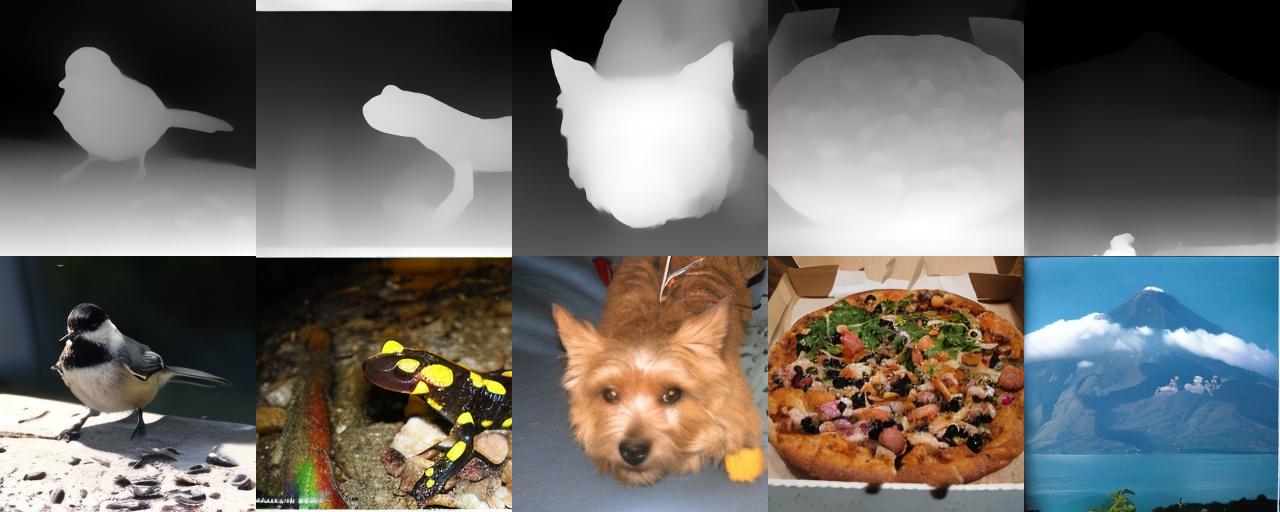}
        
    prefill adapter, ctrl-G=1.0  

    \includegraphics[width=0.49\linewidth]{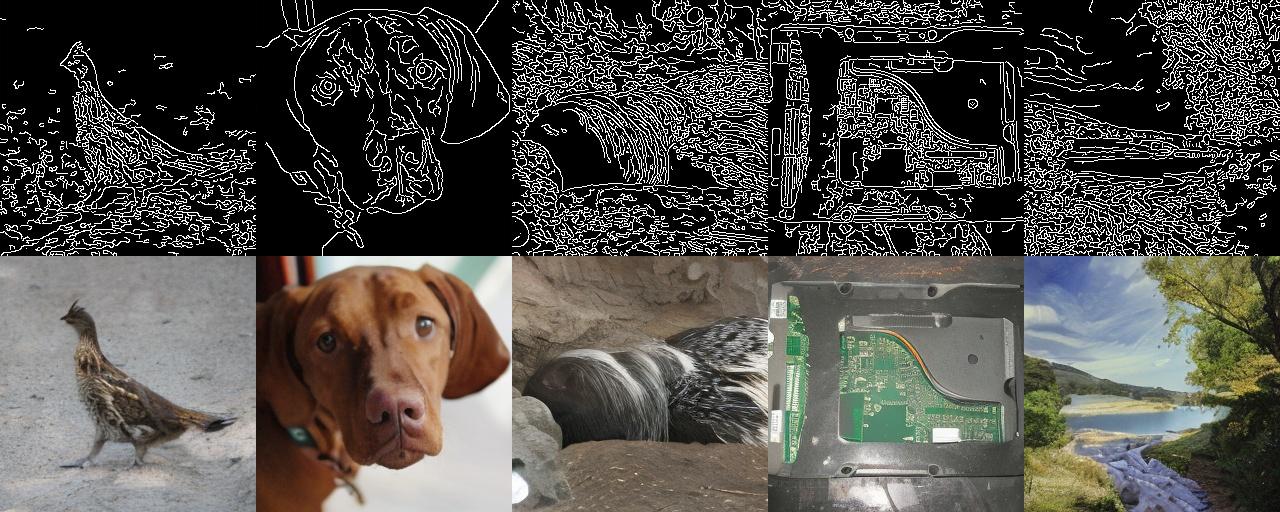}\hfill
    \includegraphics[width=0.49\linewidth]{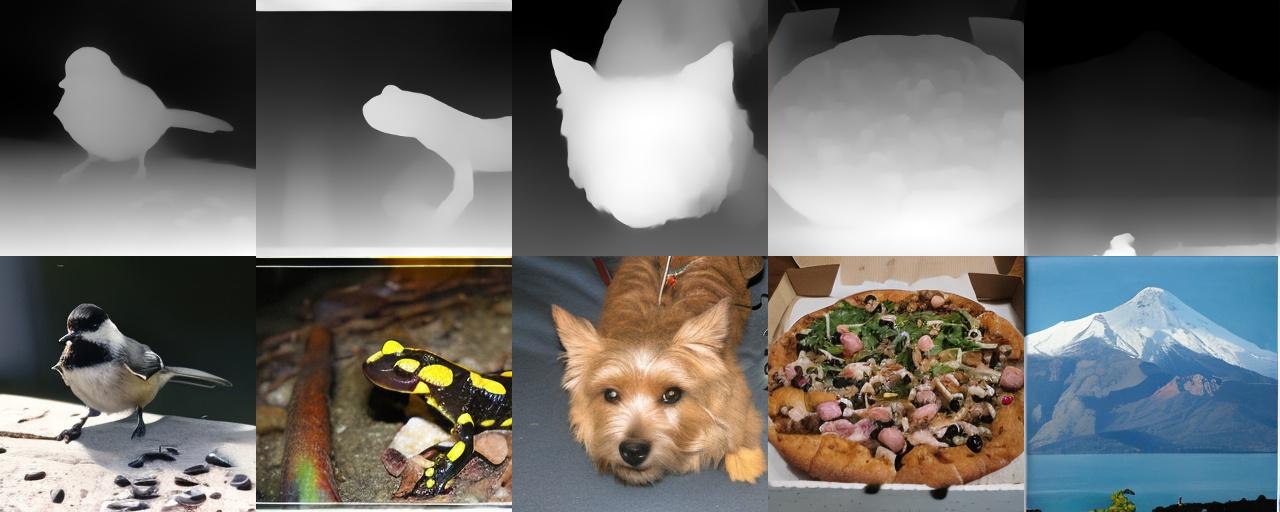}
    \caption{Additional examples of SiT generations. (CFG=3.0, temp=1.0, Euler ODE, steps=64, proj-G)}
    \label{fig:sit-ex1}
\end{figure}

\begin{figure}
    \centering
    prefill 
    
    \includegraphics[width=0.49\linewidth]{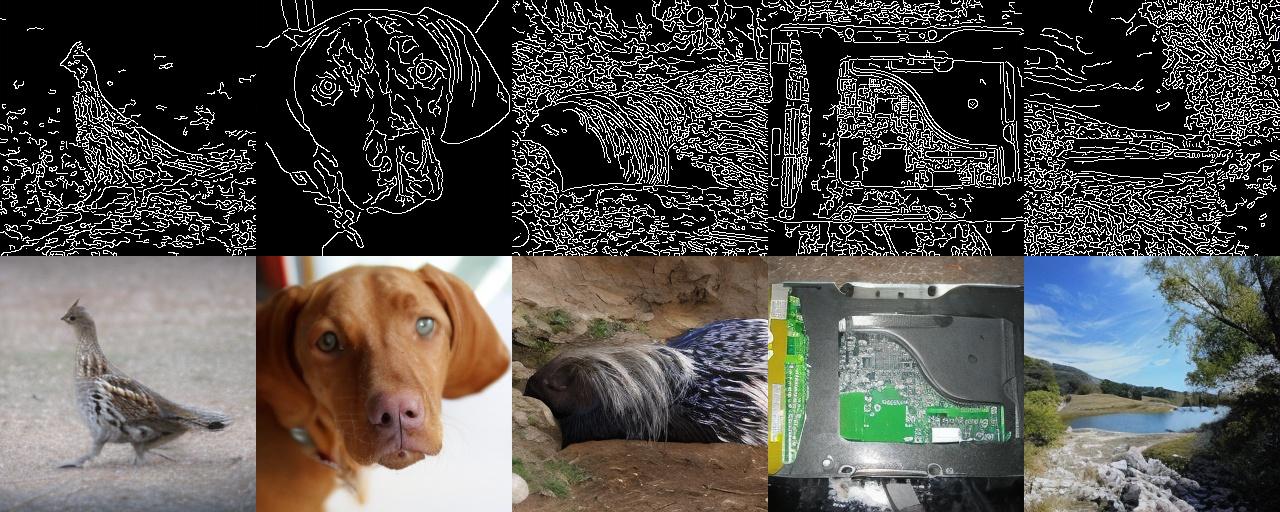}\hfill
    \includegraphics[width=0.49\linewidth]{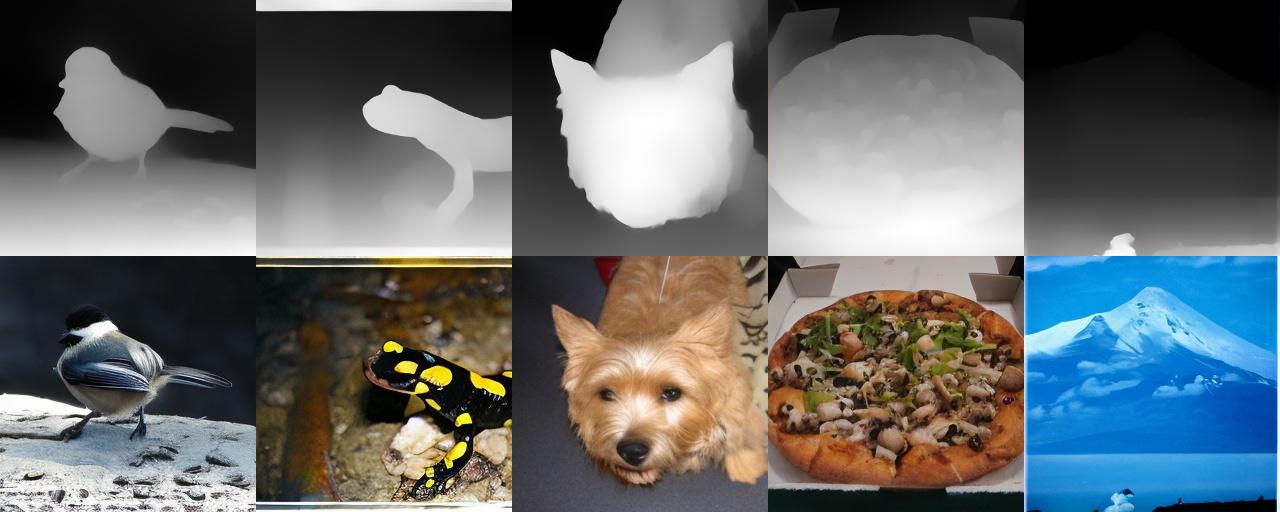}

    add adapter 

    \includegraphics[width=0.49\linewidth]{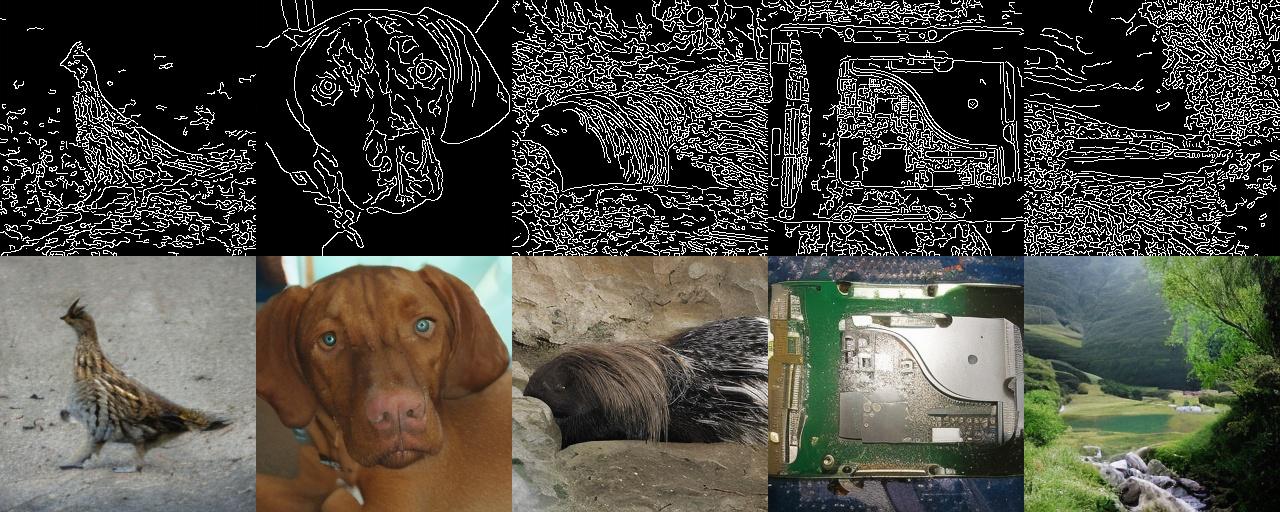}\hfill
    \includegraphics[width=0.49\linewidth]{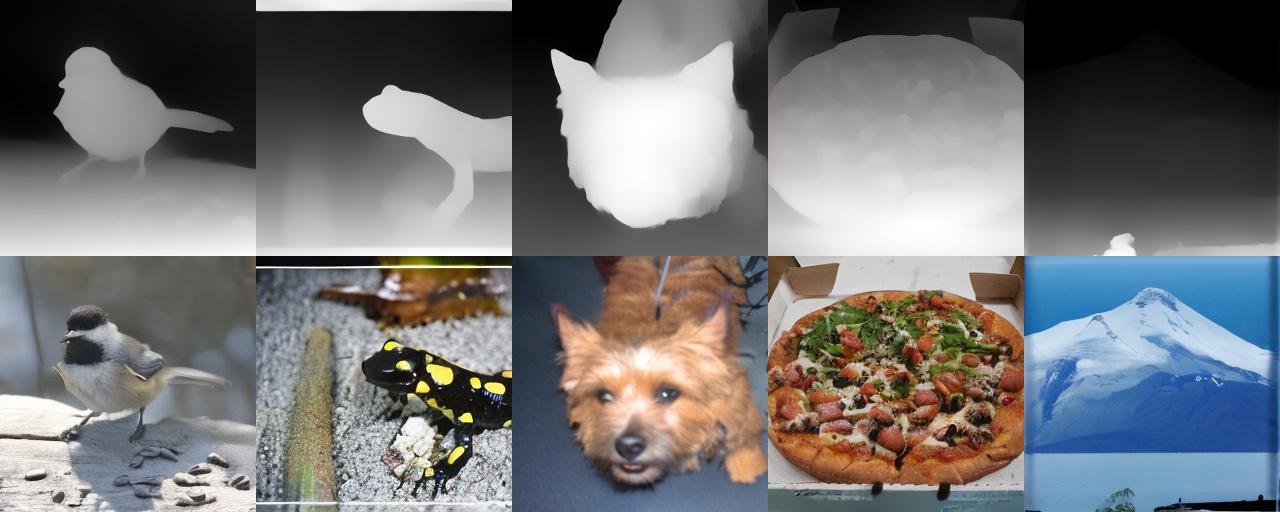}
    
    prefill adapter 

    \includegraphics[width=0.49\linewidth]{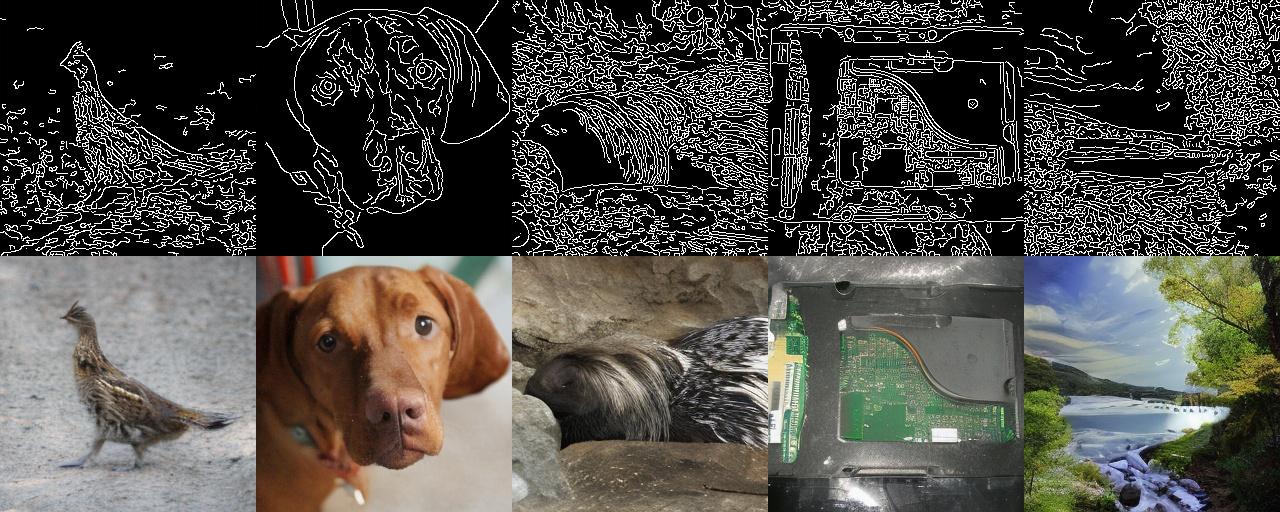}\hfill
    \includegraphics[width=0.49\linewidth]{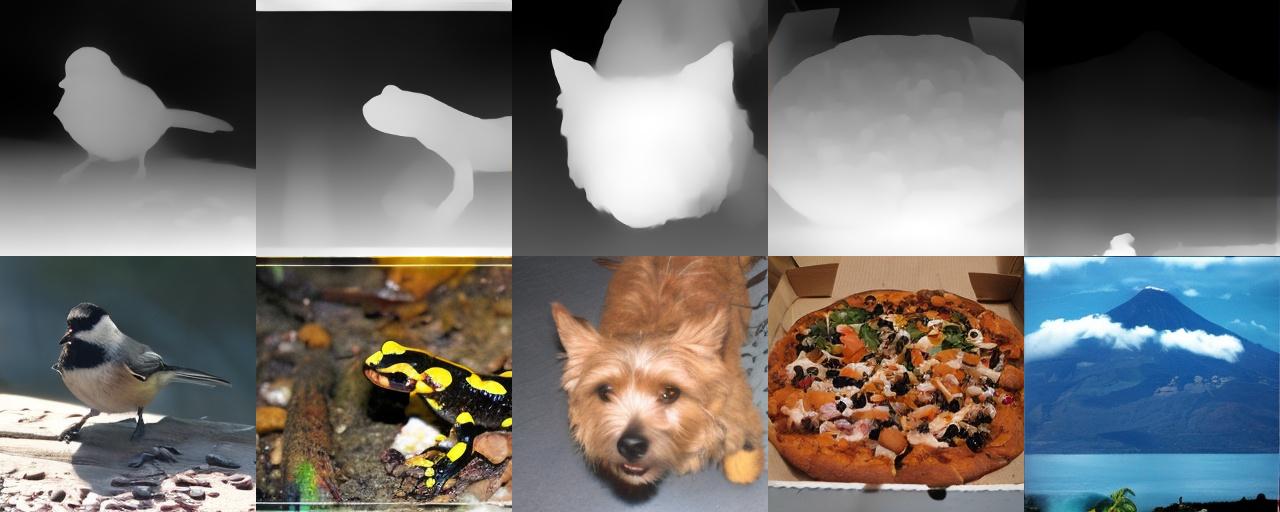}
    \caption{Additional examples of SiT generations with limited control data -- ``organism'' classes excluded. (CFG=3.0, temp=1.0, Euler ODE, steps=64, ctrl-G=1.0, proj-G)}
    \label{fig:sit-ex2}
\end{figure}

We include additional example generations over various experimental settings in \cref{fig:var-ex1,fig:var-ex2,fig:sit-ex1,fig:sit-ex2}. Readers can observe the effect of ctrl-G (canny: facial features of the dog, depth: seeds/stones next to the bird) across \cref{fig:var-ex1,fig:sit-ex1}. \cref{fig:var-ex2} show further examples of VAR ``forgetting'' ``organism'' concepts from ImageNet-pretraining. We note these are more subtle for depth (plumage on bird, stripes rather than spots on salamander).

\section{Ctrl-G on Pre-trained Adapters for T2I Models}\label{app:ctrlnet}

\begin{figure}
    \centering
    \includegraphics[width=0.57\linewidth]{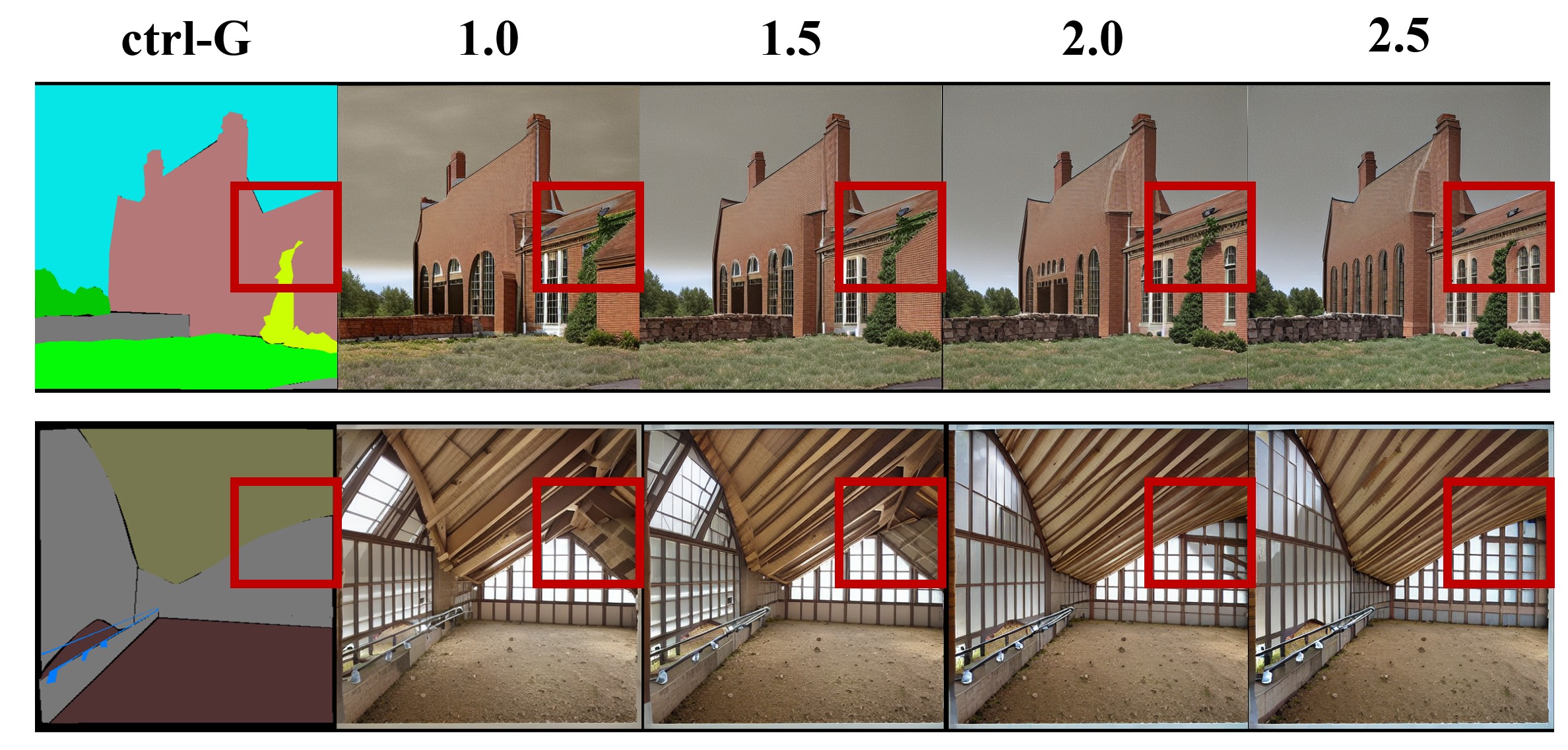}
    \includegraphics[width=0.4\linewidth]{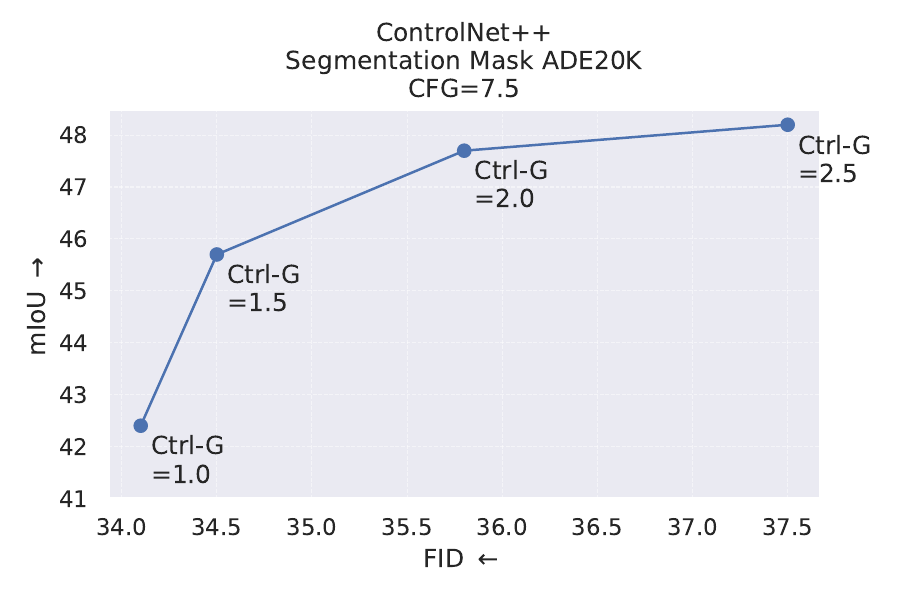}
    \caption{Ctrl-G can be applied to pre-trained opensource control adapters out of the box without the need for additional training. \textbf{Left}: Qualitative effect of increasing ctrl-G for ControlNet++ -- consistency can be visually improved. \textbf{Right}: Quantitative tradeoff between consistency (segmentation mIoU) and quality (FID).}
    \label{fig:t2i-tradeoff}
\end{figure}

We demonstrate that ctrl-G can be directly applied to existing opensource adapters without the need for any additional training. We show the effect of increasing ctrl-G on ControlNet++ \citep{controlnet_plus_plus}, a control adapter for StableDiffusion 1.5 \citep{LDM} that produces 512$\times$512 resolution text-prompt-conditioned images. \cref{fig:t2i-tradeoff} illustrates qualitative improvements in control consistency (the shape of vegetation on the building, the curve of the window), whilst also quantitatively showing the control consistency vs FID tradeoff that ctrl-G introduces, aligning with earlier results in \cref{fig:ctrl-g}. We use the evaluation code of \citet{controlnet_plus_plus}\footnote{\url{https://github.com/liming-ai/ControlNet_Plus_Plus}} and calculate FID between the validation and generation images. 

\end{document}